\documentclass[]{article}

\usepackage{authblk}
\usepackage{amsmath}
\usepackage{amsfonts}
\usepackage{booktabs}
\usepackage{tabularx}
\usepackage{graphicx}
\usepackage{subfig}
\usepackage{booktabs}
\usepackage{afterpage}
\usepackage{rotating}
\usepackage{forest}
\usepackage[]{siunitx}
\usepackage{url}
\usepackage[space]{grffile}
\usepackage{multirow}
\usepackage[a4paper, portrait, margin=1in]{geometry}

\DeclareMathOperator*{\argmax}{arg\,max}  
\DeclareMathOperator*{\argmin}{arg\,min} 
\providecommand{\keywords}[1]{\textbf{\textit{Index terms---}} #1}

\begin{document}

\title{A Tutorial on Modeling and Inference in Undirected Graphical Models for Hyperspectral Image Analysis}
\author[]{Utsav B. Gewali}
\author[]{Sildomar T. Monteiro}
\affil[]{Chester F. Carlson Center for Imaging Science, Rochester Institute of Technology, Rochester, NY\\Email: ubg9540@rit.edu}
\date{}
\maketitle

\begin{abstract}
Undirected graphical models have been successfully used to jointly model the spatial and the spectral dependencies in earth observing hyperspectral images. They produce less noisy, smooth, and spatially coherent land cover maps and give top accuracies on many datasets. Moreover, they can easily be combined with other state-of-the-art approaches, such as deep learning. This has made them an essential tool for remote sensing researchers and practitioners. However, graphical models have not been easily accessible to the larger remote sensing community as they are not discussed in standard remote sensing textbooks and not included in the popular remote sensing software and toolboxes. In this tutorial, we provide a theoretical introduction to Markov random fields and conditional random fields based spatial-spectral classification for land cover mapping along with a detailed step-by-step practical guide on applying these methods using freely available software. Furthermore, the discussed methods are benchmarked on four public hyperspectral datasets for a fair comparison among themselves and easy comparison with the vast number of methods in literature which use the same datasets. The source code necessary to reproduce all the results in the paper is published on-line to make it easier for the readers to apply these techniques to different remote sensing problems.
\end{abstract}


\section{Introduction} 
Land cover mapping (also called land cover classification and land cover segmentation) is the process of identifying the materials under each pixel of a spaceborne or an airborne image to create a map showing the spatial distribution of materials over the imaged geographical region. Hyperspectral imaging is an important technology for land cover mapping as it allows for the separation of scene materials into finer classes, compared to other sensing modalities, such as panchromatic, synthetic aperture radar, and multispectral imaging. This is because hyperspectral image at each pixel captures more information about the chemical properties of the materials by recording the reflectance/radiance at hundreds of narrow contiguous bands over the visible and infrared region. Due to its advantages, hyperspectral land cover mapping have been applied to a variety of problems such as separating various species of trees in the forest~\cite{dalponte2013tree}, identifying roads, buildings, trees, and other land covers in urban areas~\cite{heiden2007determination}, mapping the presence of minerals in soil and rocks~\cite{murphy2013mapping}, differentiating weeds from crops in agricultural land~\cite{martin2011assessing}, and studying the change in an area by comparing land cover maps at multiple dates~\cite{pu2008invasive}. 

The radiance or reflectance spectrum at each pixel of a hyperspectral image captures the interaction between light and the material, which is dependent on the atomic and the molecular structure of the material and can be used as a signature to discriminate different classes of materials. Traditional classifiers only utilized the spectrum at the pixel to determine the class of the pixel. These are called pixel-wise classifiers. Pixel-wise classifiers can be hand-designed by human experts, such as USGS's tetracoder~\cite{clark2003imaging}, or be statistical and machine learning based, such as support vector machines (SVM)~\cite{melgani2004classification}, and random forests~\cite{ham2005investigation}. Classifiers can also be supervised (requiring a set of labeled sample spectra belonging to each materials in the scene a priori) or unsupervised (not requiring any labeled spectra a priori). Pixel-wise classifiers tend to produce noisy and spatially incoherent land cover maps due to the spatial variations in illumination, shadows, purity of pixels, viewing geometry, atmospheric conditions, and noise across the image. This problem can be alleviated by combining spatial contextual information with the spectral information.       

\subsection{Spatial-spectral classification}
Spatial information can be utilized together with spectral information to produce more accurate and spatially coherent land cover maps~\cite{fauvel2013advances}. Land covers in the environment tend to be much larger than the ground pixel size of the sensors leading to regions of pixels belonging to a common material class. Additionally, some land cover classes are more likely to exists in close vicinity than others and some land cover classes are highly unlikely to occur together. This leads to strong relationships between the neighboring pixel labels in an image. For example, if a pixel belongs to a class, say building, there is a high probability that the surrounding pixels also belong to the same class, building. Similarly, the probability of neighboring pixels of a building pixel belonging to the road class or the parking lot class is typically much higher than them belonging to the forest class or the bare soil class in an urban scene. These kind of relationships can be exploited by spatial-spectral classifiers. Even though there are exceptions, e.g., \cite{camps2006composite}, \cite{wang2014spatial}, and \cite{li2017spectral}, the vast majority of spatial-spectral classifiers can be categorized into two distinct groups---methods that perform spatial-spectral feature extraction followed by pixel-wise classification and methods that combine undirected graphical model and pixel-wise classification.

Spatial-spectral feature extraction utilizes the spectra of the neighborhood of pixels around the pixel to compute feature for the pixel. The features are designed to simultaneously capture the spatial context and be highly discriminative. Classical approach used hand-designed spatial-spectral features, such as co-occurrence matrices~\cite{haralick1973textural} and extended morphological features~\cite{benediktsson2005classification}. Modern methods utilized feature representation which is learned from the data itself in supervised or unsupervised manner using sparse dictionary learning and deep learning approaches,e.g., \cite{fang2014spectral}, \cite{chen2014deep},  and \cite{yue2015spectral}.     

Undirected graphical models (UGMs)~\cite{koller2009probabilistic} are powerful and flexible probabilistic models that can represent the complex relationships occurring between the different scene elements in hyperspectral images~\cite{tarabalka2010svm,zhong2010learning}. They can be combined with pixel-wise classifiers (with or without spatial-spectral features) to enforce spatial contexts. There are other ad-hoc methods, such as majority voting over segmented image~\cite{tarabalka2009spectral}, that can be used for spatial contexts. However, compared to them UGMs are more principled approach which can model more complex relationships and has sound theory, theoretical guarantees, and efficient inference algorithms. In this tutorial, we provide an introduction to the theory of undirected graphical models, review graphical model based spatial-spectral classification methods published in literature, show how to implement graphical model based spatial-spectral classifiers using freely available toolboxes, and benchmark the discussed methods on four public hyperspectral datasets. We experiment with different supervised machine learning based classifiers and two classes of UGMs. Since popular remote sensing software, such as ENVI, do not include UGM based processing, we have published the code necessary to reproduce all the results in this paper here~\footnote{\url{https://github.com/UBGewali/tutorial-UGM-hyperspectral}}.  

This tutorial is organized as follows. Section~\ref{tutorial_related_works} reviews UGM based methods in remote sensing literature, Section~\ref{tutorial_background} provides a brief background on the UGM theory and inference algorithms, Section~\ref{tutorial_experiments} discusses the benchmarking experiments and results, and Section~\ref{tutorial_summary} summaries the tutorial. 

\section{Undirected graphical models in remote sensing}
\label{tutorial_related_works}
Since their introduction to remote sensing in \cite{solberg1996markov} and \cite{schroder1998spatial}, undirected graphical models have been widely used to model spatial dependencies in remotely sensed images for land cover mapping. Classical approaches utilize a grid-structured pairwise pixel-based Markov random field (MRF) to model the pixel dependencies and use optimization algorithms, such as iterated conditional mode (ICM), simulated annealing, and graph cuts for inference, see the reviews by \cite{schindler2012overview} and  \cite{moser2013land}. These models are defined over a grid-structured graph with each node representing a pixel label and edges present between 4-connected neighboring pixels. The unary potentials are defined at each node and captures the spectral information while the pairwise potentials are defined at edges connecting neighboring pixels and captures the spatial information. Since these models only contain unary and pairwise interactions, they are called pairwise models. Unary potentials are derived from pixel-wise classifiers, such as Gaussian maximum likelihood classifier~\cite{jackson2002adaptive}, logistic regressions~\cite{li2012spectral}, probabilistic support vector machines~\cite{tarabalka2010svm}, Gaussian mixture models~\cite{li2014hyperspectral}, Gaussian processes~\cite{yao2010novel}, and ensemble methods~\cite{merentitis2014ensemble,xia2015spectral,li2015hyperspectral}. Potts model (including its contrast sensitive version) is the most popular pairwise potential function. However, an edge based Potts potential function~\cite{tarabalka2010svm}, which only encourages neighboring pixels to have same label only if there is no edge in the intensity between them, has shown to produce better results. The hyperparameters of potential functions are mostly chosen by grid search over validation set. However, optimization schemes such as genetic algorithm~\cite{tso1999classification}, Ho–Kashyap method~\cite{moser2013combining}, and Bayesian optimization~\cite{gewali2017Using} have shown to perform better than grid search.

Conditional random fields (CRF) are type of MRF that model conditional distributions by making potential functions dependent on the input features. \cite{zhong2010learning} first utilized a CRF to map the land covers in hyperspectral images. The standard CRF uses with log-linear potential functions which have many parameters that have to be learned from the data. Hence, large number of training pixels is required while using CRF for land cover mapping. This problem has been tackled by modifying the CRF to use simpler functions for pairwise potentials. Some of the proposed pairwise potential functions are based on Euclidean distance between spectra~\cite{zhang2012simplified}, Mahalanobis distance between spectra~\cite{zhong2014support}, and Euclidean distance between pixel coordinated and spectra combined with class label cost~\cite{zhao2015detail}. These methods utilize grid-structured graph and learn parameters by maximum likelihood estimation. These methods utilize loopy belief propagation during learning and inference.

Though not as common as supervised classification, undirected graphical models have been combined with unsupervised classifiers, such as k-means and ISODATA, for mapping ground covers without any user supplied ground truth~\cite{li2014review}. Tree-structured pairwise Markov random field which performs binary segmentation recursively at each level of the tree has been proposed for unsupervised classification~\cite{d2003tree}. Active learning strategies have also been developed for UGM based spatial-spectral land cover mapping. These methods iteratively select pixels from a set of unlabeled pixels based on some strategy and ask user to manually label them so that the those pixels can be added to the training set to further improve the model. \cite{sun2015mrf} proposed selecting the unlabeled pixels which are differently labeled by the pixel-wise classifier and the combination of pixel-wise classifier and UGM. Similarly, \cite{li2011hyperspectral, li2013spectral} defined heuristics, such as breaking ties, over the posterior class probabilities predicted by the random field to select the appropriate unlabeled pixels. Methods that combine results from ensembles of UGMs trained on different features~\cite{zhong2007multiple} or ensembles of UGMs with different forms of unary potential functions~\cite{zhong2014hybrid} also been published.

Higher order graphical models contain potential functions defined over more than two nodes and are more expressive compared to the grid-structured pairwise models with unary and pairwise potentials. Since, they include potential functions over group of pixels rather than just individual pixels, they can model complex dependencies between various regions, structures, and objects in the image. \cite{zhong2011modeling} proposed using robust $P^n$ model for higher order modeling of hyperspectral images. In this method, the image is first segmented using a clustering (unsupervised classification) algorithm. Then, higher order potentials are defined over each segment to encourage all the pixels in each segment to be assigned the same label, which are used along with the unary and pairwise potentials. The $P^n$ model has been widely used for land cover mapping using other remote sensing modalities as well~\cite{montoya2015semantic, wegner2013higher, li2015robust, niemeyer2016hierarchical}. Similarly, \cite{zhao2016high} proposed a novel higher order potential over each segment based on the distance between the segments and the similarity between the pixels in the segments. Higher order relationships can also be modeled the simple pairwise model if each node of the graph is representing a label for the entire segment (group of pixels) rather than a single pixel~\cite{roscher2014superpixel,zhang2015superpixel,tuia2016getting}. These methods are less expressive than higher order potential based methods in that all the pixels in each segment are strictly assigned to the same label. In a different application, \cite{albert2017higher} used two CRF layers, one to model land cover (broader classes such as buildings, grasses etc.) and another to model land use (finer classes such as residential buildings, non-residential buildings, urban grass, grass land etc.) simultaneously in aerial imagery using intra-layer and inter-layer interactions.  

Undirected graphical models have also been used for sub-pixel mapping of remote sensing images~\cite{kasetkasem2005super, zhao2015sub,wang2013subpixel}. These methods produce land cover maps at a scale smaller than the size of a image pixel. This is done by estimating the contents of pixels using classification or unmixing, using the derived material proportions as the unary potentials of UGM at finer resolution, and using pairwise potentials at finer resolution to enforce spatial contexts.

Various studies~\cite{solberg1996markov,liu2008using,melgani2003markov} have also utilized three-dimensional grid-structured UGMs for modeling spatio-temporal dependencies in a time-series of multispectral satellite images for land cover classification and change detection. Apart from land cover mapping, undirected graphical models have also been used to model the spatially dependent parameters of Bayesian hyperspectral unmixing frameworks~\cite{altmann2014residual, altmann2015bayesian, eches2011enhancing, eches2013adaptive}. Continuous Markov random fields have been successfully used to model textures in hyperspectral images, e.g. \cite{rellier2004texture},  however they are beyond the scope of this tutorial.  

\section{Background}   
\label{tutorial_background}
Undirected graphical models define the joint distribution of a set of variables over the structure of an undirected graph~\cite{koller2009probabilistic,nowozin2011structured}. The nodes of the undirected graph represent the variables while the edges between the nodes express the conditional independence relationship between the variables.  

Let $\mathbf{y}=\left[y_1,...,y_N\right]$ be a vector of N variables whose joint probability distribution is defined over an undirected graph $G$ such that following conditional independence relationships are true.  

\paragraph*{Local Markov property} 
Each node is conditionally independent of all of the other nodes given its neighboring nodes. 
\begin{equation}
p(y_i|\mathbf{y}_{\setminus i}) = p(y_i|\mathcal{N}(y_i)), 
\end{equation}
where $\mathcal{N}(y_i)$ is the set of neighbors of $y_i$.
\paragraph*{Global Markov property} 
Two nodes are conditionally independent if all the path between them along the edges in the graph is blocked by an observed node.
\begin{equation}
p(y_i|y_j,y_S) = p(y_i|y_S),
\end{equation}
where $y_S$ are the set of nodes separating $y_i$ and $y_j$ in $G$.
\paragraph*{}
Then, the Hammersley-Clifford theorem~\cite{besag1974spatial} states that the joint distribution of the variables $y_1,...,y_N$ can be factorized as 
\begin{equation}
p\left(\mathbf{y}\right) = \frac{1}{Z} \prod_{ C \in \mathcal{C}(G)}  \psi_C \left( \mathbf{y_C} \right),
\label{eq:UGM_def1}
\end{equation}
where $\mathcal{C}(G)$ is the set of all the cliques of $G$. A clique (also called maximal subgraphs) is a subset of nodes of a graph that has an edge between every pair of nodes. $\mathbf{y_C}$ denotes a vector of all of the nodes inside the clique $C$. The functions $\psi_C \left( \mathbf{y_C} \right)$ are arbitrary non-negative functions that define the interaction between the variables inside the clique $C$ and are called potential functions. $Z$ is a normalizing constant given by    
\begin{equation}
Z = \sum_{\mathbf{y} \in \mathcal{Y}} \prod_{ C \in \mathcal{C}(G)}  \psi_C \left( \mathbf{y_C} \right),
\label{eq:UGM_def2}
\end{equation}
and is called a partition function. The partition function computes the sum of the product of potential functions over the set of all possible configurations of the variables $y_1,...,y_N$, denoted by $\mathcal{Y}$. Division by the partition function makes the product of potential functions a valid probability that sums to one. 

In this way, depending upon the structure of the undirected graph $G$ and the form of the potential functions $\psi_C \left( \mathbf{y_C} \right)$, UGMs can model a wide variety of families of probability distributions over the variables $y_1,...,y_N$. UGMs can represent the probability distribution of both real and discrete variables, however this tutorial only focuses on discrete UGMs. Also, though we include a broad introduction to UGMs in the tutorial, only pairwise models are explored in detail in explanation and experimentation. It is because pairwise models are the most widely tested and established models in remote sensing. Also, a good understanding of pairwise models is essential to understand the newer, more complex, higher order models.

\subsection{Markov random fields}
Markov random fields (MRF) is another name for UGMs~\cite{murphy2012machine,blake2011}. However, in this tutorial, similar to many literature, the term MRF is primarily used to denote models representing unconditional distributions (models not conditioned on input features), in order to contrast them with the conditional random fields (CRF).     

It is very common to represent (\ref{eq:UGM_def1}) in terms of energies by choosing the potentials to be of exponential family, $\psi_C(\mathbf{y_C}|\mathbf{w}) = \exp(-E_C(\mathbf{y_C}|\mathbf{w}))$ where $E_C(\mathbf{y_C};\mathbf{w})=-\log(\psi_C(\mathbf{y_C}|\mathbf{w}))$ is the clique energy function. Since, the potential functions contain free parameters in practice, the potential functions and the energy functions have been parameterized with parameter vector $\mathbf{w}$. Then, the joint probability of the MRF is 
\begin{equation}
p(\mathbf{y}|\mathbf{w}) = \frac{1}{Z(\mathbf{w})} \exp \left( -E\left(\mathbf{y};\mathbf{w}\right) \right),
\label{eq:MRF_def}  
\end{equation}
where $E\left(\mathbf{y};\mathbf{w}\right)=\sum_{ C \in \mathcal{C}(G)}E_C(\mathbf{y_C};\mathbf{w})$ is the total energy and  $Z(\mathbf{w}) = \sum_{\mathcal{Y}} \exp \left( -E\left(\mathbf{y};\mathbf{w}\right) \right)$. 

Pairwise MRF is the simplest MRF formulation which expresses the total energy as the sum of unary energies and pairwise energies. The unary energy is defined for all the nodes and the pairwise energy is defined for all the edges in the graph. Each node has a different unary energy based on the value assigned to it, with the likely assignments having lower energy. Similarly, each edge exhibits different pairwise energy for different configuration of possible values of the two nodes at its ends, with likely configurations having lower energy. The total energy in a pairwise MRF is    
\begin{equation}
E\left(\mathbf{y};\mathbf{w}\right) = \sum_{i \in V} E_i\left(y_i;\mathbf{w}_1\right) + \sum_{(i,j) \in D} E_{ij}\left(y_i,y_j;\mathbf{w}_2\right),
\label{eq:MRF_pairwise} 
\end{equation}
where $E_i\left(y_i;\mathbf{w}_1\right)$ is the unary energy of the i\textsuperscript{th} variable when its value is $y_i$ and $E_{ij}\left(y_i,y_j;\mathbf{w}_2\right)$ is the pairwise energy between the i\textsuperscript{th} and the  j\textsuperscript{th} variables when their values are $y_i$ and $y_j$ respectively.  $V$ is the set of all nodes and $D$ is the set of all edges in the graph $G$. $\mathbf{w}=[\mathbf{w}_1,\mathbf{w}_2]$ are the parameters of the energy functions.

\subsection{Conditional random fields}
The conditional random field (CRF)~\cite{sutton2011} is a type of MRF whose clique potentials are conditioned on input features. It is a discriminative version of MRF that models $p(\mathbf{y}|\mathbf{x})$ instead of $p(\mathbf{y})$, where $\mathbf{x} = \left[x_1,...,x_N\right]$  are the input features for $\mathbf{y} = \left[y_1,...,y_N\right]$, as 
\begin{equation}
p\left(\mathbf{y}|\mathbf{x},\mathbf{w}\right) = \frac{1}{Z(\mathbf{x},\mathbf{w})} \prod_{ C \in \mathcal{C}(G)}  \psi_C \left( \mathbf{y_C}|\mathbf{x},\mathbf{w}\right),
\label{eq:CRF_def1}
\end{equation}
where $Z(\mathbf{x},\mathbf{w}) = \sum_{\mathbf{y} \in \mathcal{Y}} \prod_{ C \in \mathcal{C}(G)}  \psi_C \left( \mathbf{y_C|\mathbf{x},\mathbf{w}} \right)$ and $\mathbf{w}$ is the vector of potential function parameters. The potential functions of CRFs are most commonly represented by the log-linear function as $\psi_C \left( \mathbf{y_C}|\mathbf{x},\mathbf{w}\right)=\exp\left( \mathbf{w_C}^T\phi(\mathbf{x_C},\mathbf{y_C})\right)$, where $\phi(\mathbf{x_C},\mathbf{y_C})$ is a feature function and $\mathbf{w_C}$ is a weight vector. In terms of energies, the clique energy $E_C \left( \mathbf{y_C};\mathbf{x},\mathbf{w}\right)= -\mathbf{w_C}^T\phi(\mathbf{x_C},\mathbf{y_C})$. The feature function produces an arbitrary length vector of features dependent on  $\mathbf{x_C}$ and $\mathbf{y_C}$.

Similar to MRF, a pairwise CRF model can be defined as

\begin{equation}
\label{eq:pairwise_crf}
p(\mathbf{y}|\mathbf{x},\mathbf{w}) = \frac{1}{Z(\mathbf{x},\mathbf{w})} \exp \left( \sum_{i \in V} \mathbf{w}^T_1 \phi_1(x_i, y_i) + \sum_{(i,j) \in D}  \mathbf{w}^T_{y_i,y_j} \phi_2(x_i,x_j) \right),
\end{equation}
where $\mathbf{w}=\{\mathbf{w}_1,\mathbf{w}_{y_i,y_j}\}$ are the parameters and $Z(\mathbf{x},\mathbf{w})$ is the partition function. $\phi_1(x_i,y_i)$ is the unary feature function for $i^\text{th}$ variable as a function of the input feature $x_i$ and label $y_i$. $\phi_2(x_i,x_j)$ is the pairwise feature function between the $i^\text{th}$ and $j^\text{th}$  variables, $y_i$ and $y_j$, as a function of inputs features $x_i$ and $x_j$. Separate parameter vectors are defined for each possible combinations of $y_i$ and $y_j$, represented by $\mathbf{w}^T_{y_i,y_j}$. The pairwise energy is obtained by multiplying the pairwise feature vector by appropriate pairwise weight vector based on the values of $y_i$ and $y_j$. The weight vector and the feature functions can be function can be function of the node index and be different at different nodes, however in the above formulation it is assumed that they are same across the graph.  

The main advantages of CRFs over MRFs is that being discriminative model rather than generative model they can better use data for classification and that potentials in CRFs can be made more data-dependent than MRFs due to the use of input features, while the main disadvantage of CRF is that they require larger training data and longer training time~\cite{murphy2012machine}.

\subsection{Parameter learning}
Since MRF typically have very few parameters it is very common to tune the parameters of MRF by grid-search over validation set. The parameters of CRF cannot be tuned in this manner as they are substantially large in number. Parameters of CRF can be learned by maximum likelihood estimation~\cite{nowozin2011structured}. The log-likelihood of the CRF is 
\begin{equation}
l(\mathbf{w}) = \frac{1}{N} \sum_i p(\mathbf{y}^{(i)}|\mathbf{x}^{(i)},\mathbf{w}),
\end{equation}
where N is the number of samples and  $(\mathbf{y}^{(i)},\mathbf{x}^{(i)})$ is the i\textsuperscript{th} training pair. The gradient of log-likelihood with respect to clique parameters~\cite{murphy2012machine} is given by
\begin{equation}
\frac{\partial l(\mathbf{w})}{\partial \mathbf{w_C}} = \frac{1}{N} \sum_i \left[ \psi_C \left( \mathbf{y_C}^{(i)}|\mathbf{x}^{(i)},\mathbf{w}\right)  - \mathbb{E}_{\mathbf{y_C}}\left[\psi_C \left( \mathbf{y_C}|\mathbf{x}^{(i)},\mathbf{w}\right)\right] \right].
\end{equation}

The log-likelihood function can be maximized using a gradient based optimizer. The second term in the derivative calculates expectation over marginal probability of the clique. Marginal probabilities can be estimated by the inference methods discussed below. Since, each iteration of gradient optimization requires performing inference once, maximum likelihood parameter estimation is computationally expensive for graphical models. There are other alternatives for parameter learning, such as maximizing pseudo-likelihood~\cite{besag1975statistical} and maximum margin learning~\cite{taskar2004max}. Methods like maximum likelihood and maximum pseudo-likelihood can be used for parameter estimation in MRF but maximum margin learning is only for CRFs.

\subsection{Inference}
The size for the solution space $\mathcal{Y}$ for a discrete undirected graphical model of $N$ variables where each of the $N$ variables can take $M$ distinct value is $M^N$. Hence, brute-force inference by enumerating cost of all configurations of the variables in not computationally feasible, unless the graph is very small.     

There are two popular inference approaches for undirected graphical models--maximum a posteriori (MAP) inference and probabilistic inference (also called marginal inference). Both inference approaches are NP-hard for general graphs and arbitrary potential functions however for restricted graph structure and potential function, exact inference is tractable. For example, for graphs with no loops such as chains and trees exact inference is possible by method called belief propagation~\cite{pearl1988probabilistic}. Similarly, Graph-cuts can be used to efficiently find the exact MAP inference in pairwise graphical model of binary variables if the total energy function is sub-modular~\cite{greig1989exact,kolmogorov2004energy}. For cases where exact inference is not tractable, a variety of efficient approximate algorithms have been developed. The readers are encouraged to read \cite{nowozin2011structured} for in depth coverage of exact and approximate inference algorithms.  

\paragraph*{MAP inference}
The MAP inference finds the configuration of $\mathbf{y}$ that maximizes the joint probability or equivalently minimizes the total energy as
\begin{equation}
    \mathbf{y}^{*} = \argmax_{\mathbf{y} \in \mathcal{Y}} p(\mathbf{y};\mathbf{x},\mathbf{w}) = \argmin_{\mathbf{y} \in \mathcal{Y}} E(\mathbf{y};\mathbf{x},\mathbf{w}).
\end{equation}
Some of the common algorithms used for MAP inference are iterated conditional mode (ICM), simulated annealing, graph cuts and move-making algorithms, belief propagations (including loopy and tree reweighted belief propagations), Markov chain Monte Carlo, and linear programming relaxations.   

\paragraph*{Probabilistic inference}
The probabilistic inference finds the value of the log partition function and the marginal probabilities of the cliques:
\begin{enumerate}
\item $\log Z(\mathbf{x},\mathbf{w})$
\item $p(\mathbf{y_C}|\mathbf{x},\mathbf{w}), \forall C \in \mathcal{C}(G), \forall \mathbf{y_C} \in \mathcal{Y_C}$.
\end{enumerate}
Once the marginal probability of individual variables is calculated, the label with the highest probability is generally assigned to the variable, which is sometimes called the maximum of marginals inference. This is equivalent to minimizing the expected Hamming loss while MAP inference is equivalent to minimizing the expected 0/1 loss~\cite{nowozin2011structured}. Apart from being an important inference technique, probabilistic inference is essential for maximum likelihood and other parameter estimation techniques as probabilistic inference is performed once per gradient calculation in these methods. Some of the common algorithms for probabilistic inference are belief propagations (including loopy and tree reweighted belief propagations), mean field inference, and Markov chain Monte Carlo. 

\paragraph*{}
Both inference techniques can be applied for CRFs and MRFs, however the equations in this Subsection are particularly for CRFs as the terms are conditioned for input features. The input features should be neglected when using them to denote MRFs.

\section{Experimental evaluation}
\label{tutorial_experiments}
In this section, we apply pairwise MRFs and CRFs on hyperspectral images for spatial-spectral classification. First we compare different grid-structured pairwise models defined over pixel labels, and in the second part we compare theses grid-structured pixel-based pairwise models against pairwise models defined over superpixels (image segments).   

\subsection{Hyperspectral datasets}
\label{tutorial_datasets}
We experiment on four widely used public hyperspectral datasets---(a) Indian pines~\cite{baumgardner2015band}, (b) Salinas~\cite{gualtieri1999support}, (c) University of Pavia~\cite{dell2003exploiting}, and (d) Pavia Center~\cite{licciardi2009decision} \footnote{obtained from \url{http://www.ehu.eus/ccwintco/index.php?title=Hyperspectral_Remote_Sensing_Scenes}}. The Indian Pines dataset contains an image collected by the Airborne Visible/Infrared Imaging Spectrometer (AVRIS) and a corresponding ground truth map, with the identity of the materials. The Indian Pines image captures an area of 2 $\times$ 2 miles, covering agricultural land and forest in Northwest Tippecanoe County, Indiana.  It is 145 $\times$ 145 pixels in size and its pixel diameter is around \SI{4}{\metre}. In our experiments, only the ground cover classes which were present in 200 or more pixels were used, bring the total number of classes from 16 to 12. The Salinas dataset also consist of an AVRIS image and a ground truth map. It was collected in the south of the city of Greenfield in the Salinas Valley in California and images a 512 $\times$ 217 farming area scene with 16 ground covers. Both images contain 220 spectral bands, with wavelengths ranging from \SI{400}{\nano\metre} to \SI{2500}{\nano\metre}. Twenty water absorption bands were removed from the images as pre-processing. Both of the images are not atmospherically compensated, with the pixels measured in spectral radiance.

The University of Pavia and the Pavia Center datasets were collected by the Reflective Optics System Imaging Spectrometer (ROSIS) over city of Pavia in northern Italy. The ROSIS sensor has 115 bands over the visible and near-infrared spectral range, with wavelengths ranging from \SI{430}{\nano\metre} to \SI{860}{\nano\metre}. Each pixel has a ground sampling distance of \SI{1.3}{\metre}. Twelve noisy bands were removed from the University of Pavia image and thirteen noisy bands were removed from the Pavia Center image. The University of Pavia is 610 $\times$ 340 pixels in size and the Pavia center image is 1096 $\times$ 715 pixels in size. There are nine different material classes in both the images and  ground truth land cover maps of the images are available. Both the images have been atmospherically compensated, with the spectra being measured in terms of reflectance.

\subsection{Grid-structured MRFs and CRFs}
\label{tutorial_methodology}
\begin{figure}
\centering
\includegraphics[]{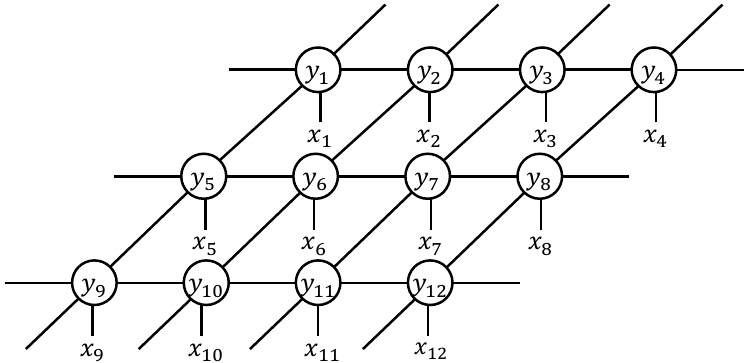}
\caption{Grid-structured graph used for MRFs and CRFs.}
\label{fig:CRF_grid}
\end{figure}

Grid-structured pairwise model is the simplest and the most widely used undirected graphical model for land cover classification. In this model, each node represents a pixel label and there is an edge between nodes representing 4-connected neighboring pixels, as shown in Figure~\ref{fig:CRF_grid}. 
In the figure, $y_i$ represents the class label assigned to the $i^{th}$ pixel and $x_i$  represent the spectrum (or any feature derived from the spectrum) measured at that pixel. There are as many nodes as there are pixels and there are edges between the nodes representing neighboring pixels. Each node can take one of the discrete values representing the class of the material. Its value indicate what material is present at the pixel location whose label is represented by that node. 

In the experiments, for the MRFs the unary energy at each pixel is derived from pixel-wise classifiers, by using the negative logarithm of the class-conditional probability, $E_i\left(y_i=c\right) = - \ln \left( P\left(  y_i=c \mid x_i \right) \right)$.  The Potts model is used for pairwise energy. The Potts model promotes spatial smoothness by penalizing when neighboring pixels are assigned different class labels. 
It is defined as 
\begin{equation}
\label{eq:potts}
E_{ij}\left(y_i,y_j\right) = 
\begin{cases}
    0,& \text{if } y_i = y_j\\
    \beta,              & \text{otherwise},
\end{cases}
\end{equation}
where $\beta$ is the cost of the labels $y_i$ and $y_j$  being different. There are many other choices for pairwise energy function, however, the Potts model is most popular for spatial-spectral classification in literature. The parameters of the Potts model can be learned using maximum likelihood estimation under probabilistic parameter learning framework. But, it is far more common to tune the parameter using grid search over a validation set. In the experiments, the parameter $\beta$ was chosen by grid search from \{0.001,0.01,0.1,1,10\}. The MRF inference was performed by 15 iterations of graph cuts with expansion move algorithm~\cite{boykov2001fast,kolmogorov2004energy,boykov2004experimental} using \cite{szeliski2008comparative}~\footnote{\url{http://vision.middlebury.edu/MRF/code/}}.

The CRF was implemented using the JGMT~\footnote{\url{https://people.cs.umass.edu/~domke/JGMT/}} toolbox. The CRF model used is exactly same as described by \eqref{eq:pairwise_crf}. A vector consisting of the class probabilities for all the M classes predicted by pixel-wise classifiers was used as the unary feature function, $\phi_1(x_i, y_i) = [P\left(  y_i=1 \mid x_i \right), P\left(  y_i=2 \mid x_i \right), ..., P\left(  y_i=M \mid x_i \right) ]^T$.  The pairwise feature function used was a constant of 1, i.e, the pairwise energy for each configuration of the two nodes was simply equal to an element in the weight vector. The truncated fitting with the clique logistic loss based on tree re-weighted belief propagation with five iterations was used for parameter learning and inference~\cite{domke2013learning}. Typically, when CRFs are used in computer vision, the parameters are learned using many labeled images in the training set. But, in remote sensing, most of the time a large number of labeled training images are not available and parameters have to be learned from few labeled pixels of the test image. This is modeled during learning by making unlabeled pixels of the test image as latent nodes and the training pixels as observed nodes during training.

\subsubsection{Results}
\begin{figure}
\centering
\includegraphics[]{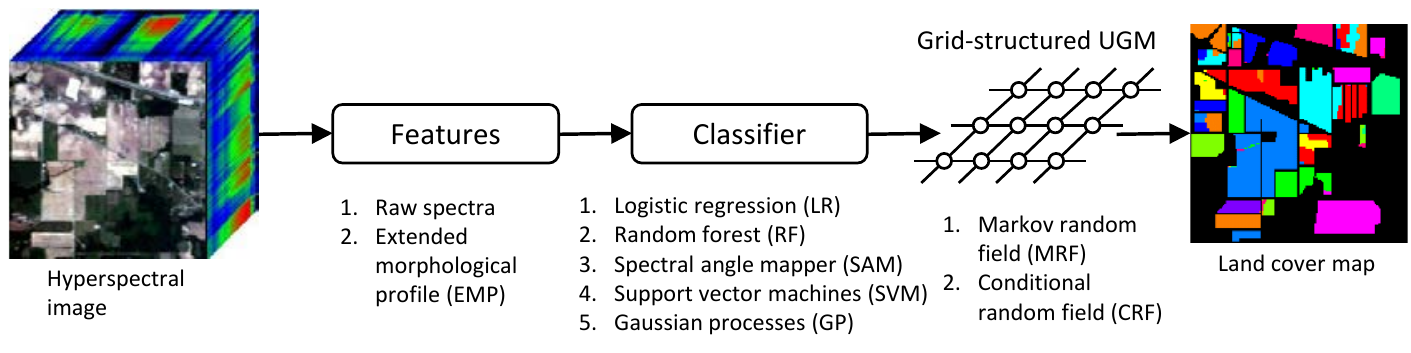}
\caption{Pipeline for testing grid-structured models.}
\label{fig:pipeline_grid}
\end{figure}

The land cover maps were generated using methods consisting of feature extraction, pixel-wise classification, and the MRF or the CRF, one after another. We experiment with two features and various pixel-wise classifiers in order to perform a comprehensive analysis. The workflow used for the experimentation is shown in Figure~\ref{fig:pipeline_grid}.


\begin{table*}[!htb]
\tiny
\caption{Performance on the Indian Pines dataset measured in overall accuracy.}
\label{tab:OATable_pines}
\centering	
\tabcolsep=0.09cm
\noindent\makebox[\textwidth]{%
\begin{tabular}{@{\extracolsep{4pt}}lrrrrrrrrrrrr@{}}
\toprule
       & \multicolumn{12}{c}{Number of training pixels per class}\\
\cline{2-13}
       & \multicolumn{3}{c}{20} & \multicolumn{3}{c}{60} & \multicolumn{3}{c}{100} & \multicolumn{3}{c}{140}\\
\cline{2-4}
\cline{5-7}
\cline{8-10}
\cline{11-13}
Methods & Best & Mean & SD & Best & Mean & SD & Best & Mean & SD & Best & Mean & SD\\
\midrule
LR & 67.17 & 61.98 & 2.99 & 76.33 & 72.53 & 1.67 & 80.00 & 76.76 & 1.60 & 82.33 & 79.49 & 1.42 \\
LR-MRF & 86.00 & 78.48 & 3.87 & 92.67 & 87.64 & 2.78 & 94.00 & 89.26 & 2.25 & 94.00 & 91.49 & 1.51 \\
LR-CRF & 72.67 & 50.19 & 13.83 & 93.33 & 86.60 & 5.34 & 96.17 & 91.16 & 4.40 & 97.33 & 93.80 & 2.24 \\
RF & 65.33 & 60.88 & 2.44 & 74.33 & 71.92 & 1.44 & 79.00 & 75.57 & 1.77 & 81.00 & 78.02 & 1.73 \\
RF-MRF & 81.33 & 75.68 & 3.64 & 90.17 & 86.17 & 2.32 & 93.00 & 89.76 & 1.73 & 95.33 & 91.92 & 1.12 \\
RF-CRF & 68.17 & 45.77 & 17.78 & 94.33 & 83.52 & 15.41 & 96.83 & 90.43 & 9.15 & 98.00 & 93.94 & 3.52 \\
SAM & 61.50 & 56.97 & 1.94 & 67.50 & 64.31 & 2.12 & 71.00 & 66.93 & 2.07 & 71.00 & 68.53 & 1.08 \\
SAM-MRF & 84.67 & 77.42 & 3.17 & 94.00 & 90.48 & 1.66 & 96.50 & 93.94 & 1.56 & 98.33 & 95.67 & 1.04 \\
SAM-CRF & 57.67 & 44.65 & 7.52 & 79.83 & 65.02 & 9.86 & 83.67 & 70.76 & 7.60 & 84.50 & 73.85 & 6.87 \\
SVM(SE) & 71.67 & 65.31 & 4.52 & 80.33 & 77.37 & 2.21 & 84.83 & 82.14 & 1.43 & 87.17 & 84.43 & 1.46 \\
SVM(SE)-MRF & 86.67 & 80.42 & 5.27 & 93.17 & 89.33 & 1.85 & 96.67 & 91.99 & 2.24 & 96.50 & 93.81 & 1.44 \\
SVM(SE)-CRF & 71.83 & 51.95 & 12.05 & 95.17 & 89.84 & 4.28 & 97.00 & 91.04 & 15.79 & 98.83 & 96.03 & 1.22 \\
SVM(ESAM) & 59.83 & 52.96 & 3.48 & 70.33 & 67.03 & 1.92 & 74.83 & 71.85 & 1.52 & 78.00 & 74.72 & 1.71 \\
SVM(ESAM)-MRF & 74.50 & 65.16 & 6.17 & 83.00 & 79.96 & 2.28 & 85.50 & 82.43 & 1.53 & 87.67 & 84.42 & 1.68 \\
SVM(ESAM)-CRF & 65.33 & 49.12 & 8.94 & 87.17 & 77.53 & 13.81 & 91.83 & 83.31 & 14.47 & 94.50 & 88.77 & 3.64 \\
GP(SE) & 64.83 & 59.47 & 2.30 & 77.50 & 75.48 & 1.06 & 83.67 & 80.81 & 1.56 & 86.50 & 83.47 & 1.63 \\
GP(SE)-MRF & 83.00 & 74.68 & 4.72 & 92.50 & 88.76 & 2.12 & 95.33 & 91.59 & 2.18 & 96.00 & 93.33 & 1.39 \\
GP(SE)-CRF & 64.67 & 44.86 & 11.03 & 86.83 & 76.16 & 13.99 & 93.83 & 76.94 & 27.73 & 96.83 & 83.23 & 25.62 \\
GP(ESAM) & 59.50 & 55.47 & 2.45 & 73.33 & 69.27 & 1.85 & 76.67 & 73.89 & 1.54 & 79.67 & 77.07 & 1.52 \\
GP(ESAM)-MRF & 83.00 & 73.17 & 3.80 & 89.33 & 85.58 & 1.70 & 91.50 & 88.24 & 2.05 & 94.17 & 90.48 & 1.80 \\
GP(ESAM)-CRF & 58.83 & 41.06 & 8.06 & 85.50 & 74.03 & 4.54 & 90.00 & 75.02 & 23.04 & 92.33 & 81.04 & 24.74 \\
EMP-LR & 93.00 & 89.09 & 2.41 & 97.83 & 95.17 & 1.28 & 98.67 & 96.84 & 1.24 & 98.83 & 97.70 & 0.59 \\
EMP-LR-MRF & 93.50 & 89.56 & 2.35 & \textbf{98.33} & 95.48 & 1.52 & \textbf{99.50} & 97.26 & 1.19 & 99.17 & 98.11 & 0.58 \\
EMP-LR-CRF & 79.67 & 56.78 & 13.07 & 98.17 & 95.72 & 1.11 & \textbf{99.50} & 97.49 & 1.00 & 99.50 & 98.37 & 0.58 \\
EMP-RF & 94.67 & \textbf{90.94} & 2.41 & \textbf{98.33} & 96.56 & 0.86 & \textbf{99.50} & 98.30 & 0.60 & \textbf{100.00} & 98.78 & 0.57 \\
EMP-RF-MRF & 94.50 & 90.84 & 2.53 & \textbf{98.33} & \textbf{96.67} & 0.80 & \textbf{99.50} & \textbf{98.40} & 0.61 & \textbf{100.00} & 98.80 & 0.58 \\
EMP-RF-CRF & 73.67 & 54.14 & 15.84 & 97.33 & 94.98 & 1.68 & 99.33 & 97.28 & 1.19 & 99.67 & 98.35 & 0.66 \\
EMP-SAM & 92.33 & 88.72 & 2.15 & 96.50 & 94.89 & 0.94 & 99.00 & 97.22 & 0.86 & 99.00 & 97.87 & 0.58 \\
EMP-SAM-MRF & \textbf{96.17} & 90.43 & 2.91 & \textbf{98.33} & 96.63 & 0.89 & \textbf{99.50} & 98.24 & 0.72 & 99.67 & 98.78 & 0.50 \\
EMP-SAM-CRF & 59.67 & 42.92 & 11.13 & 85.00 & 63.16 & 12.25 & 89.50 & 66.03 & 15.17 & 88.00 & 72.47 & 9.72 \\
EMP-SVM(SE) & 93.50 & 89.80 & 1.92 & 98.00 & 95.61 & 1.19 & 99.33 & 97.87 & 0.86 & 99.33 & 98.32 & 0.55 \\
EMP-SVM(SE)-MRF & 94.50 & 90.56 & 1.80 & 98.00 & 96.28 & 1.15 & 99.33 & 98.27 & 0.73 & 99.67 & 98.71 & 0.56 \\
EMP-SVM(SE)-CRF & 75.00 & 57.07 & 13.17 & 97.67 & 96.39 & 1.31 & \textbf{99.50} & 98.33 & 0.68 & 99.67 & \textbf{98.81} & 0.52 \\
EMP-SVM(ESAM) & 85.33 & 81.57 & 2.29 & 95.83 & 90.80 & 1.67 & 96.17 & 94.22 & 0.97 & 96.67 & 95.28 & 0.96 \\
EMP-SVM(ESAM)-MRF & 86.50 & 82.27 & 2.16 & 95.83 & 91.34 & 1.63 & 97.00 & 94.66 & 1.00 & 97.00 & 95.69 & 0.96 \\
EMP-SVM(ESAM)-CRF & 69.33 & 54.58 & 8.87 & 93.67 & 90.05 & 3.09 & 97.00 & 94.78 & 1.95 & 98.17 & 95.86 & 1.00 \\
EMP-GP(SE) & 91.00 & 87.14 & 1.94 & 96.67 & 94.54 & 1.12 & 98.17 & 96.83 & 0.76 & 98.67 & 97.54 & 0.64 \\
EMP-GP(SE)-MRF & 94.50 & 87.87 & 2.47 & 97.67 & 95.30 & 1.26 & 98.83 & 97.22 & 0.78 & 99.33 & 97.94 & 0.69 \\
EMP-GP(SE)-CRF & 71.67 & 51.86 & 11.33 & 93.83 & 86.55 & 15.29 & 96.17 & 89.80 & 15.70 & 98.00 & 91.82 & 15.93 \\
EMP-GP(ESAM) & 88.83 & 85.42 & 2.13 & 95.67 & 92.79 & 1.33 & 97.67 & 95.71 & 1.01 & 98.17 & 96.67 & 0.77 \\
EMP-GP(ESAM)-MRF & 89.67 & 86.14 & 2.20 & 96.00 & 93.44 & 1.57 & 98.17 & 96.21 & 0.92 & 98.67 & 97.06 & 0.77 \\
EMP-GP(ESAM)-CRF & 69.33 & 50.69 & 9.68 & 92.67 & 78.47 & 24.14 & 96.50 & 84.61 & 25.96 & 97.67 & 91.77 & 15.82 \\
\bottomrule
\end{tabular}
}
\end{table*}
\begin{table*}[!htb]
\tiny
\caption{Performance on the University of Pavia dataset measured in overall accuracy.}
\label{tab:OATable_paviaU}
\centering	
\tabcolsep=0.09cm
\noindent\makebox[\textwidth]{%
\begin{tabular}{@{\extracolsep{3pt}}lrrrrrrrrrrrr@{}}
\toprule
      & \multicolumn{12}{c}{Number of training pixels per class}\\
\cline{2-13}
       & \multicolumn{3}{c}{20} & \multicolumn{3}{c}{60} & \multicolumn{3}{c}{100} & \multicolumn{3}{c}{140}\\
\cline{2-4}
\cline{5-7}
\cline{8-10}
\cline{11-13}
Methods & Best & Mean & SD & Best & Mean & SD & Best & Mean & SD & Best & Mean & SD\\
\midrule
LR & 78.89 & 73.13 & 5.11 & 83.78 & 78.99 & 2.59 & 84.00 & 80.69 & 2.24 & 84.67 & 81.73 & 1.49 \\
LR-MRF & 93.11 & 82.84 & 7.06 & 92.67 & 88.99 & 2.91 & 94.67 & 90.26 & 2.15 & 93.11 & 90.64 & 1.49 \\
LR-CRF & 42.67 & 21.34 & 9.97 & 79.56 & 68.51 & 7.44 & 92.22 & 77.83 & 18.97 & 93.56 & 77.22 & 23.01 \\
RF & 80.22 & 75.41 & 2.71 & 83.33 & 80.21 & 1.55 & 86.44 & 83.15 & 1.61 & 86.67 & 84.24 & 1.58 \\
RF-MRF & 91.33 & 81.92 & 3.93 & 93.78 & 90.57 & 1.95 & 97.78 & 93.39 & 2.01 & 97.78 & 94.47 & 1.67 \\
RF-CRF & 44.00 & 21.84 & 10.48 & 88.67 & 71.62 & 6.93 & 94.89 & 72.00 & 31.33 & 97.11 & 79.35 & 27.57 \\
SAM & 79.56 & 74.59 & 2.76 & 81.33 & 78.26 & 1.79 & 83.78 & 79.56 & 2.12 & 85.11 & 80.88 & 1.98 \\
SAM-MRF & 90.22 & 83.06 & 4.80 & 94.00 & 91.04 & 1.82 & 97.11 & 93.77 & 1.29 & 97.78 & 94.99 & 1.36 \\
SAM-CRF & 43.11 & 21.30 & 9.79 & 69.56 & 58.99 & 6.22 & 84.22 & 68.69 & 7.01 & 88.67 & 72.33 & 5.77 \\
SVM(SE) & 86.22 & 77.59 & 3.15 & 90.22 & 85.84 & 2.64 & 92.67 & 88.96 & 1.49 & 91.33 & 90.02 & 0.95 \\
SVM(SE)-MRF & 96.89 & 84.40 & 4.95 & 97.11 & 93.29 & 2.65 & 98.67 & 95.90 & 1.23 & 97.78 & 95.92 & 1.40 \\
SVM(SE)-CRF & 41.33 & 21.07 & 9.34 & 94.44 & 73.64 & 7.75 & 98.67 & 76.19 & 30.23 & 99.33 & 89.24 & 21.61 \\
SVM(ESAM) & 79.11 & 73.22 & 4.08 & 81.11 & 79.01 & 1.36 & 84.44 & 81.40 & 1.79 & 86.89 & 82.85 & 1.92 \\
SVM(ESAM)-MRF & 88.22 & 78.01 & 5.96 & 91.11 & 85.13 & 2.27 & 93.56 & 87.99 & 2.46 & 94.22 & 90.00 & 2.13 \\
SVM(ESAM)-CRF & 42.67 & 21.19 & 9.44 & 88.67 & 68.13 & 9.18 & 90.22 & 70.35 & 27.73 & 95.56 & 86.93 & 15.16 \\
GP(SE) & 82.22 & 77.20 & 2.62 & 90.22 & 85.80 & 1.98 & 92.44 & 88.89 & 1.45 & 93.56 & 90.07 & 1.58 \\
GP(SE)-MRF & 91.78 & 85.53 & 3.91 & 97.11 & 94.68 & 1.34 & 98.22 & 96.37 & 1.09 & 98.89 & 96.73 & 1.07 \\
GP(SE)-CRF & 39.56 & 20.69 & 8.89 & 89.11 & 67.82 & 9.98 & 98.67 & 84.59 & 8.04 & 97.56 & 85.83 & 15.96 \\
GP(ESAM) & 80.44 & 76.33 & 2.46 & 83.56 & 80.87 & 1.43 & 86.22 & 83.95 & 1.46 & 88.89 & 85.48 & 1.89 \\
GP(ESAM)-MRF & 90.44 & 84.24 & 4.00 & 94.22 & 90.49 & 1.75 & 97.33 & 93.85 & 1.59 & 97.78 & 95.39 & 1.25 \\
GP(ESAM)-CRF & 39.78 & 20.58 & 9.07 & 84.67 & 68.71 & 9.31 & 95.78 & 85.61 & 6.88 & 96.00 & 81.67 & 24.36 \\
EMP-LR & 93.56 & 89.80 & 3.09 & 98.44 & 96.59 & 1.09 & 98.89 & 97.46 & 1.04 & \textbf{99.78} & 98.14 & 0.99 \\
EMP-LR-MRF & 93.78 & 89.87 & 3.18 & 98.44 & 96.77 & 1.07 & 98.89 & 97.61 & 1.02 & \textbf{99.78} & 98.33 & 0.79 \\
EMP-LR-CRF & 42.89 & 22.10 & 10.35 & 97.11 & 74.96 & 14.58 & 98.67 & 77.96 & 30.89 & 99.33 & 92.79 & 15.98 \\
EMP-RF & \textbf{99.33} & 95.73 & 2.60 & \textbf{99.56} & 98.50 & 0.62 & \textbf{100.00} & 99.06 & 0.59 & \textbf{99.78} & \textbf{99.24} & 0.43 \\
EMP-RF-MRF & \textbf{99.33} & \textbf{95.74} & 2.61 & \textbf{99.56} & \textbf{98.51} & 0.61 & \textbf{100.00} & \textbf{99.09} & 0.59 & \textbf{99.78} & \textbf{99.24} & 0.43 \\
EMP-RF-CRF & 44.22 & 22.41 & 10.89 & 96.89 & 77.84 & 8.41 & 99.11 & 86.62 & 21.32 & 99.33 & 96.22 & 4.30 \\
EMP-SAM & 94.44 & 90.16 & 2.53 & 98.22 & 95.81 & 1.07 & 99.11 & 97.56 & 0.92 & \textbf{99.78} & 98.09 & 0.68 \\
EMP-SAM-MRF & 97.56 & 91.08 & 2.83 & 98.44 & 96.50 & 1.13 & 99.33 & 97.93 & 0.84 & \textbf{99.78} & 98.41 & 0.82 \\
EMP-SAM-CRF & 40.00 & 19.83 & 8.38 & 76.89 & 54.59 & 12.48 & 92.44 & 60.71 & 12.69 & 89.33 & 61.59 & 15.13 \\
EMP-SVM(SE) & 95.33 & 90.13 & 3.22 & 99.11 & 96.59 & 1.70 & 99.33 & 97.87 & 0.80 & 99.56 & 98.50 & 0.70 \\
EMP-SVM(SE)-MRF & 95.56 & 90.16 & 3.30 & 99.11 & 96.73 & 1.61 & 99.11 & 97.90 & 0.82 & \textbf{99.78} & 98.61 & 0.72 \\
EMP-SVM(SE)-CRF & 41.33 & 21.46 & 9.81 & 86.67 & 73.39 & 13.57 & 98.44 & 88.29 & 15.79 & \textbf{99.78} & 89.92 & 21.84 \\
EMP-SVM(ESAM) & 93.11 & 89.02 & 2.54 & 98.44 & 96.46 & 1.16 & 99.11 & 97.51 & 0.93 & 98.89 & 98.07 & 0.66 \\
EMP-SVM(ESAM)-MRF & 94.22 & 89.27 & 2.67 & 98.44 & 96.48 & 1.16 & 99.11 & 97.64 & 0.75 & 99.33 & 98.11 & 0.70 \\
EMP-SVM(ESAM)-CRF & 44.00 & 21.48 & 10.15 & 94.67 & 73.11 & 14.08 & 98.89 & 77.41 & 30.71 & 98.67 & 89.53 & 21.74 \\
EMP-GP(SE) & 94.44 & 90.75 & 2.22 & 98.67 & 97.06 & 0.90 & 99.33 & 98.10 & 0.90 & \textbf{99.78} & 98.50 & 0.83 \\
EMP-GP(SE)-MRF & 94.22 & 90.78 & 2.23 & 98.89 & 97.13 & 0.99 & 99.33 & 98.18 & 0.84 & 99.56 & 98.50 & 0.79 \\
EMP-GP(SE)-CRF & 40.67 & 20.04 & 8.92 & 91.33 & 70.01 & 9.18 & 96.00 & 84.30 & 6.50 & 96.44 & 83.52 & 21.03 \\
EMP-GP(ESAM) & 94.89 & 89.97 & 2.38 & 98.00 & 96.15 & 1.87 & 99.78 & 97.77 & 0.89 & 99.33 & 98.35 & 0.75 \\
EMP-GP(ESAM)-MRF & 95.11 & 90.25 & 2.46 & 98.89 & 96.46 & 1.27 & 99.78 & 97.78 & 0.90 & 99.33 & 98.38 & 0.77 \\
EMP-GP(ESAM)-CRF & 39.11 & 20.07 & 8.81 & 82.44 & 69.16 & 7.28 & 94.22 & 84.75 & 5.91 & 96.67 & 80.86 & 24.39 \\
\bottomrule
\end{tabular}
}
\end{table*}
\begin{table*}[!htb]
\tiny
\caption{Performance on the Pavia Center dataset measured in overall accuracy.}
\label{tab:OATable_paviaC}
\centering	
\tabcolsep=0.09cm
\noindent\makebox[\textwidth]{%
\begin{tabular}{@{\extracolsep{3pt}}lrrrrrrrrrrrr@{}}
\toprule
         & \multicolumn{12}{c}{Number of training pixels per class}\\
\cline{2-13}
       & \multicolumn{3}{c}{20} & \multicolumn{3}{c}{60} & \multicolumn{3}{c}{100} & \multicolumn{3}{c}{140}\\
\cline{2-4}
\cline{5-7}
\cline{8-10}
\cline{11-13}
    Methods   & Best & Mean & SD & Best & Mean & SD & Best & Mean & SD & Best & Mean & SD\\
\midrule
LR & 90.22 & 86.13 & 2.70 & 94.22 & 90.30 & 1.82 & 94.22 & 91.75 & 1.42 & 94.67 & 92.92 & 1.22 \\
LR-MRF & 95.11 & 91.24 & 3.13 & 97.78 & 94.64 & 2.05 & 98.89 & 95.88 & 1.61 & 98.67 & 96.78 & 0.92 \\
LR-CRF & 33.33 & 13.37 & 6.13 & 85.11 & 51.61 & 12.35 & 96.67 & 77.68 & 9.19 & 97.78 & 82.84 & 8.32 \\
RF & 91.56 & 85.24 & 2.26 & 93.56 & 89.79 & 1.61 & 94.67 & 91.85 & 1.50 & 95.11 & 92.16 & 1.15 \\
RF-MRF & 94.44 & 89.81 & 2.67 & 96.89 & 94.00 & 1.77 & 97.33 & 95.09 & 1.36 & 97.78 & 95.76 & 1.04 \\
RF-CRF & 33.11 & 13.54 & 6.53 & 82.44 & 52.37 & 12.19 & 96.67 & 78.00 & 8.92 & 96.00 & 80.94 & 15.33 \\
SAM & 90.89 & 86.35 & 2.05 & 92.22 & 89.44 & 1.57 & 93.56 & 90.16 & 1.65 & 93.33 & 90.93 & 1.07 \\
SAM-MRF & 97.56 & 93.37 & 2.33 & 98.22 & 95.78 & 1.29 & 98.44 & 95.79 & 1.48 & 97.78 & 96.28 & 0.98 \\
SAM-CRF & 33.11 & 13.19 & 5.91 & 70.00 & 47.62 & 10.37 & 82.22 & 65.60 & 9.69 & 89.11 & 69.69 & 8.80 \\
SVM(SE) & 93.56 & 88.19 & 2.87 & 94.22 & 91.99 & 1.43 & 96.44 & 93.66 & 1.71 & 96.22 & 94.39 & 1.13 \\
SVM(SE)-MRF & 96.67 & 91.57 & 3.10 & 97.78 & 95.28 & 1.41 & 98.67 & 96.46 & 1.40 & 98.44 & 97.09 & 0.77 \\
SVM(SE)-CRF & 32.67 & 13.53 & 6.55 & 82.67 & 51.67 & 11.63 & 98.67 & 78.48 & 9.05 & 99.56 & 84.03 & 8.20 \\
SVM(ESAM) & 91.56 & 88.32 & 1.79 & 93.78 & 90.63 & 1.39 & 94.00 & 91.67 & 1.36 & 93.56 & 91.76 & 0.97 \\
SVM(ESAM)-MRF & 95.78 & 91.83 & 2.62 & 96.00 & 93.85 & 1.37 & 97.11 & 94.71 & 1.12 & 96.22 & 94.77 & 1.12 \\
SVM(ESAM)-CRF & 33.11 & 13.39 & 6.14 & 84.44 & 52.04 & 12.19 & 95.33 & 77.17 & 8.78 & 96.89 & 83.44 & 7.40 \\
GP(SE) & 94.00 & 89.24 & 2.09 & 94.89 & 92.35 & 1.48 & 95.78 & 93.63 & 1.36 & 96.44 & 94.59 & 1.08 \\
GP(SE)-MRF & \textbf{97.78} & 92.83 & 2.52 & 98.44 & 95.72 & 1.52 & 98.67 & 96.81 & 1.13 & 98.89 & 97.07 & 0.94 \\
GP(SE)-CRF & 32.67 & 13.10 & 5.46 & 80.44 & 50.04 & 12.03 & 96.67 & 75.24 & 8.88 & 97.56 & 81.61 & 7.40 \\
GP(ESAM) & 92.67 & 88.73 & 1.90 & 93.78 & 91.30 & 1.24 & 95.33 & 92.27 & 1.56 & 94.67 & 92.79 & 1.09 \\
GP(ESAM)-MRF & 96.44 & 92.91 & 2.44 & 97.56 & 95.49 & 1.05 & 98.00 & 96.08 & 1.03 & 98.00 & 96.24 & 1.08 \\
GP(ESAM)-CRF & 32.89 & 13.26 & 5.83 & 81.56 & 49.64 & 10.65 & 96.00 & 75.68 & 8.60 & 96.00 & 80.81 & 7.82 \\
EMP-LR & 92.44 & 88.40 & 3.17 & 97.33 & 95.24 & 1.28 & 98.89 & 96.96 & 1.04 & 99.33 & 97.33 & 0.99 \\
EMP-LR-MRF & 94.00 & 88.75 & 3.24 & 98.44 & 95.71 & 1.20 & 98.89 & 97.11 & 1.09 & 99.33 & 97.54 & 0.94 \\
EMP-LR-CRF & 33.11 & 13.33 & 6.02 & 84.00 & 53.02 & 12.57 & 96.00 & 78.81 & 9.32 & 98.44 & 80.83 & 20.10 \\
EMP-RF & 97.11 & 93.99 & 1.92 & \textbf{99.11} & 97.28 & 1.14 & \textbf{99.56} & 98.31 & 0.93 & 99.56 & 98.67 & 0.73 \\
EMP-RF-MRF & 97.33 & \textbf{94.09} & 1.94 & \textbf{99.11} & \textbf{97.30} & 1.17 & \textbf{99.56} & \textbf{98.33} & 0.87 & 99.56 & \textbf{98.73} & 0.73 \\
EMP-RF-CRF & 33.11 & 13.49 & 6.39 & 86.22 & 53.29 & 12.46 & 96.44 & 78.85 & 9.18 & 98.44 & 82.87 & 15.62 \\
EMP-SAM & 93.33 & 90.41 & 1.58 & 97.56 & 95.67 & 1.20 & 98.89 & 97.14 & 1.11 & 99.11 & 97.84 & 0.75 \\
EMP-SAM-MRF & 95.11 & 91.39 & 1.96 & 97.56 & 96.17 & 1.02 & 99.33 & 97.72 & 0.81 & 99.33 & 98.03 & 0.76 \\
EMP-SAM-CRF & 31.78 & 13.16 & 5.59 & 60.44 & 45.31 & 8.59 & 78.22 & 61.77 & 8.12 & 80.44 & 63.37 & 11.31 \\
EMP-SVM(SE) & 96.67 & 91.89 & 2.39 & 98.44 & 96.53 & 1.13 & \textbf{99.56} & 98.10 & 0.92 & \textbf{99.78} & 98.47 & 1.05 \\
EMP-SVM(SE)-MRF & 96.67 & 91.98 & 2.37 & 98.44 & 96.58 & 1.16 & \textbf{99.56} & 98.20 & 0.90 & \textbf{99.78} & 98.49 & 1.05 \\
EMP-SVM(SE)-CRF & 32.00 & 13.52 & 6.43 & 85.56 & 53.56 & 12.41 & 97.33 & 79.01 & 9.64 & 99.33 & 78.97 & 23.98 \\
EMP-SVM(ESAM) & 93.33 & 90.36 & 2.05 & 97.56 & 94.87 & 1.27 & 98.44 & 96.73 & 1.01 & 99.11 & 97.10 & 0.88 \\
EMP-SVM(ESAM)-MRF & 93.56 & 90.42 & 2.00 & 97.56 & 94.83 & 1.28 & 98.44 & 96.71 & 1.12 & 98.89 & 97.24 & 0.84 \\
EMP-SVM(ESAM)-CRF & 32.22 & 13.27 & 5.83 & 84.00 & 51.61 & 12.24 & 97.33 & 75.64 & 14.93 & 97.11 & 82.35 & 15.15 \\
EMP-GP(SE) & 94.67 & 90.81 & 2.37 & 98.67 & 96.68 & 0.98 & 99.11 & 97.95 & 0.89 & 99.56 & 98.21 & 0.75 \\
EMP-GP(SE)-MRF & 95.11 & 91.18 & 2.40 & 98.67 & 96.73 & 0.99 & 99.11 & 98.01 & 0.85 & 99.56 & 98.29 & 0.77 \\
EMP-GP(SE)-CRF & 29.33 & 13.02 & 5.13 & 79.33 & 49.99 & 11.26 & 95.33 & 74.41 & 8.50 & 95.78 & 80.31 & 7.96 \\
EMP-GP(ESAM) & 93.56 & 90.05 & 2.17 & 97.78 & 95.44 & 1.18 & 98.89 & 97.25 & 1.02 & 98.89 & 97.67 & 0.84 \\
EMP-GP(ESAM)-MRF & 93.56 & 90.10 & 2.17 & 98.00 & 95.61 & 1.14 & 98.89 & 97.34 & 1.06 & 98.89 & 97.70 & 0.78 \\
EMP-GP(ESAM)-CRF & 28.22 & 12.93 & 4.87 & 68.67 & 48.16 & 10.15 & 96.22 & 72.67 & 8.88 & 94.67 & 80.07 & 6.93 \\
\bottomrule
\end{tabular}
}
\end{table*}

\begin{table*}[!htb]
\tiny
\caption{Performance on the Salinas dataset measured in overall accuracy.}
\label{tab:OATable_salinas}
\centering	
\tabcolsep=0.09cm
\noindent\makebox[\textwidth]{%
\begin{tabular}{@{\extracolsep{3pt}}lrrrrrrrrrrrr@{}}
\toprule
       & \multicolumn{12}{c}{Number of training pixels per class}\\
\cline{2-13}
       & \multicolumn{3}{c}{20} & \multicolumn{3}{c}{60} & \multicolumn{3}{c}{100} & \multicolumn{3}{c}{140}\\
\cline{2-4}
\cline{5-7}
\cline{8-10}
\cline{11-13}
    Methods & Best & Mean & SD & Best & Mean & SD & Best & Mean & SD & Best & Mean & SD\\
\midrule
LR & 93.13 & 90.16 & 1.81 & 95.75 & 93.45 & 1.03 & 96.13 & 94.71 & 0.86 & 96.13 & 95.08 & 0.73 \\
LR-MRF & 96.50 & 94.13 & 2.24 & 99.25 & 97.46 & 1.48 & 99.00 & 97.90 & 0.80 & 98.88 & 97.94 & 0.63 \\
LR-CRF & 49.88 & 23.98 & 8.50 & 96.88 & 81.18 & 15.93 & 98.63 & 94.70 & 2.48 & 98.50 & 96.84 & 1.49 \\
RF & 91.50 & 88.88 & 1.49 & 94.63 & 92.02 & 1.11 & 95.63 & 93.38 & 1.02 & 94.88 & 93.64 & 0.77 \\
RF-MRF & 99.00 & 95.60 & 2.17 & 99.13 & 98.09 & 1.13 & 99.38 & 98.47 & 0.65 & 99.88 & 98.62 & 0.68 \\
RF-CRF & 49.75 & 24.36 & 8.49 & 97.63 & 81.19 & 15.67 & 98.50 & 94.53 & 2.57 & 99.13 & 97.17 & 1.88 \\
SAM & 90.63 & 88.72 & 1.25 & 92.63 & 90.91 & 0.92 & 93.75 & 91.95 & 0.95 & 94.13 & 91.97 & 0.98 \\
SAM-MRF & 98.00 & 94.04 & 2.02 & 99.38 & 97.63 & 1.29 & 99.25 & 98.38 & 0.63 & 99.63 & 98.70 & 0.45 \\
SAM-CRF & 47.75 & 22.93 & 8.09 & 83.13 & 70.20 & 8.31 & 89.38 & 77.52 & 6.33 & 92.38 & 80.20 & 6.52 \\
SVM(SE) & 93.50 & 89.96 & 2.04 & 95.75 & 93.62 & 1.03 & 96.38 & 95.00 & 0.76 & 96.50 & 95.13 & 0.67 \\
SVM(SE)-MRF & 98.63 & 93.72 & 3.00 & 99.38 & 97.66 & 1.76 & 99.50 & 98.64 & 0.55 & 99.88 & 98.86 & 0.60 \\
SVM(SE)-CRF & 49.63 & 24.33 & 8.50 & 96.38 & 81.43 & 15.92 & 99.13 & 94.60 & 3.06 & 99.50 & 97.52 & 1.53 \\
SVM(ESAM) & 90.25 & 84.37 & 3.86 & 93.63 & 91.70 & 1.32 & 94.63 & 92.91 & 0.90 & 95.25 & 93.21 & 0.88 \\
SVM(ESAM)-MRF & 97.00 & 89.63 & 3.51 & 98.50 & 96.17 & 1.84 & 99.13 & 97.48 & 1.07 & 99.50 & 98.25 & 0.60 \\
SVM(ESAM)-CRF & 49.75 & 24.12 & 8.55 & 96.25 & 78.46 & 20.70 & 96.63 & 92.83 & 2.56 & 98.25 & 95.27 & 2.28 \\
GP(SE) & 92.88 & 90.10 & 1.33 & 95.13 & 92.79 & 1.11 & 95.88 & 94.35 & 0.92 & 96.38 & 94.65 & 0.72 \\
GP(SE)-MRF & 97.25 & 92.75 & 2.56 & 99.00 & 95.98 & 2.22 & 98.50 & 96.87 & 1.61 & 99.13 & 97.59 & 1.17 \\
GP(SE)-CRF & 43.25 & 22.05 & 7.68 & 93.38 & 77.75 & 10.39 & 96.13 & 86.53 & 15.92 & 97.75 & 92.54 & 4.40 \\
GP(ESAM) & 92.00 & 89.72 & 1.16 & 93.75 & 92.42 & 0.77 & 95.00 & 93.50 & 0.83 & 95.50 & 93.93 & 0.73 \\
GP(ESAM)-MRF & 96.25 & 92.33 & 1.89 & 98.88 & 96.45 & 2.11 & 98.88 & 97.64 & 0.89 & 98.75 & 98.00 & 0.53 \\
GP(ESAM)-CRF & 39.38 & 20.65 & 7.46 & 95.25 & 71.65 & 20.82 & 97.38 & 84.79 & 21.68 & 98.00 & 93.92 & 3.24 \\
EMP-LR & 98.75 & 96.33 & 1.37 & 99.50 & 98.45 & 0.62 & 99.88 & 98.98 & 0.45 & 99.88 & 99.11 & 0.47 \\
EMP-LR-MRF & 98.75 & 96.68 & 1.25 & 99.50 & 98.74 & 0.47 & 99.88 & 99.12 & 0.46 & \textbf{100.00} & 99.35 & 0.38 \\
EMP-LR-CRF & 49.50 & 24.43 & 8.32 & 99.00 & 79.71 & 21.28 & 99.63 & 96.29 & 3.13 & 99.88 & 98.96 & 1.70 \\
EMP-RF & \textbf{99.63} & 98.35 & 1.03 & 99.88 & 99.19 & 0.43 & \textbf{100.00} & 99.45 & 0.28 & \textbf{100.00} & 99.64 & 0.25 \\
EMP-RF-MRF & \textbf{99.63} & \textbf{98.43} & 1.01 & 99.88 & \textbf{99.30} & 0.39 & \textbf{100.00} & \textbf{99.50} & 0.26 & \textbf{100.00} & \textbf{99.67} & 0.24 \\
EMP-RF-CRF & 49.63 & 24.38 & 8.65 & \textbf{100.00} & 81.98 & 15.99 & 99.63 & 96.28 & 2.90 & \textbf{100.00} & 99.00 & 1.61 \\
EMP-SAM & 99.13 & 96.68 & 1.65 & 99.63 & 98.71 & 0.85 & \textbf{100.00} & 99.25 & 0.34 & 99.88 & 99.38 & 0.27 \\
EMP-SAM-MRF & 99.38 & 97.17 & 1.50 & 99.75 & 99.03 & 0.55 & \textbf{100.00} & 99.38 & 0.31 & \textbf{100.00} & 99.51 & 0.25 \\
EMP-SAM-CRF & 45.25 & 23.37 & 7.73 & 85.13 & 69.08 & 10.11 & 95.38 & 82.08 & 7.85 & 96.75 & 84.17 & 8.61 \\
EMP-SVM(SE) & 98.75 & 96.11 & 1.48 & 99.75 & 98.93 & 0.66 & \textbf{100.00} & 99.39 & 0.26 & \textbf{100.00} & 99.46 & 0.39 \\
EMP-SVM(SE)-MRF & 98.88 & 96.46 & 1.45 & 99.88 & 99.06 & 0.64 & \textbf{100.00} & 99.45 & 0.26 & \textbf{100.00} & 99.54 & 0.33 \\
EMP-SVM(SE)-CRF & 50.00 & 24.46 & 8.60 & 98.88 & 80.62 & 21.18 & 99.75 & 96.40 & 3.16 & \textbf{100.00} & 98.98 & 1.57 \\
EMP-SVM(ESAM) & 98.50 & 96.31 & 1.62 & 99.25 & 98.16 & 0.92 & 99.63 & 98.81 & 0.51 & 99.75 & 99.20 & 0.38 \\
EMP-SVM(ESAM)-MRF & 98.50 & 96.39 & 1.53 & 99.75 & 98.35 & 1.00 & 99.63 & 99.01 & 0.47 & 99.88 & 99.31 & 0.36 \\
EMP-SVM(ESAM)-CRF & 49.88 & 24.23 & 8.49 & 98.00 & 82.20 & 16.11 & 99.38 & 96.08 & 3.02 & 99.75 & 98.11 & 2.77 \\
EMP-GP(SE) & 99.25 & 96.28 & 1.64 & 99.50 & 98.73 & 0.73 & 99.75 & 99.22 & 0.45 & \textbf{100.00} & 99.53 & 0.20 \\
EMP-GP(SE)-MRF & 99.25 & 96.79 & 1.67 & 99.75 & 98.76 & 0.75 & 99.88 & 99.28 & 0.46 & \textbf{100.00} & 99.55 & 0.23 \\
EMP-GP(SE)-CRF & 45.63 & 22.83 & 8.65 & 94.25 & 78.07 & 15.91 & 99.25 & 83.57 & 26.54 & 99.63 & 94.08 & 6.38 \\
EMP-GP(ESAM) & 98.63 & 95.93 & 1.48 & 99.63 & 98.33 & 0.62 & 99.63 & 99.00 & 0.35 & 99.75 & 99.17 & 0.41 \\
EMP-GP(ESAM)-MRF & 98.50 & 96.37 & 1.45 & 99.75 & 98.46 & 0.61 & 99.75 & 99.10 & 0.35 & 99.88 & 99.27 & 0.40 \\
EMP-GP(ESAM)-CRF & 45.25 & 22.13 & 7.94 & 94.13 & 74.88 & 20.87 & 98.13 & 82.38 & 26.35 & 98.63 & 94.31 & 5.52 \\
\bottomrule
\end{tabular}
}
\end{table*}
The features used in the experiments were the raw spectra and the spatial-spectral extended morphological features~\cite{benediktsson2005classification} (EMP). The EMP features were obtained by applying the principal component analysis (PCA) on the image, retaining the relevant PCA components, and then applying a series of opening and closing morphological operators with circular structured elements of increasing size. The size of the first structuring element was 2 pixels in diameter, and the size of the subsequent ones were increased at a fixed step. The fraction of variance preserved after PCA, the number of morphological operations applied, and the increment in the size of the structuring elements are the hyperparameters of the EMP features. These hyperparameters were tuned using grid search over the validation set, which consisted of randomly selected 30\% of the training pixels of each class. The remaining 70\% was used as classifier's training data. The fraction of variance preserved was chosen from $\{84\%,89\%,94\%,99\%\}$, and the number of morphological operators and the increment in the size of morphological operators were chosen from $\{2, 4, 8\}$. The MRF's parameters were also tuned using this validation set. The CRF was learned using all the training pixels.

The classifiers used were logistic regression (LR), random forest (RF), spectral angle mapper (SAM), support vector machine (SVM), and Gaussian process (GP). These were implemented using the multivariate logistic regression with L2 regularized weights in the LIBLINEAR library~\cite{fan2008liblinear}, MATLAB's random forest, probabilistic multi-output support vector machine in the LIBSVM library~\cite{chang2011libsvm}, and the Gaussian process classifiers in the GPML library~\cite{rasmussen2010gaussian} respectively. The squared exponential (SE) and the exponential spectral mapper~\cite{gewali2016novel} (ESAM) kernel/covariance functions were used with the SVM and the GP. The classifiers were trained on 70\% of the pixels remaining in the training set. The C parameter and the kernel scale in the SVM were chosen from $\{10^{-3},10^{-2},...,10^2,10^3\}$ by training the SVM on 80\% of the classifier's training data and validating over the remaining 20\%. After validation, the SVM was trained on the entire classifier's training data using the tuned hyperparameters. Using similar grid search schemes, the number of trees in the RF was chosen from the set $\{50,100,200,400\}$, and the regularization parameter in the logistic regression was chosen from $\{10^{-3},10^{-2},...,10^2,10^3\}$. The gain of ESAM was fixed to one when using it with the SVM. 

Since, the GPML library does not provide multi-class classifiers, binary classifiers were trained using one-vs-one scheme and the multi-class probabilities were estimated by \cite{wu2004probability}. Error function likelihood was used with the GP classifier and the inference was performed using Laplace approximation. The hyper-parameters of the covariance function were learned by maximizing the marginal likelihood. When using SAM, the angle between the test pixel and the classifier's training examples of each class were calculated, and the minimum angle from each class was directly used as unary energy for MRF~\cite{gewali2016spectral}. For CRF, the feature function was obtained by passing the minimum angle through $e^{-x}$ function.  

Table~\ref{tab:OATable_pines}, Table~\ref{tab:OATable_paviaU}, Table~\ref{tab:OATable_paviaC}, and Table~\ref{tab:OATable_salinas} show the performance comparison of different methods on the datasets. A varying number of training pixels per class were randomly selected for training, and the models were tested on a separate set of randomly selected pixels. The testing set consisted of 50 pixels each from all of the classes. This process was repeated for 30 independent trials to obtain the mean, the standard deviation (SD) and the best overall accuracy (OA) over these trials. In the tables, each method's name contains three parts--the first indicates whether raw pixels or EMP features were uses, the second contains the classifiers name, and the third indicate whether MRF or CRF was applied. The names of the kernel/covariance function used with the SVMs and the GPs are included in parenthesis.

\begin{table*}[!htb]
\tiny
\caption{Performance comparison using various metrics.}
\label{tab:OATable_elaborate}
\centering	
\tabcolsep=0.09cm
\noindent\makebox[\textwidth]{%
\begin{tabular}{@{}llccccc@{}} 
\toprule
\multicolumn{7}{c}{Indian Pines} \\
\midrule
Pixels/class & Methods & OA & $\kappa$ & Avg. Precision & Avg. Recall & Avg. F1 \\
\multirow{6}{*}{20} & SVM & 65.31$\pm$4.45 & 62.16$\pm$4.85 & 66.04$\pm$3.49 & 65.31$\pm$4.45 & 64.87$\pm$4.18\\
      & SVM-MRF & 80.42$\pm$5.19 & 78.64$\pm$5.66 & 82.29$\pm$5.43 & 80.42$\pm$5.19 & 79.74$\pm$5.44\\
      & SVM-CRF & 51.95$\pm$11.84 & 47.58$\pm$12.92 & 42.26$\pm$14.13 & 51.95$\pm$11.84 & 43.83$\pm$13.28\\
      & EMP-SVM & 89.80$\pm$1.89 & 88.87$\pm$2.06 & 90.47$\pm$1.79 & 89.80$\pm$1.89 & 89.77$\pm$1.89\\
      & EMP-SVM-MRF & \textbf{90.56}$\pm$1.77 & \textbf{89.70}$\pm$1.93 & \textbf{91.30}$\pm$1.52 & \textbf{90.56}$\pm$1.77 & \textbf{90.52}$\pm$1.77\\
      & EMP-SVM-CRF & 57.07$\pm$12.95 & 53.16$\pm$14.12 & 48.14$\pm$15.39 & 57.07$\pm$12.95 & 49.47$\pm$14.27\\
\midrule
\multirow{6}{*}{200} & SVM & 85.71$\pm$1.85 & 84.13$\pm$2.06 & 85.86$\pm$1.88 & 85.71$\pm$1.85 & 85.67$\pm$1.86\\
      & SVM-MRF & 95.01$\pm$1.29 & 94.46$\pm$1.44 & 95.22$\pm$1.26 & 95.01$\pm$1.29 & 94.99$\pm$1.30\\
      & SVM-CRF & 97.10$\pm$1.05 & 96.78$\pm$1.17 & 97.28$\pm$0.98 & 97.10$\pm$1.05 & 97.09$\pm$1.06\\
      & EMP-SVM & 98.91$\pm$0.54 & 98.79$\pm$0.60 & 98.93$\pm$0.53 & 98.91$\pm$0.54 & 98.91$\pm$0.54\\
      & EMP-SVM-MRF & 99.15$\pm$0.47 & 99.05$\pm$0.52 & 99.17$\pm$0.46 & 99.15$\pm$0.47 & 99.15$\pm$0.47\\
      & EMP-SVM-CRF & \textbf{99.35}$\pm$0.43 & \textbf{99.27}$\pm$0.48 & \textbf{99.37}$\pm$0.41 & \textbf{99.35}$\pm$0.43 & \textbf{99.34}$\pm$0.43\\
\midrule
\multicolumn{7}{c}{University of Pavia} \\
\midrule
Pixels/class & Methods & OA & $\kappa$ & Avg. Precision & Avg. Recall & Avg. F1 \\
\multirow{6}{*}{20} & SVM & 77.59$\pm$3.10 & 74.78$\pm$3.49 & 78.23$\pm$2.95 & 77.59$\pm$3.10 & 77.21$\pm$3.33\\
      & SVM-MRF & 84.40$\pm$4.87 & 82.45$\pm$5.48 & 85.62$\pm$4.74 & 84.40$\pm$4.87 & 84.08$\pm$4.97\\
      & SVM-CRF & 21.07$\pm$9.18 & 11.21$\pm$10.33 & 9.06$\pm$8.24 & 21.07$\pm$9.18 & 10.73$\pm$8.34\\
      & EMP-SVM & 90.13$\pm$3.17 & 88.89$\pm$3.57 & 90.94$\pm$2.92 & 90.13$\pm$3.17 & 90.12$\pm$3.17\\
      & EMP-SVM-MRF & \textbf{90.16}$\pm$3.24 & \textbf{88.92}$\pm$3.65 & \textbf{90.97}$\pm$2.99 & \textbf{90.16}$\pm$3.24 & \textbf{90.15}$\pm$3.24\\
      & EMP-SVM-CRF & 21.46$\pm$9.65 & 11.64$\pm$10.86 & 9.32$\pm$8.05 & 21.46$\pm$9.65 & 11.27$\pm$8.69\\
\midrule
\multirow{6}{*}{200} & SVM & 91.47$\pm$1.25 & 90.40$\pm$1.41 & 91.65$\pm$1.26 & 91.47$\pm$1.25 & 91.46$\pm$1.26\\
      & SVM-MRF & 96.73$\pm$0.87 & 96.32$\pm$0.98 & 96.88$\pm$0.83 & 96.73$\pm$0.87 & 96.72$\pm$0.88\\
      & SVM-CRF & 97.40$\pm$1.09 & 97.08$\pm$1.22 & 97.62$\pm$0.93 & 97.40$\pm$1.09 & 97.41$\pm$1.07\\
      & EMP-SVM & 98.96$\pm$0.46 & 98.83$\pm$0.52 & 98.98$\pm$0.44 & 98.96$\pm$0.46 & 98.95$\pm$0.46\\
      & EMP-SVM-MRF & \textbf{99.04}$\pm$0.45 & \textbf{98.92}$\pm$0.50 & \textbf{99.06}$\pm$0.43 & \textbf{99.04}$\pm$0.45 & \textbf{99.04}$\pm$0.45\\
      & EMP-SVM-CRF & 95.89$\pm$15.75 & 95.38$\pm$17.72 & 95.61$\pm$17.53 & 95.89$\pm$15.75 & 95.59$\pm$17.35\\
\midrule
\multicolumn{7}{c}{Pavia Center} \\
\midrule
Pixels/class & Methods & OA & $\kappa$ & Avg. Precision & Avg. Recall & Avg. F1 \\
\multirow{6}{*}{20} & SVM & 88.19$\pm$2.82 & 86.71$\pm$3.18 & 88.77$\pm$2.81 & 88.19$\pm$2.82 & 88.13$\pm$2.83\\
      & SVM-MRF & 91.57$\pm$3.05 & 90.52$\pm$3.43 & 92.19$\pm$2.89 & 91.57$\pm$3.05 & 91.55$\pm$3.03\\
      & SVM-CRF & 13.53$\pm$6.44 & 2.72$\pm$7.25 & 2.65$\pm$3.75 & 13.53$\pm$6.44 & 4.05$\pm$4.84\\
      & EMP-SVM & 91.89$\pm$2.35 & 90.88$\pm$2.64 & 92.38$\pm$2.20 & 91.89$\pm$2.35 & 91.84$\pm$2.41\\
      & EMP-SVM-MRF & \textbf{91.98}$\pm$2.33 & \textbf{90.97}$\pm$2.62 & \textbf{92.46}$\pm$2.17 & \textbf{91.98}$\pm$2.33 & \textbf{91.94}$\pm$2.39\\
      & EMP-SVM-CRF & 13.52$\pm$6.32 & 2.71$\pm$7.11 & 2.70$\pm$4.09 & 13.52$\pm$6.32 & 4.06$\pm$5.00\\
\midrule
\multirow{6}{*}{200} & SVM & 94.96$\pm$1.00 & 94.33$\pm$1.13 & 95.11$\pm$0.97 & 94.96$\pm$1.00 & 94.96$\pm$1.01\\
      & SVM-MRF & 97.71$\pm$0.89 & 97.42$\pm$1.00 & 97.81$\pm$0.82 & 97.71$\pm$0.89 & 97.71$\pm$0.90\\
      & SVM-CRF & 82.95$\pm$24.83 & 80.82$\pm$27.93 & 79.72$\pm$27.85 & 82.95$\pm$24.83 & 80.32$\pm$27.41\\
      & EMP-SVM & 98.61$\pm$0.73 & 98.43$\pm$0.83 & 98.64$\pm$0.71 & 98.61$\pm$0.73 & 98.61$\pm$0.74\\
      & EMP-SVM-MRF & \textbf{98.75}$\pm$0.71 & \textbf{98.59}$\pm$0.79 & \textbf{98.78}$\pm$0.68 & \textbf{98.75}$\pm$0.71 & \textbf{98.75}$\pm$0.71\\
      & EMP-SVM-CRF & 78.99$\pm$30.93 & 76.37$\pm$34.80 & 75.39$\pm$34.27 & 78.99$\pm$30.93 & 76.09$\pm$33.94\\
\midrule
\multicolumn{7}{c}{Salinas} \\
\midrule
Pixels/class & Methods & OA & $\kappa$ & Avg. Precision & Avg. Recall & Avg. F1 \\
\multirow{6}{*}{20} & SVM & 89.96$\pm$2.01 & 89.29$\pm$2.14 & 90.07$\pm$2.14 & 89.96$\pm$2.01 & 89.78$\pm$2.10\\
      & SVM-MRF & 93.72$\pm$2.95 & 93.30$\pm$3.14 & 93.19$\pm$3.95 & 93.72$\pm$2.95 & 93.08$\pm$3.57\\
      & SVM-CRF & 24.33$\pm$8.36 & 19.29$\pm$8.91 & 11.66$\pm$7.34 & 24.33$\pm$8.36 & 14.07$\pm$7.69\\
      & EMP-SVM & 96.11$\pm$1.46 & 95.85$\pm$1.55 & 96.40$\pm$1.27 & 96.11$\pm$1.46 & 96.10$\pm$1.44\\
      & EMP-SVM-MRF & \textbf{96.46}$\pm$1.42 & \textbf{96.23}$\pm$1.52 & \textbf{96.75}$\pm$1.22 & \textbf{96.46}$\pm$1.42 & \textbf{96.45}$\pm$1.42\\
      & EMP-SVM-CRF & 24.46$\pm$8.45 & 19.42$\pm$9.02 & 12.23$\pm$7.75 & 24.46$\pm$8.45 & 14.45$\pm$8.04\\
\midrule
\multirow{6}{*}{200} & SVM & 95.78$\pm$0.89 & 95.50$\pm$0.94 & 95.82$\pm$0.90 & 95.78$\pm$0.89 & 95.75$\pm$0.89\\
      & SVM-MRF & 98.97$\pm$0.51 & 98.90$\pm$0.54 & 99.03$\pm$0.47 & 98.97$\pm$0.51 & 98.96$\pm$0.51\\
      & SVM-CRF & 97.83$\pm$1.50 & 97.69$\pm$1.60 & 97.92$\pm$2.05 & 97.83$\pm$1.50 & 97.73$\pm$1.91\\
      & EMP-SVM & 99.64$\pm$0.22 & 99.62$\pm$0.24 & 99.65$\pm$0.21 & 99.64$\pm$0.22 & 99.64$\pm$0.22\\
      & EMP-SVM-MRF & \textbf{99.69}$\pm$0.20 & \textbf{99.67}$\pm$0.21 & \textbf{99.70}$\pm$0.19 & \textbf{99.69}$\pm$0.20 & \textbf{99.69}$\pm$0.20\\
      & EMP-SVM-CRF & 99.61$\pm$0.27 & 99.59$\pm$0.28 & 99.63$\pm$0.26 & 99.61$\pm$0.27 & 99.61$\pm$0.27\\
\bottomrule
\end{tabular}
}
\end{table*}
\begin{figure*}[!h]
\centering
	 \subfloat[][Ground truth]{\includegraphics{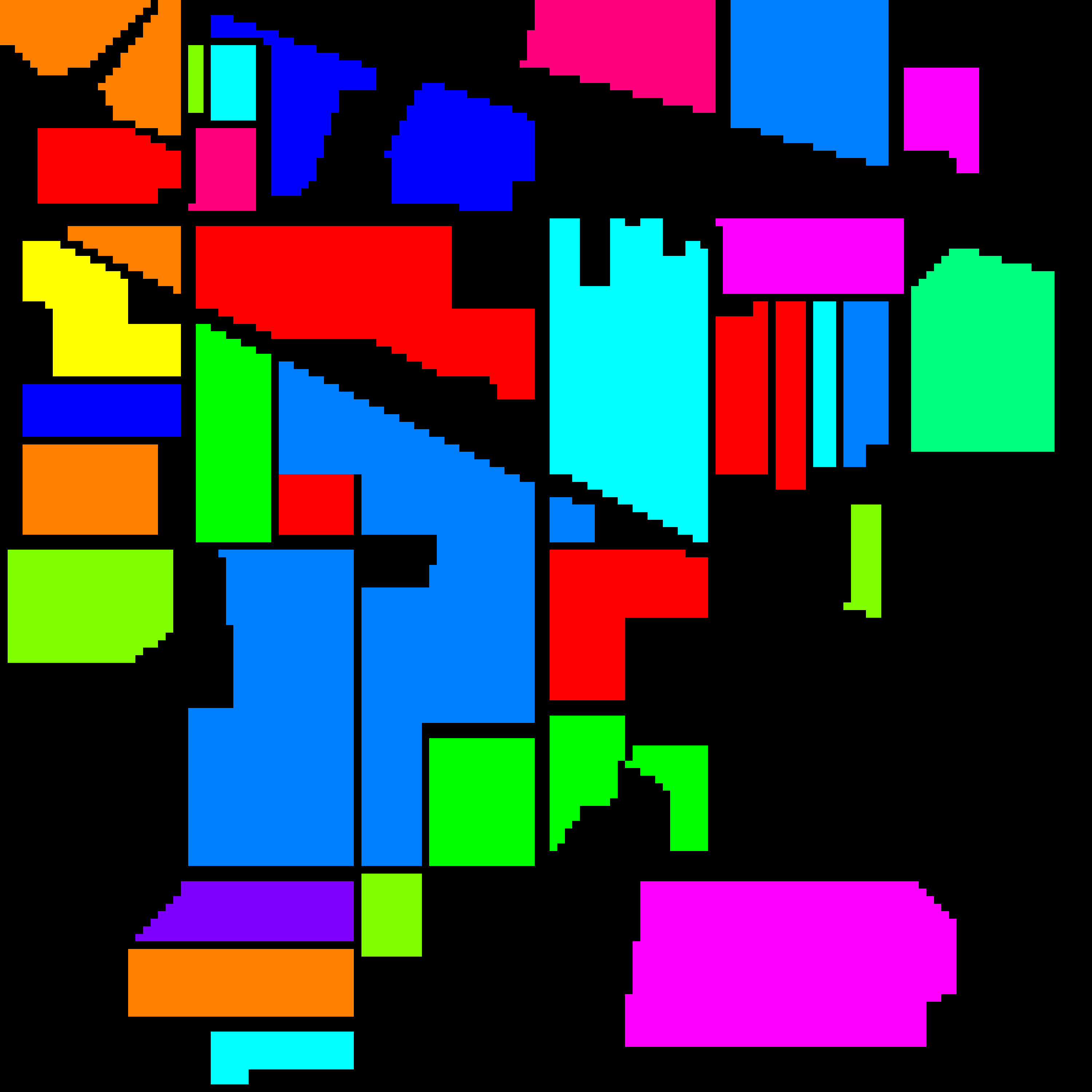}} \hfill
     \subfloat[][SVM]{\includegraphics{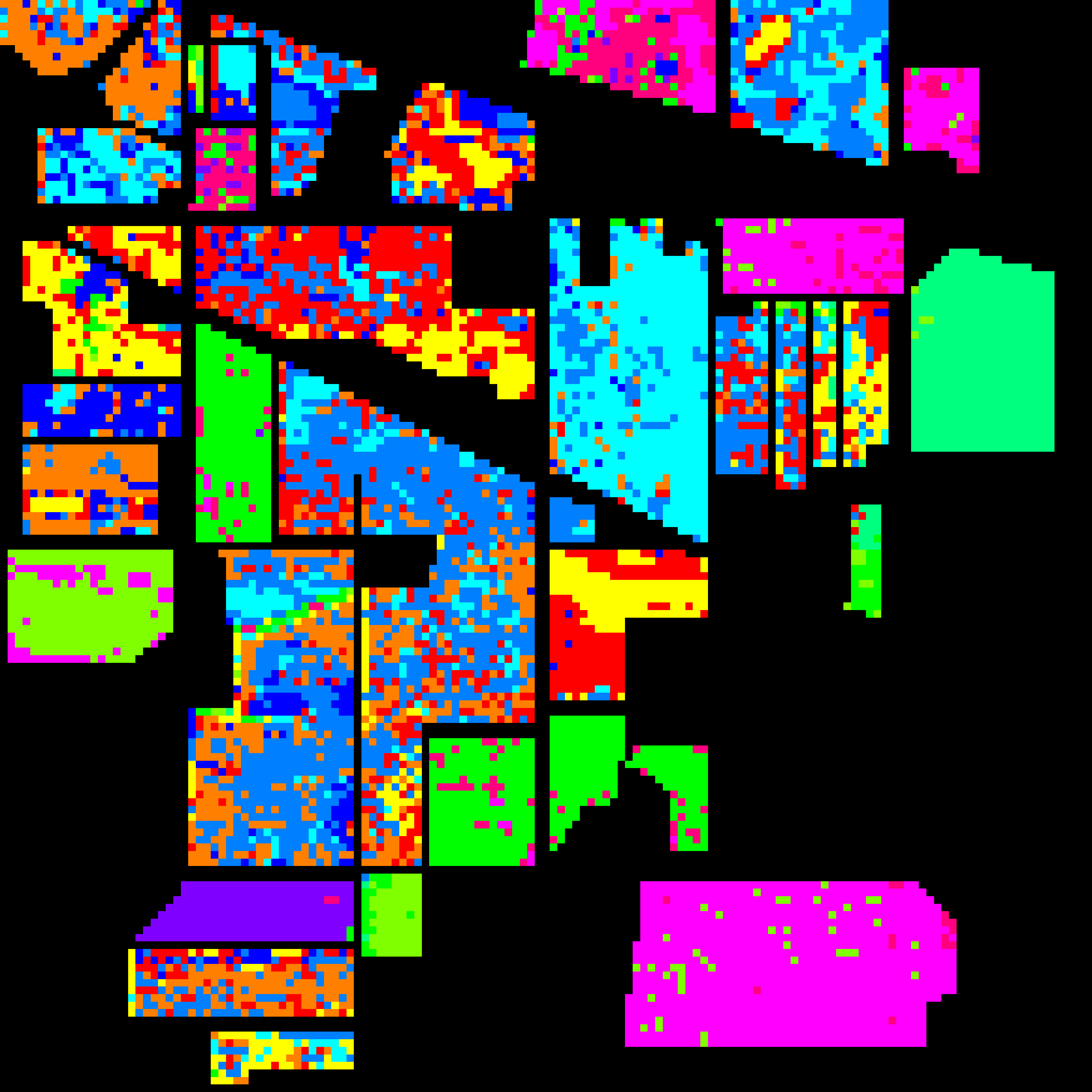}} \hfill
     \subfloat[][SVM-MRF]{\includegraphics{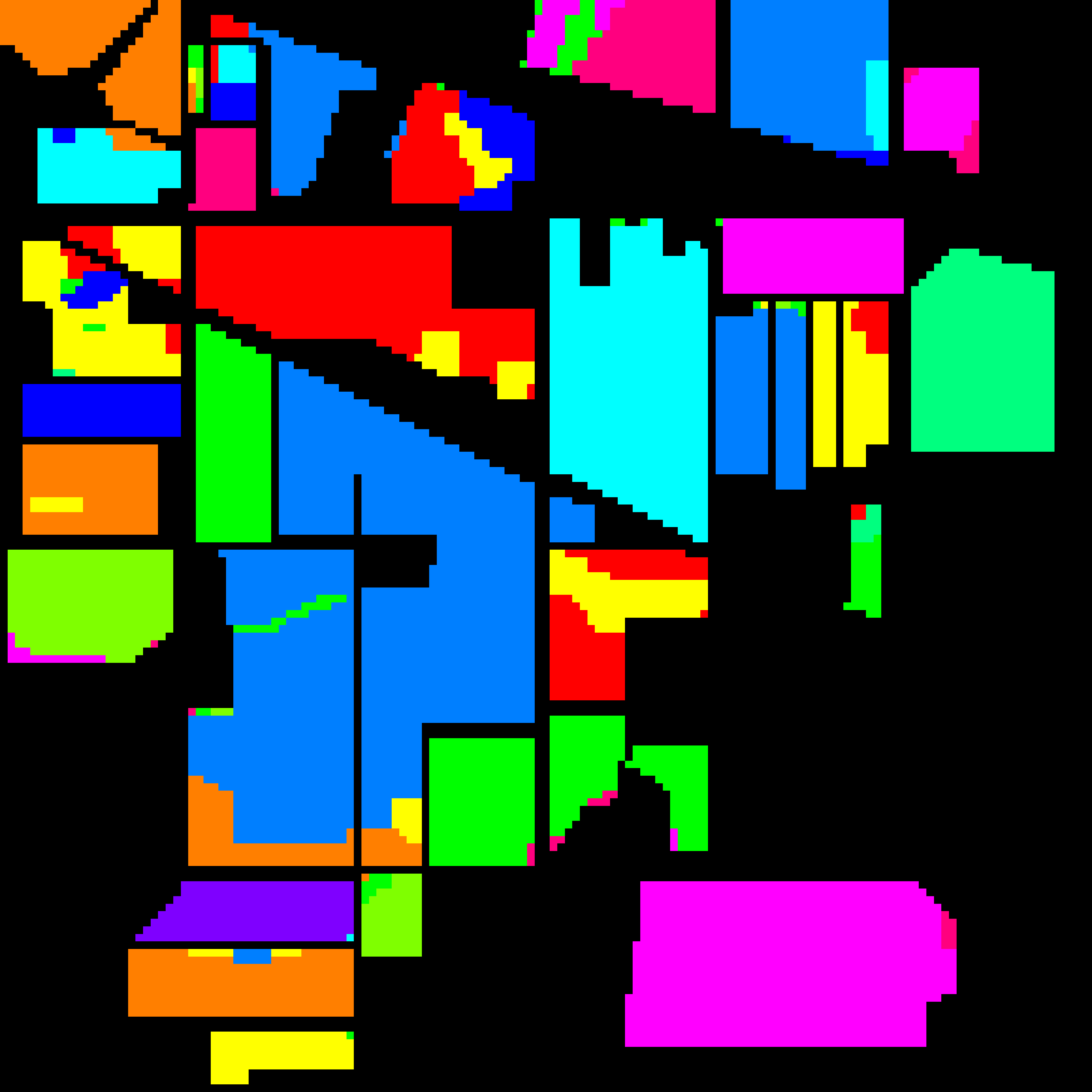}} \\
     \subfloat[][SVM-CRF]{\includegraphics{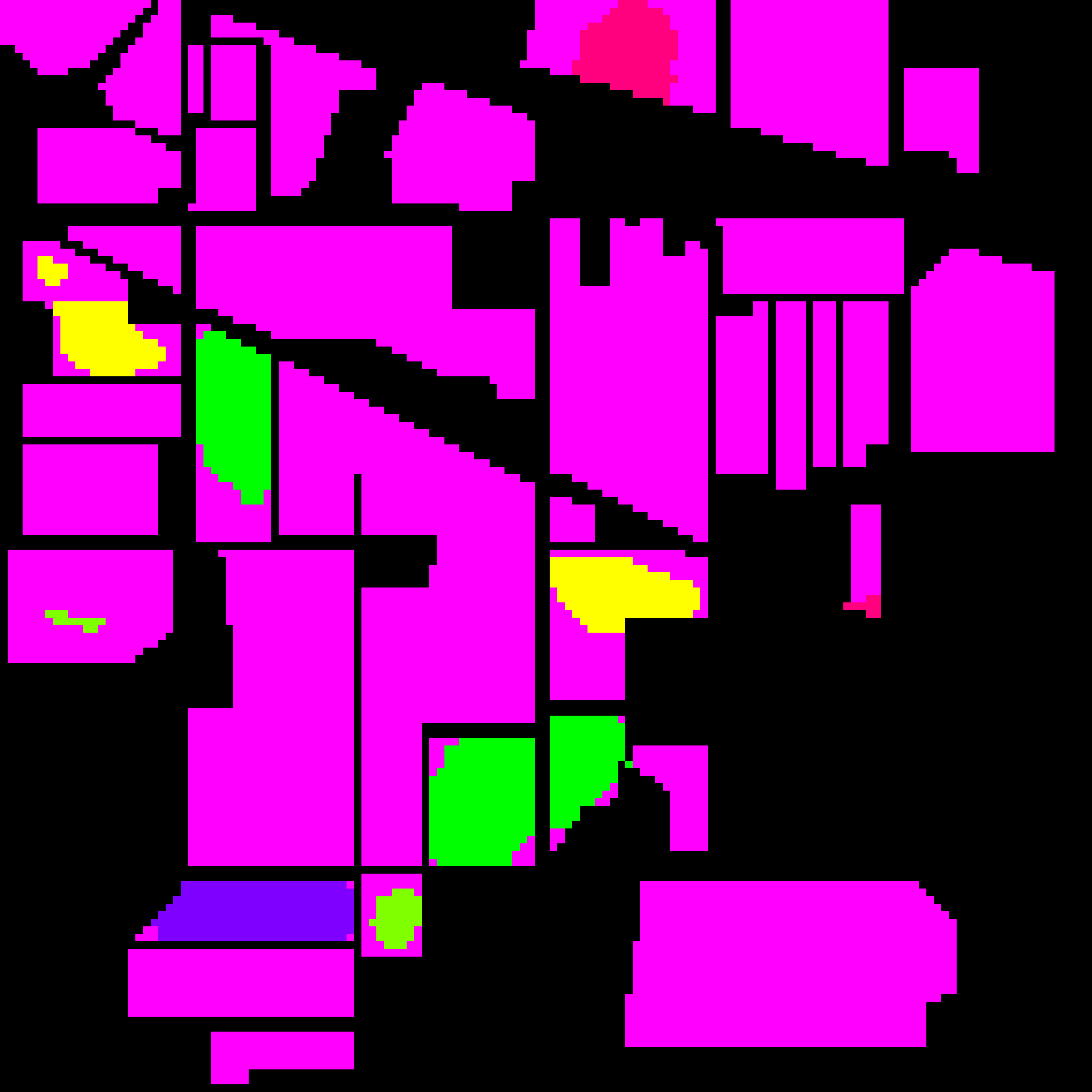}} \hfill
     \subfloat[][EMP-SVM]{\includegraphics{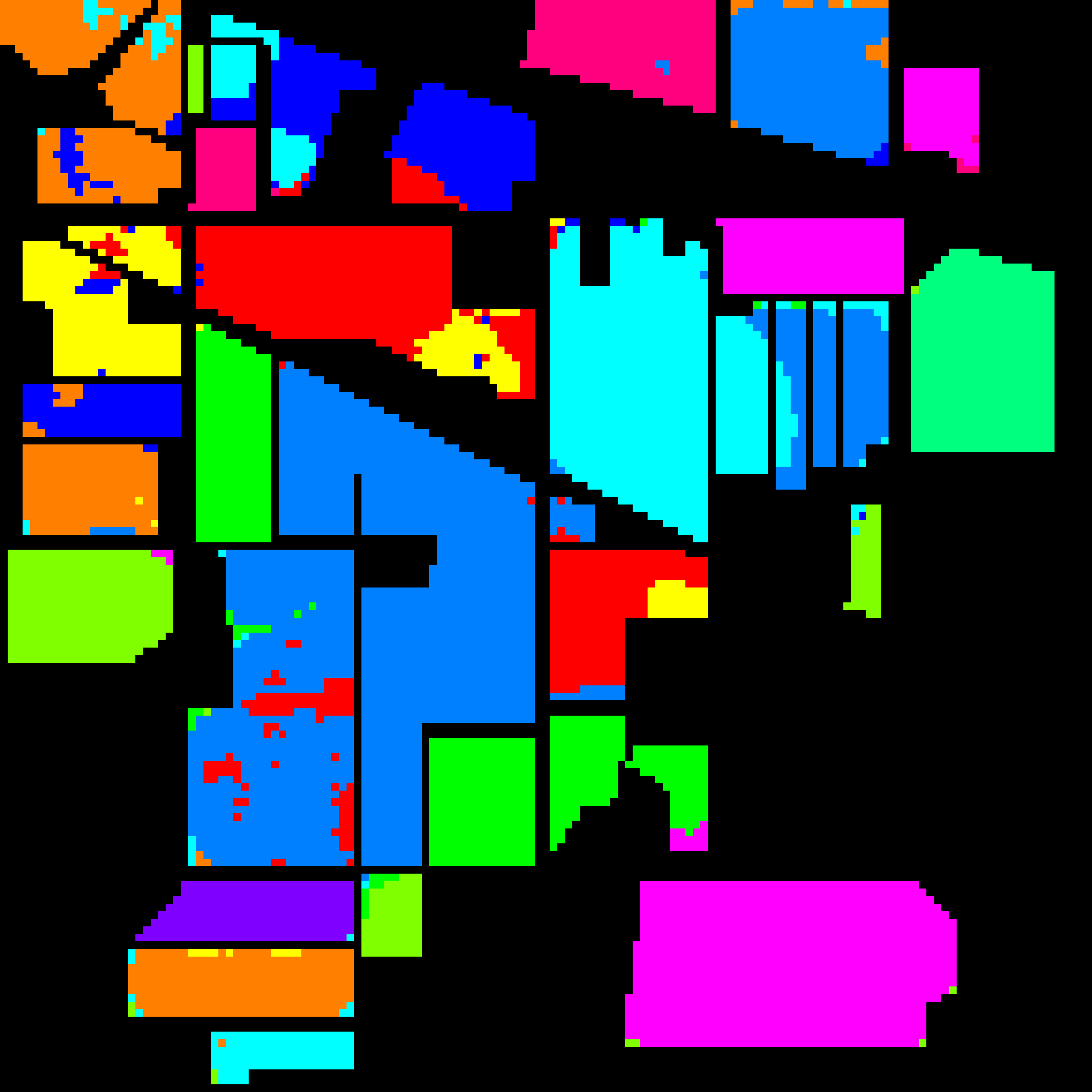}} \hfill
     \subfloat[][EMP-SVM-MRF]{\includegraphics{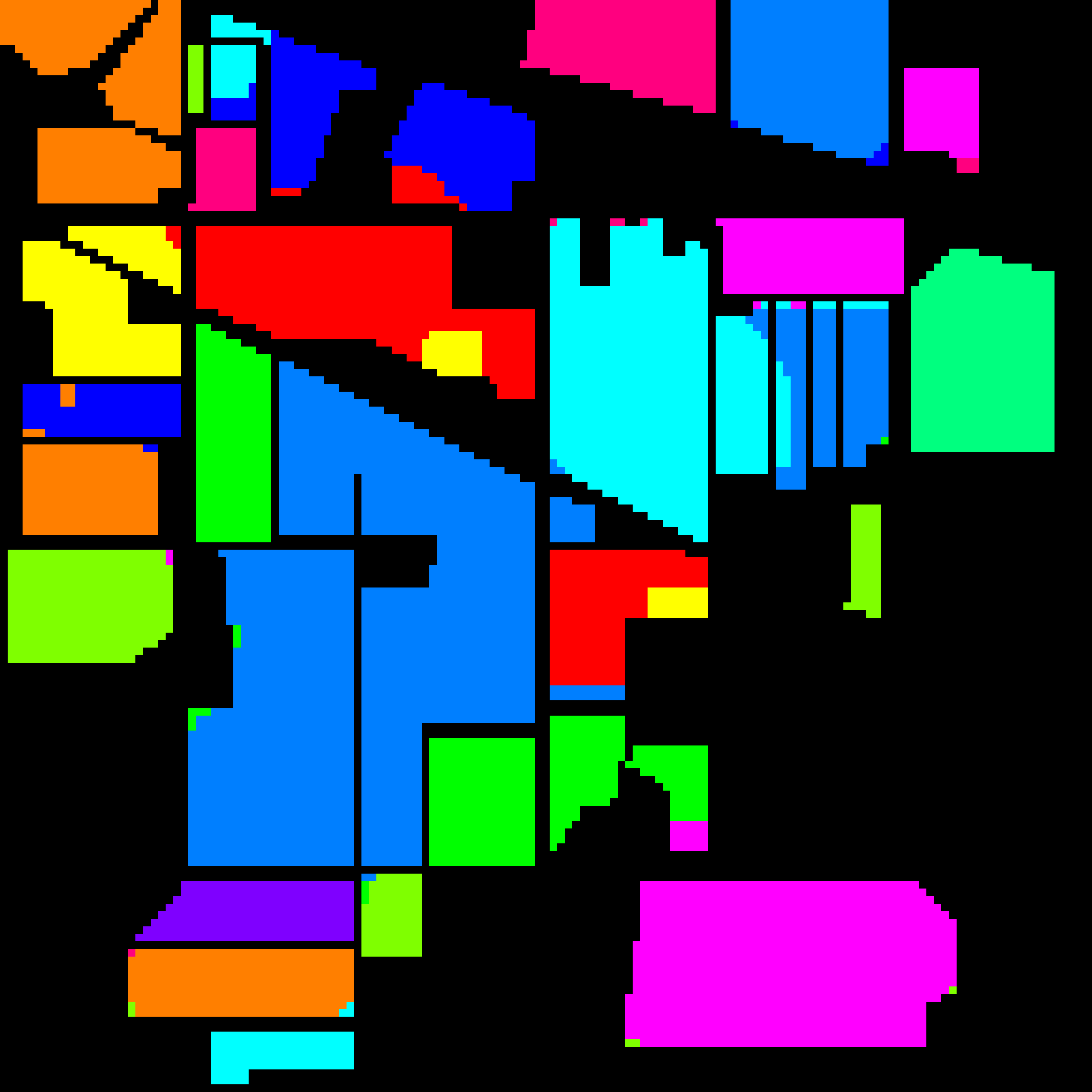}} \hfill
\caption{Predicted land cover maps for the Indian Pines image (20 training pixels per class).}
\label{fig:indian_pines_images_20}
\end{figure*}
\begin{figure*}[!h]
\centering
	 \subfloat[][Ground truth]{\includegraphics{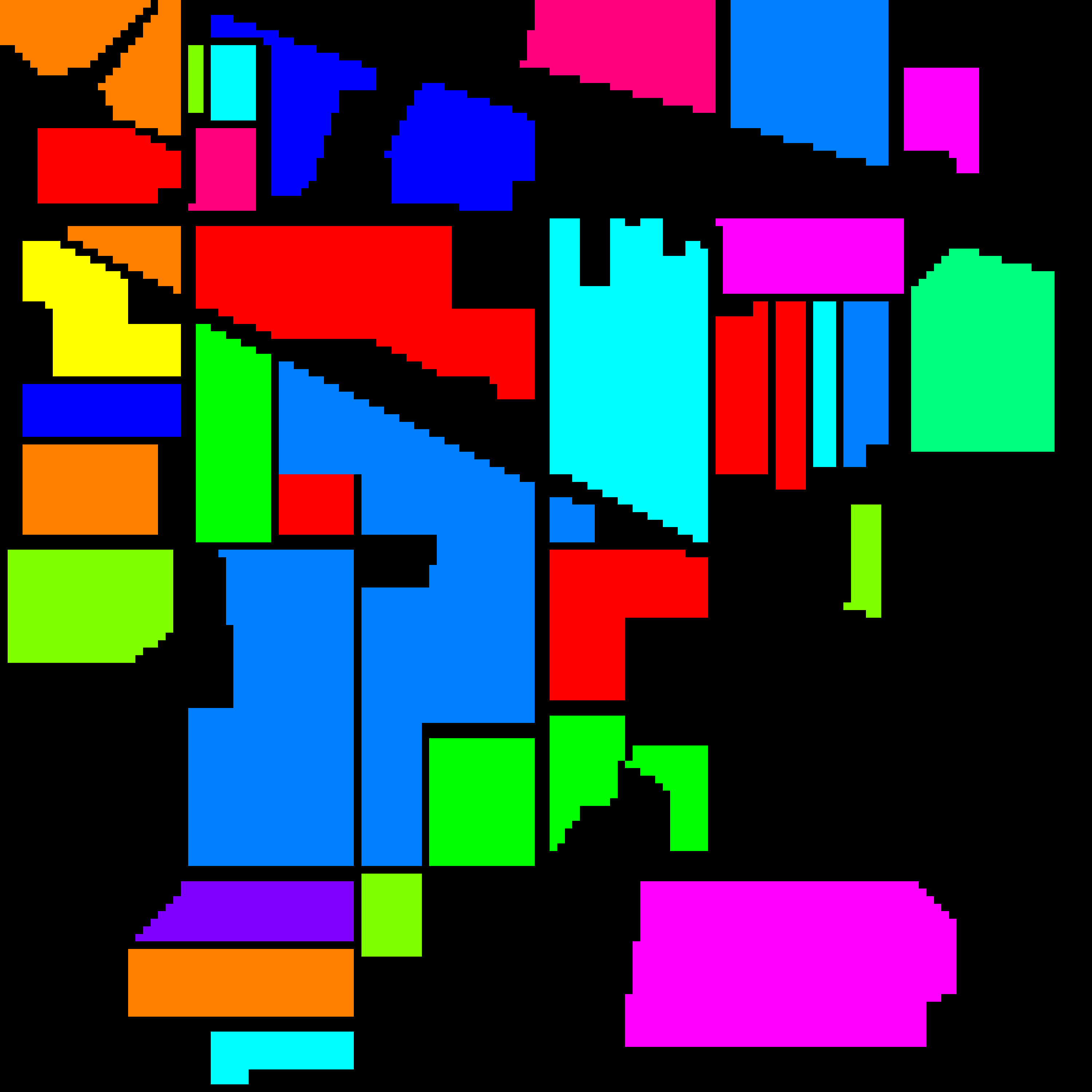}} \hfill
     \subfloat[][SVM]{\includegraphics{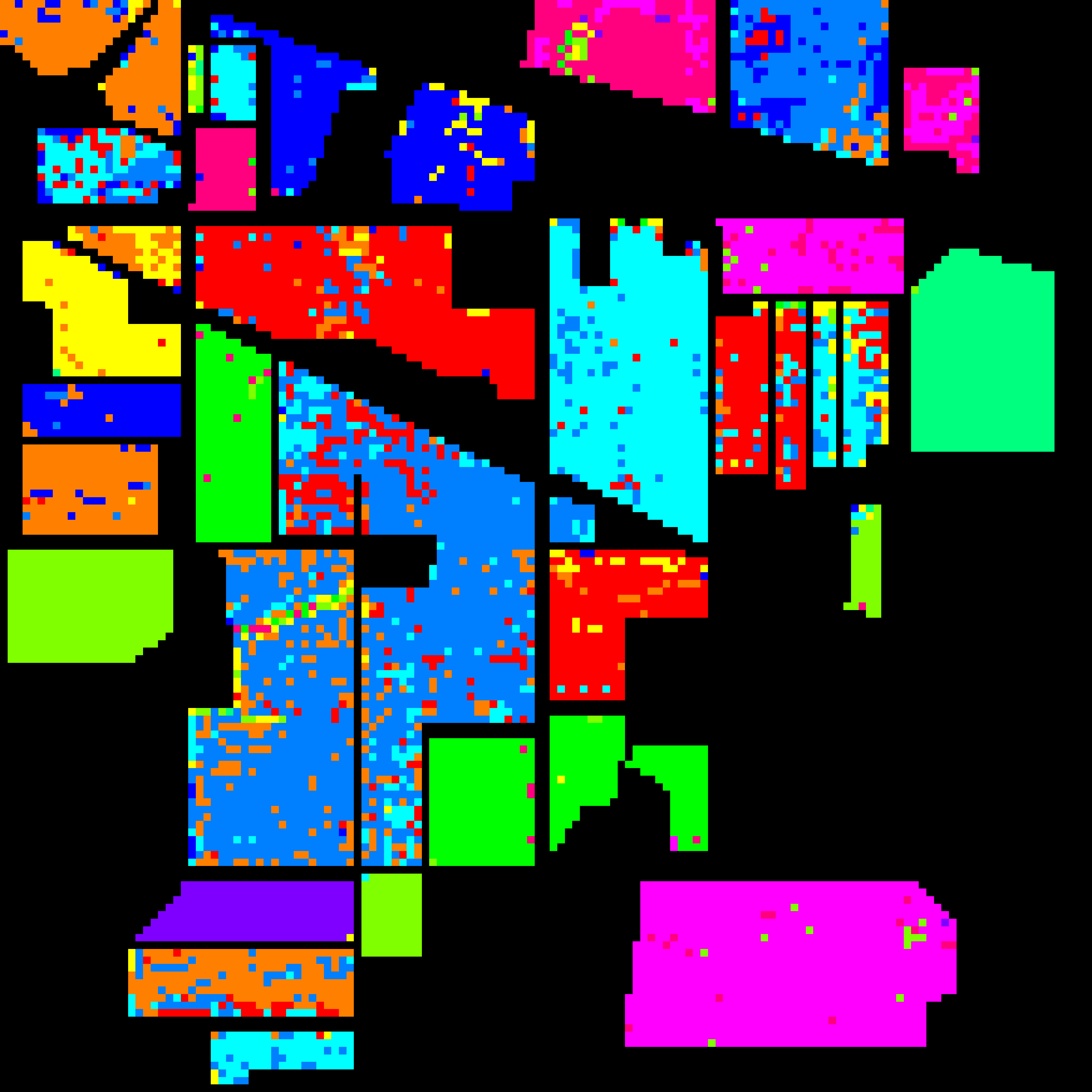}} \hfill
     \subfloat[][SVM-MRF]{\includegraphics{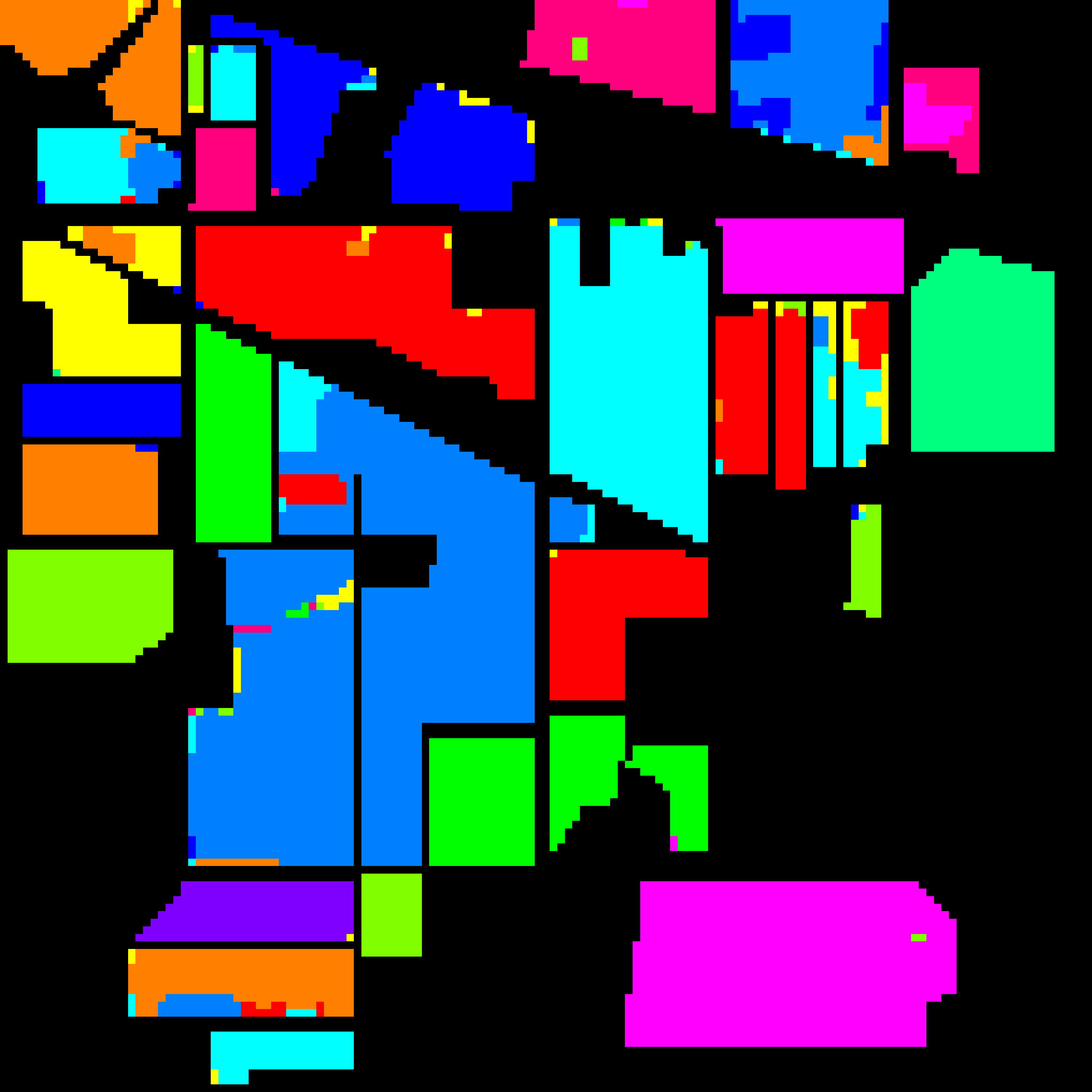}} \\
     \subfloat[][SVM-CRF]{\includegraphics{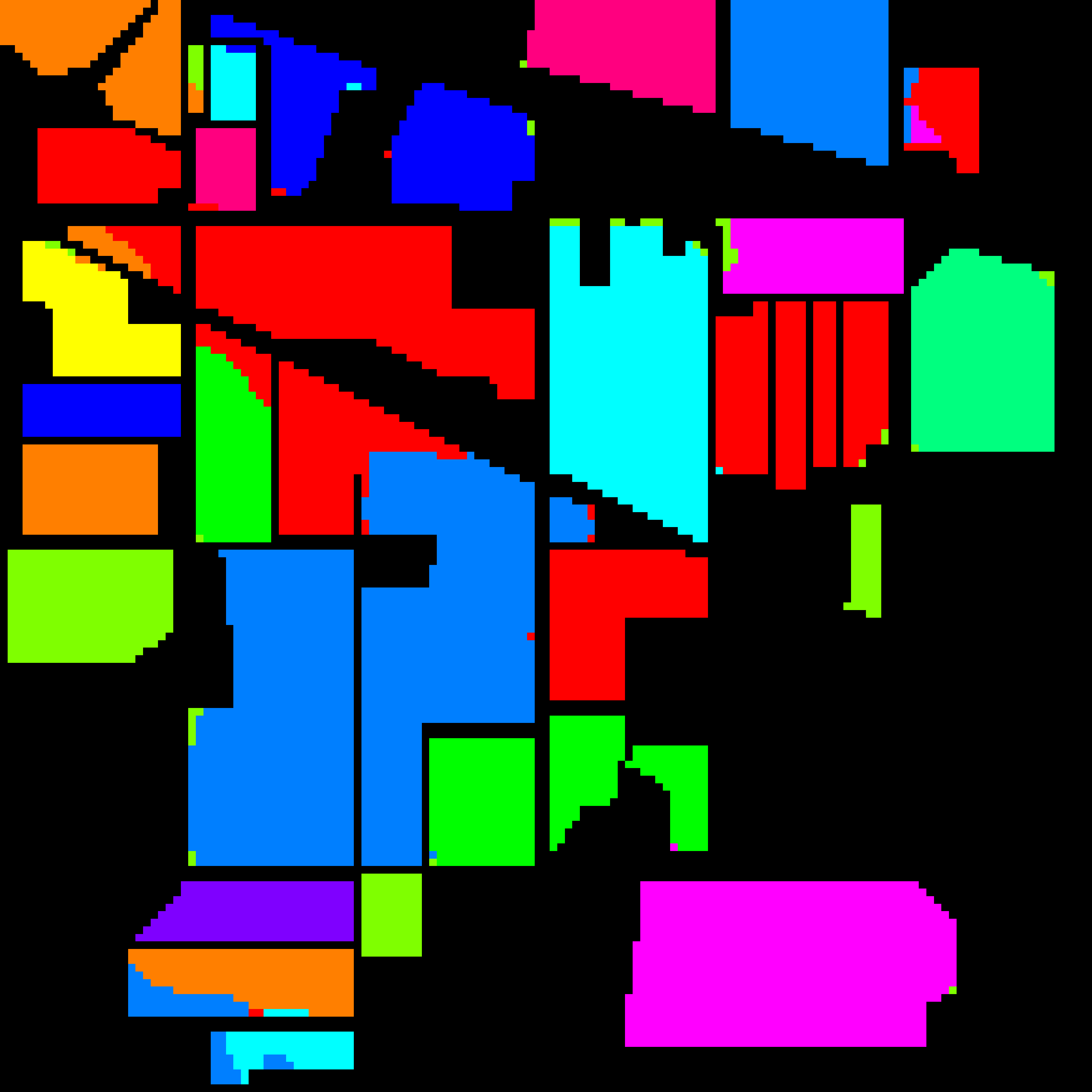}} \hfill
     \subfloat[][EMP-SVM]{\includegraphics{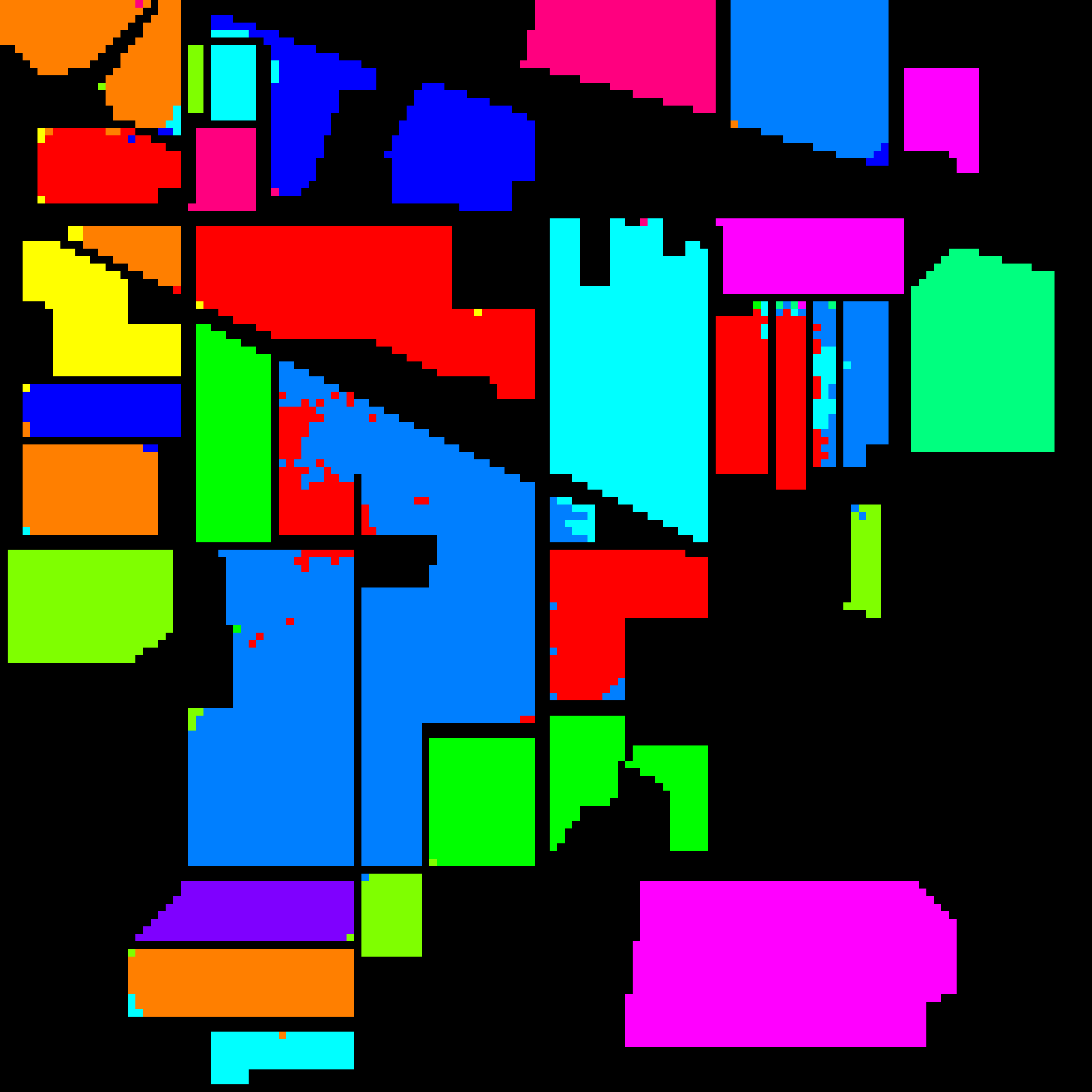}} \hfill
     \subfloat[][EMP-SVM-MRF]{\includegraphics{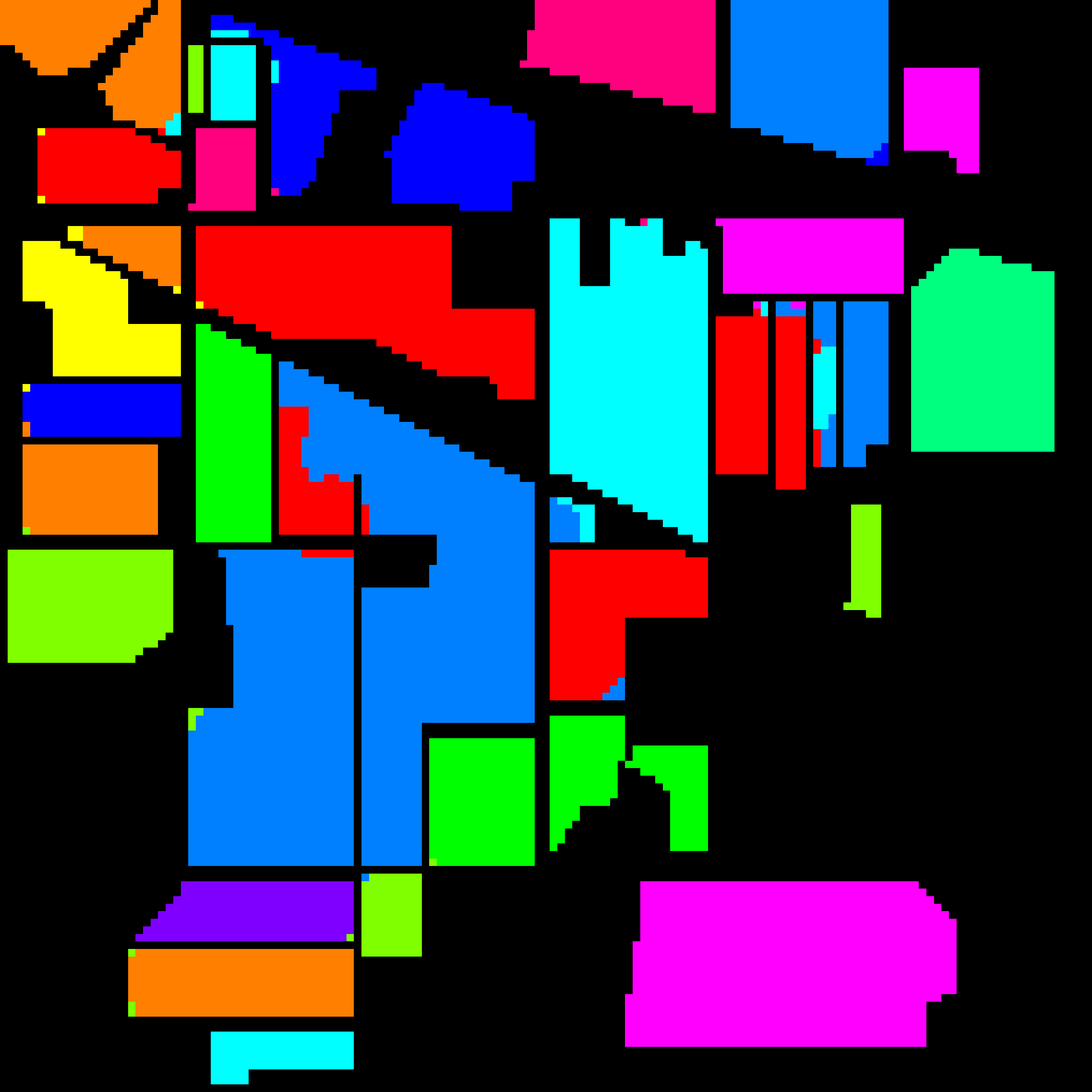}} \hfill
\caption{Predicted land cover maps for the Indian Pines image (140 training pixels per class).}
\label{fig:indian_pines_images_200}
\end{figure*}
\begin{figure*}[!h]
\centering
	 \subfloat[][Ground truth]{\includegraphics{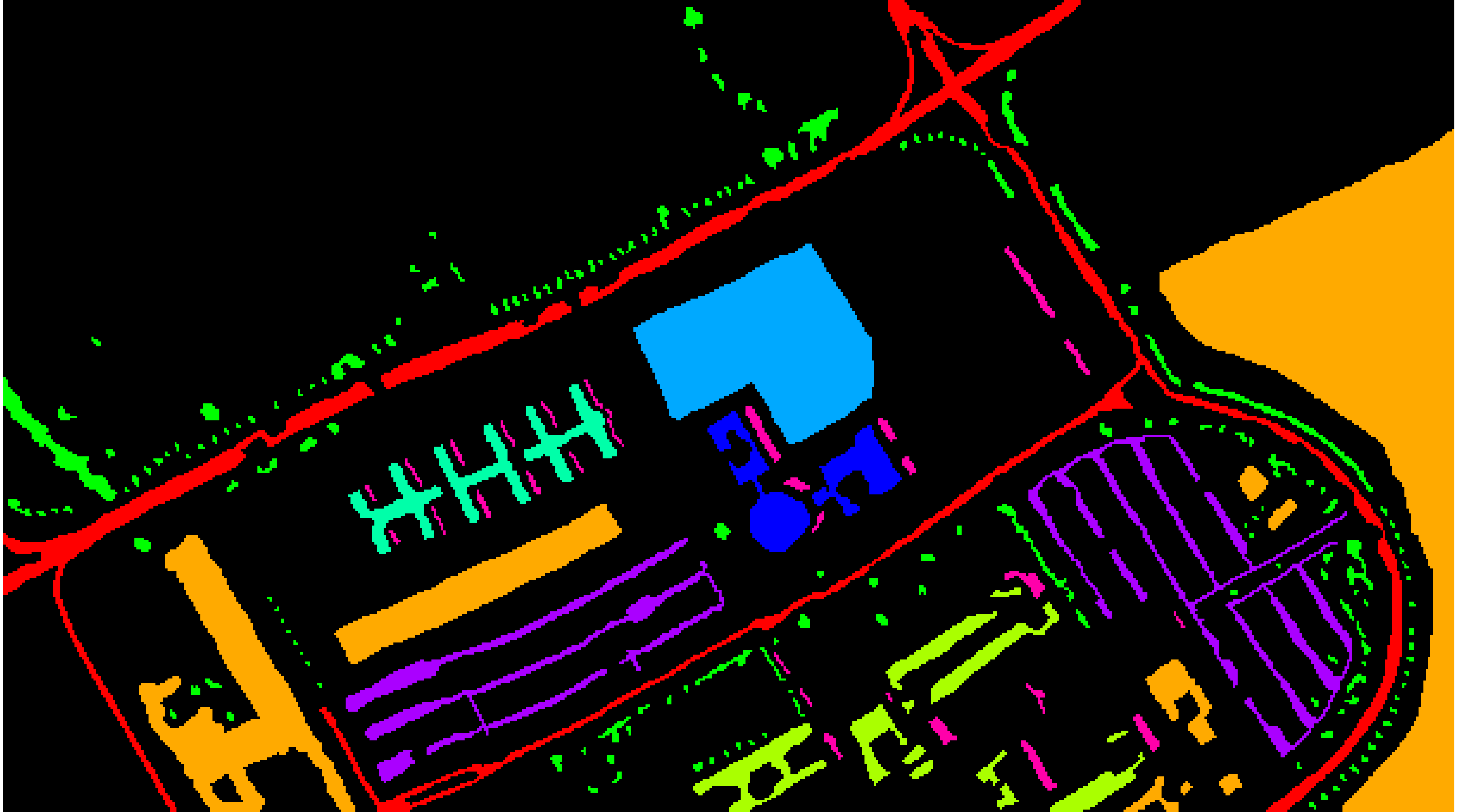}} \hfill
     \subfloat[][SVM]{\includegraphics{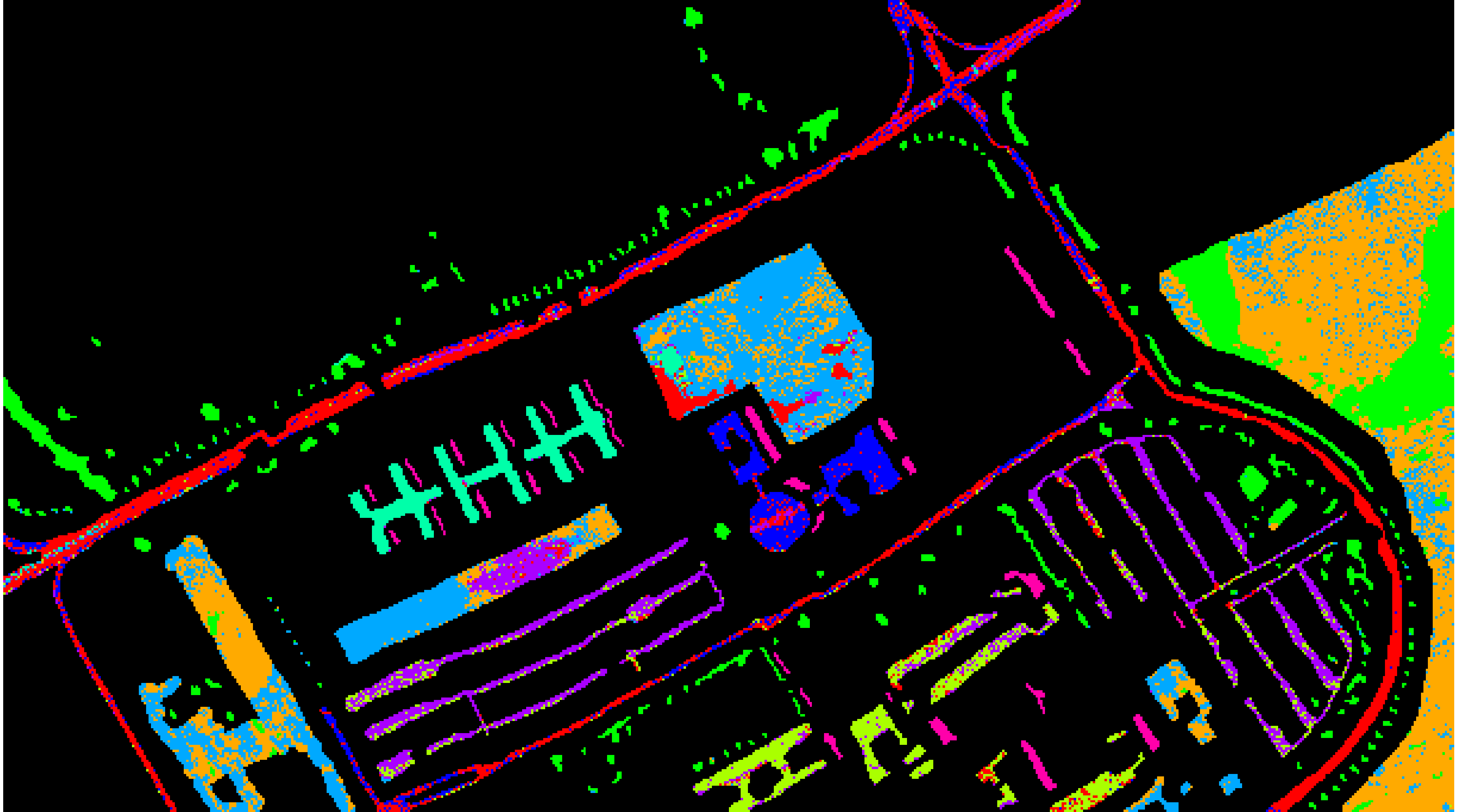}} \hfill
     \subfloat[][SVM-MRF]{\includegraphics{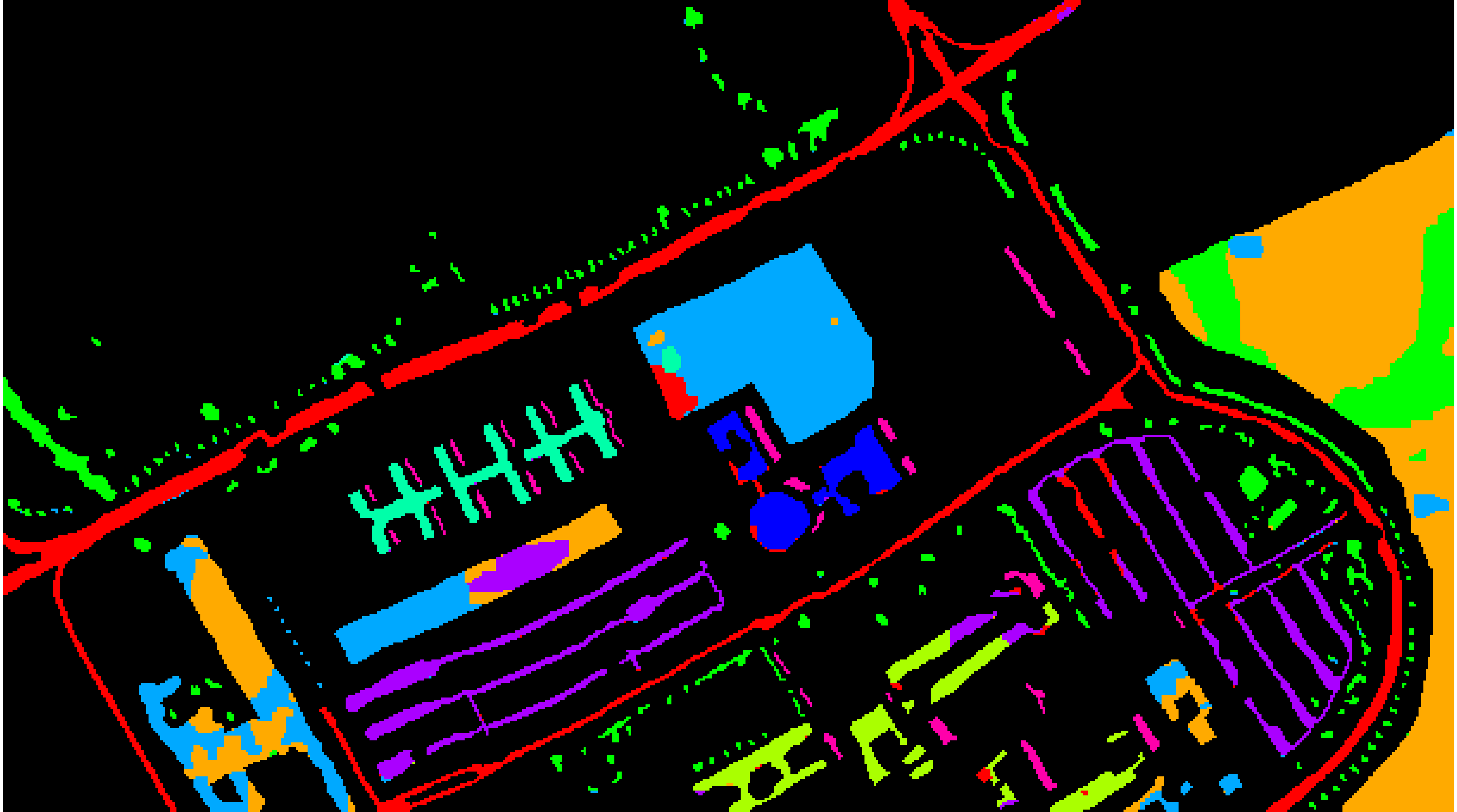}} \hfill
     \subfloat[][SVM-CRF]{\includegraphics{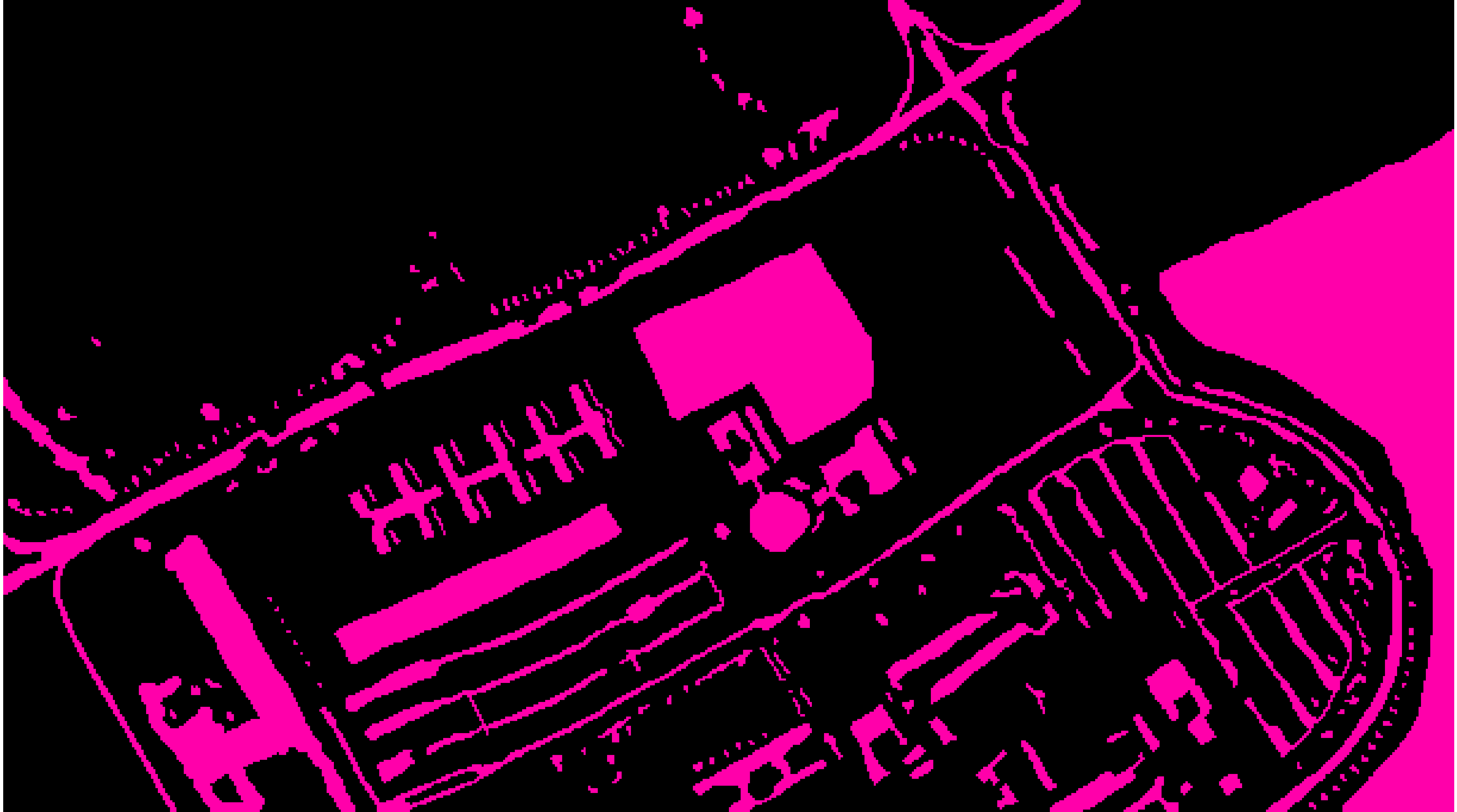}} \hfill
     \subfloat[][EMP-SVM]{\includegraphics{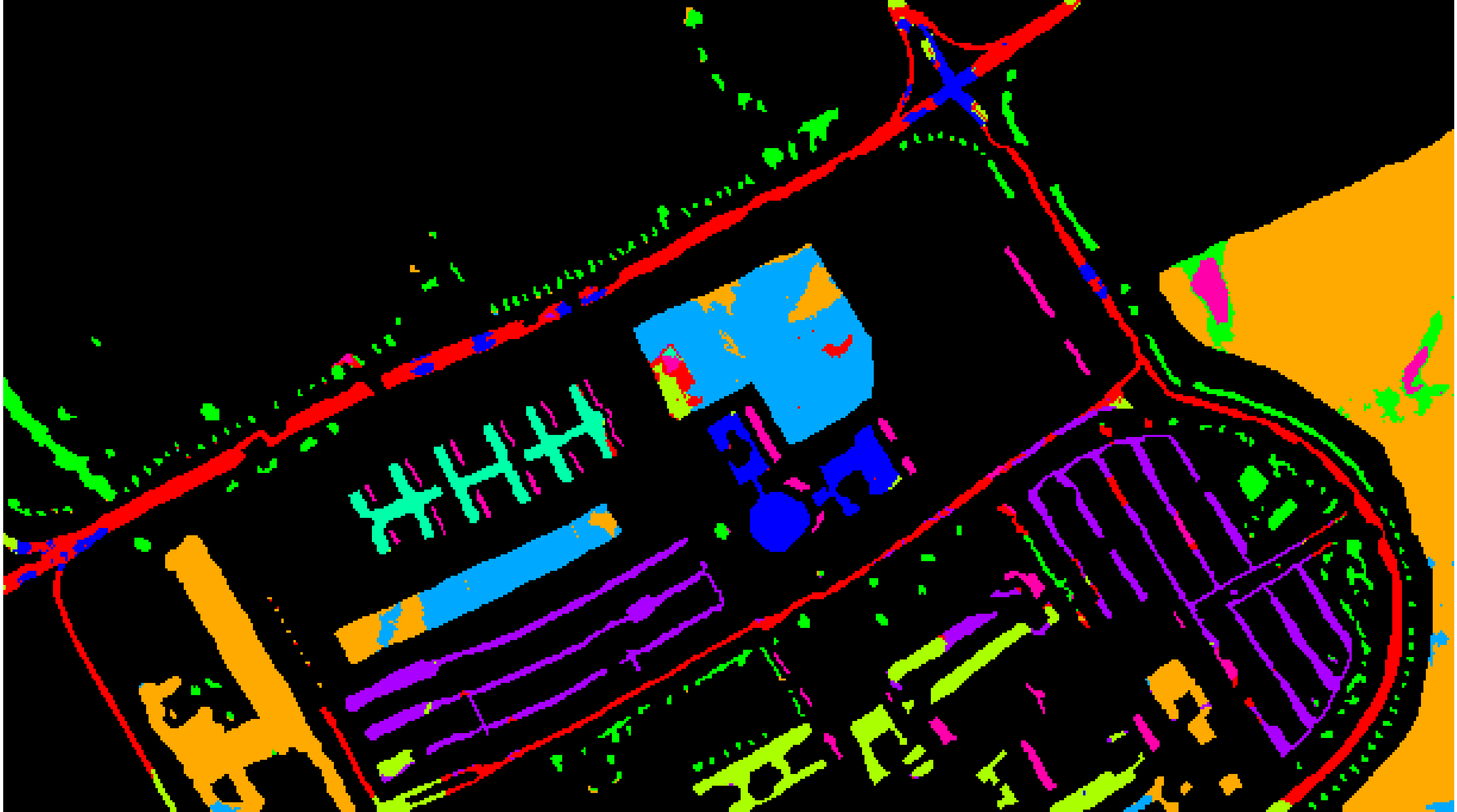}} \hfill
     \subfloat[][EMP-SVM-MRF]{\includegraphics{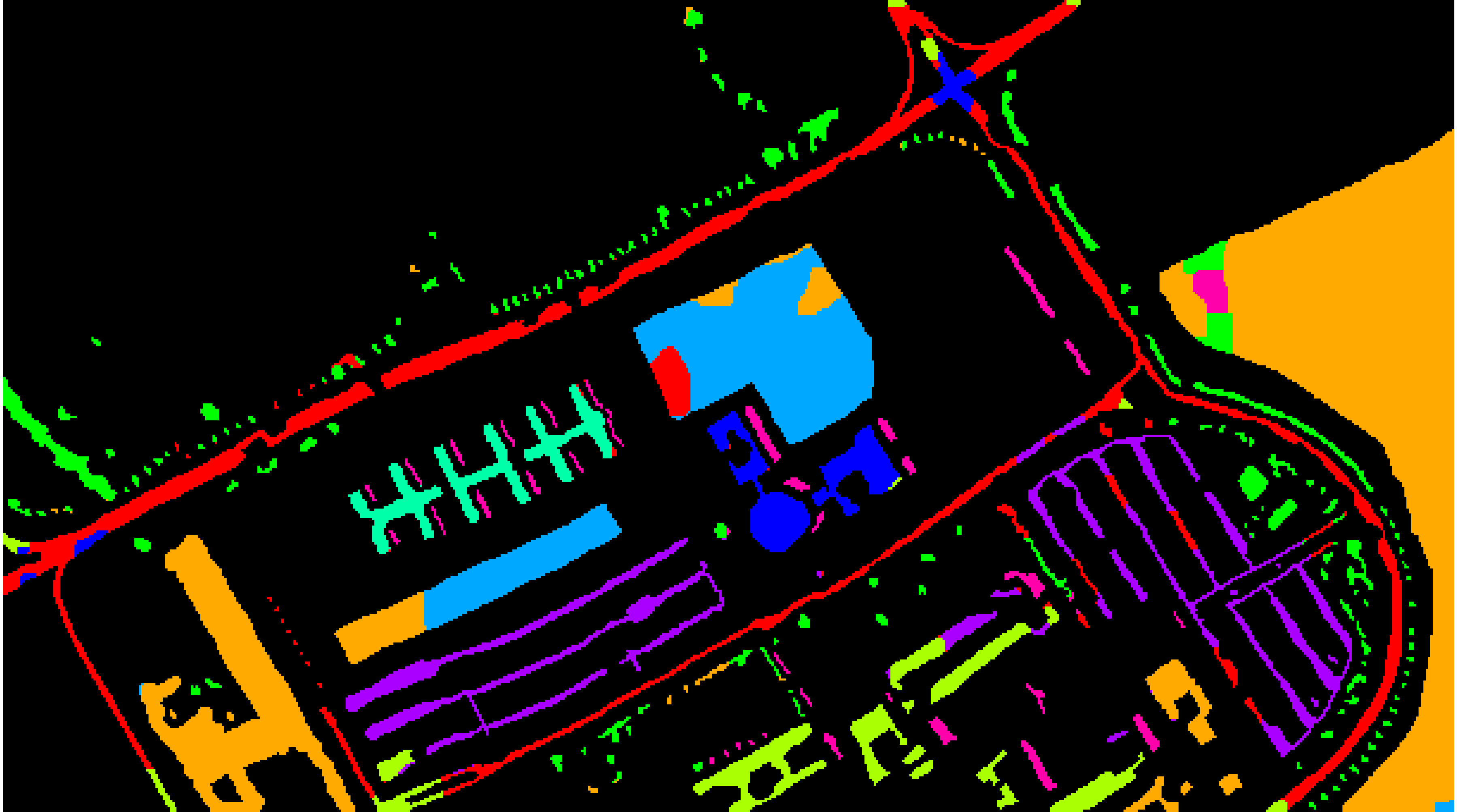}} \hfill
\caption{Predicted land cover maps for the University of Pavia image (20 training pixels per class).}
\label{fig:pavia_images_20}
\end{figure*}
\begin{figure*}[!h]
\centering
	 \subfloat[][Ground truth]{\includegraphics{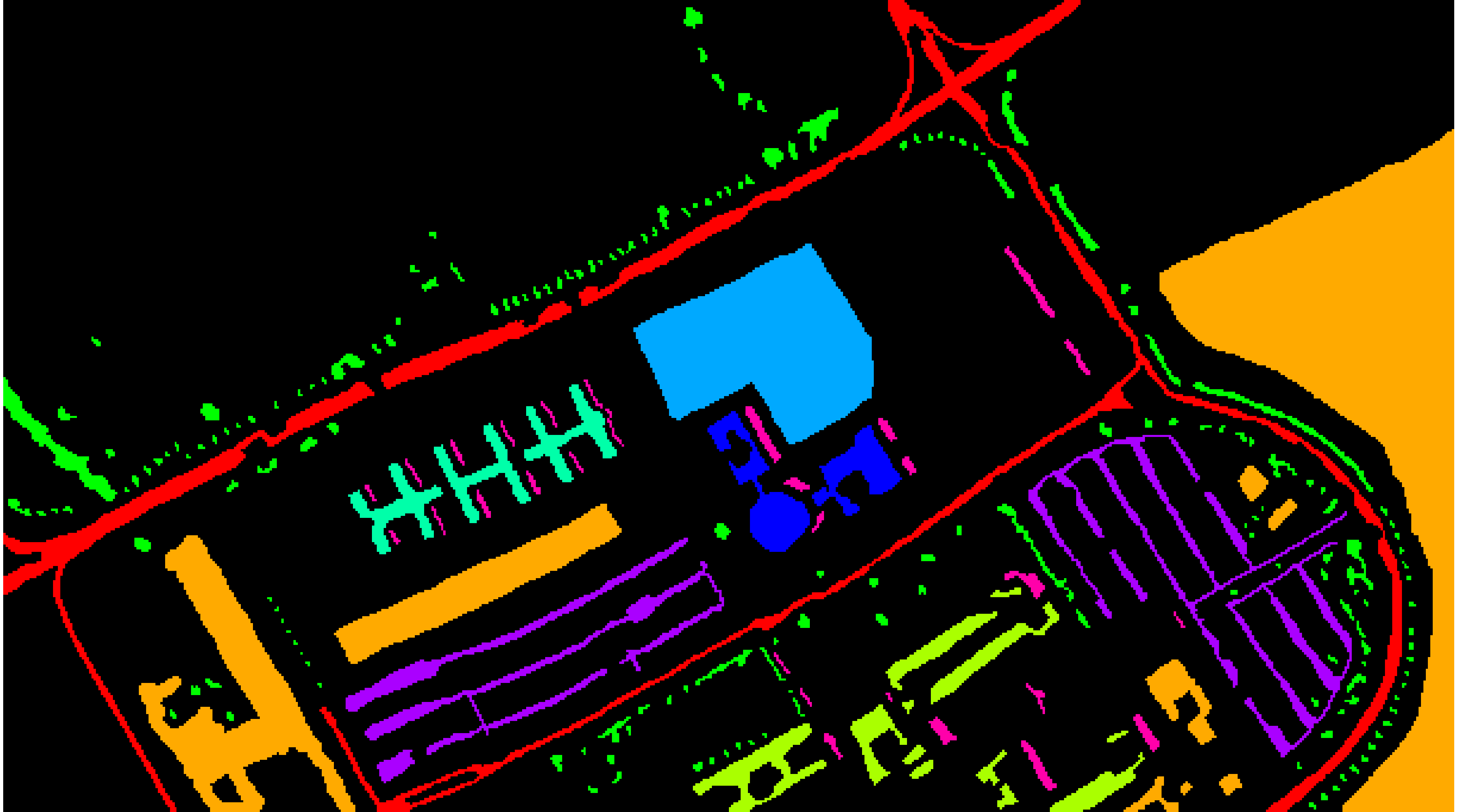}} \hfill
     \subfloat[][SVM]{\includegraphics{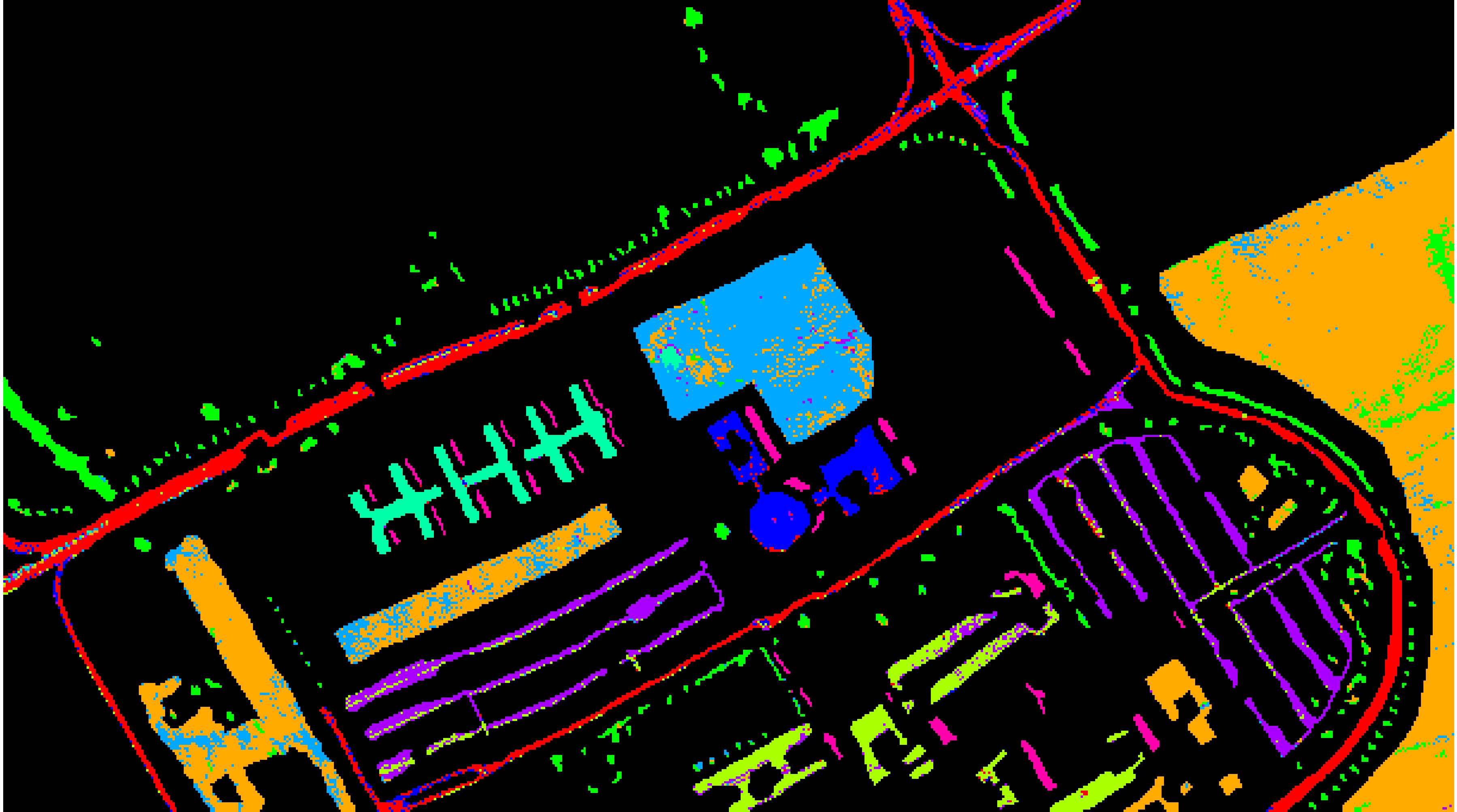}} \hfill
     \subfloat[][SVM-MRF]{\includegraphics{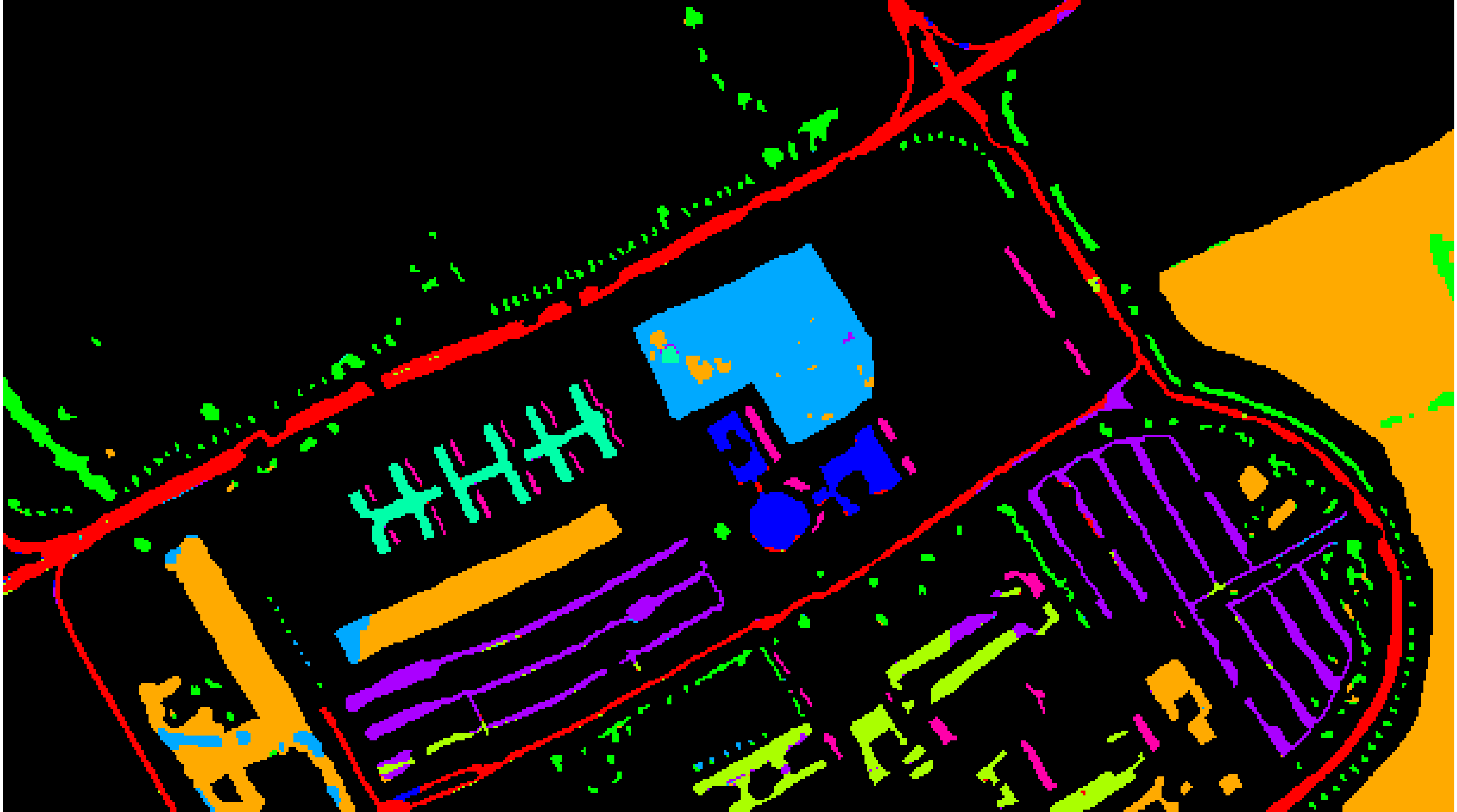}} \hfill
     \subfloat[][SVM-CRF]{\includegraphics{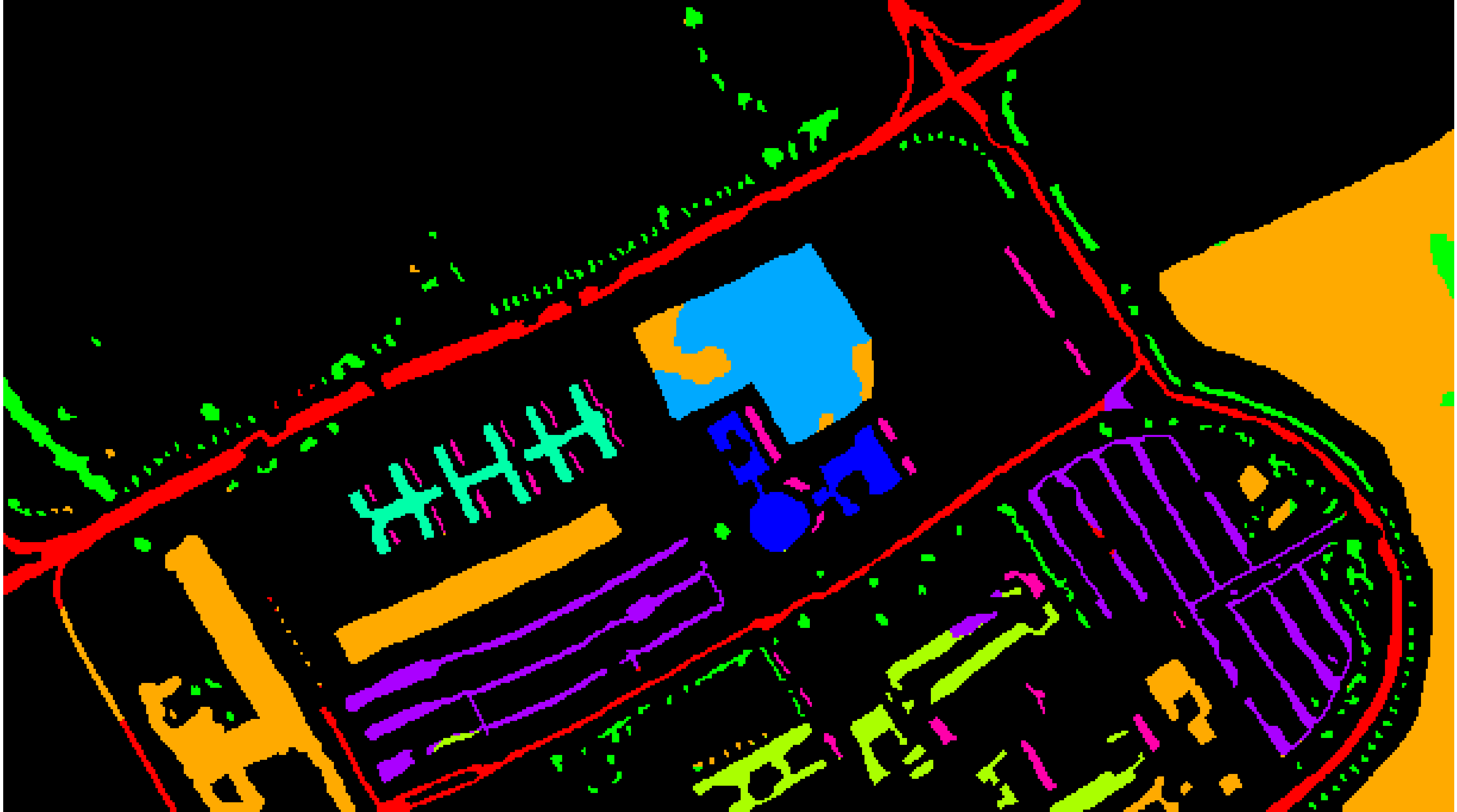}} \hfill
     \subfloat[][EMP-SVM]{\includegraphics{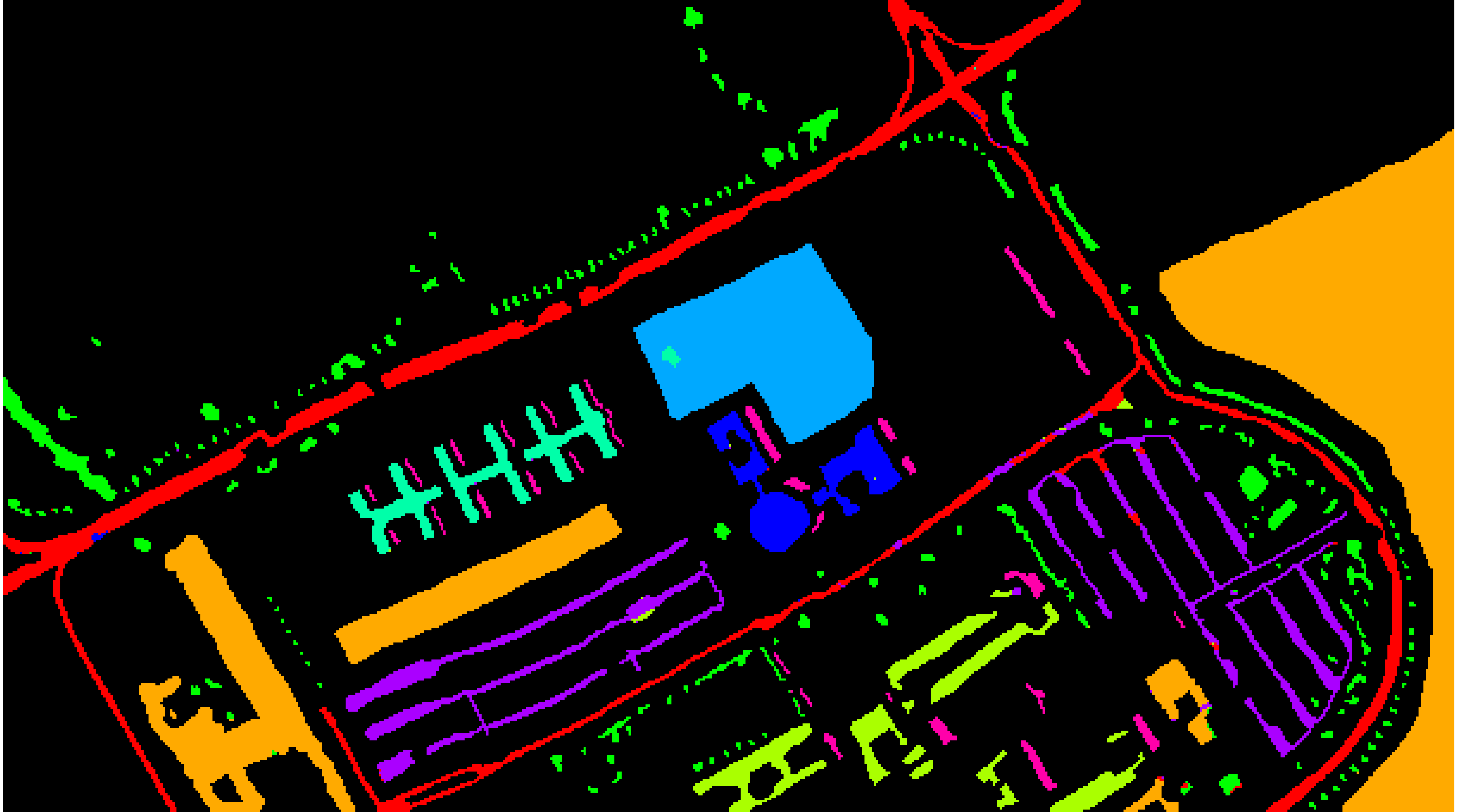}} \hfill
     \subfloat[][EMP-SVM-MRF]{\includegraphics{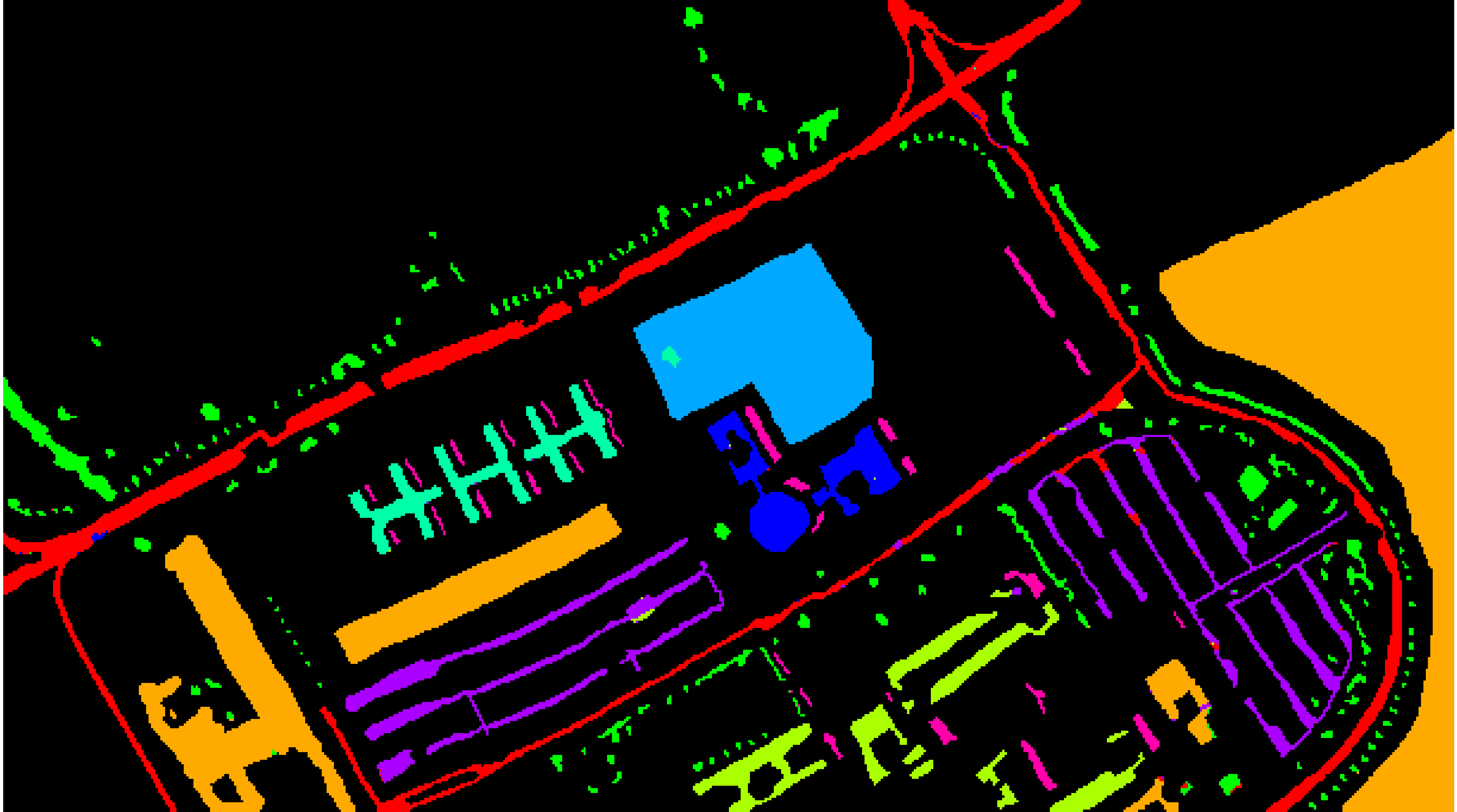}} \hfill
\caption{Predicted land cover maps for the University of Pavia image (140 training pixels per class).}
\label{fig:pavia_images_200}
\end{figure*}

\subsubsection{Discussion}
The results demonstrate the benefits of using MRFs and CRFs. The MRF increased the mean accuracy of the pixel-wise classifiers in all cases. However, we obtained mixed results from the CRF. The CRF showed high variance in performance, especially for cases with low number of training example. The performance of the CRF was poor compared to the MRF for smaller training size. However, as the size of the training set was increased the performance of CRF increased, surpassing the performance of MRF for some methods. We did not experiment with training sizes greater than 140, as at that training size other methods were already producing mean accuracies of around 99\%. The reason that CRF was not well suited in the experiment is that the training set is limited. In theory, the CRF should produce better result than the MRF when enough training data is available because it is a discriminative model and is also more expressive. The CRF with log-linear potentials is more appropriate for cases where there are multiple fully labeled training images and the separate test images. However, this is not mostly the case in remote sensing. The dependence of CRF's performance on training set size can be further seen in table \ref{tab:OATable_elaborate}, where the performance of CRF is even better when the training size is 200 pixels per class. The classification maps in Figure~\ref{fig:indian_pines_images_20}, Figure~\ref{fig:indian_pines_images_200}, Figure~\ref{fig:pavia_images_20}, and Figure~\ref{fig:pavia_images_200} qualitatively show the same trends as observed in the tables.  

All the methods performed better when EMP features were used instead of the raw spectra. The SVM and the random forest classifiers were the best pixel-wise classifiers. The GP produced results comparable to the SVM. For both GP and SVM, better results were obtained with squared exponential kernel/covariance function. This indicate the spectral angle mapper based ESAM is not necessarily better for classification. However, the combination of SAM and MRF produced good surprising results rivaling the performance of SVM and random forest based MRFs and CRFs. SAM is essentially a nearest neighbor search with the angle as the distance metric. It does not require any training and is very fast for datasets with hundreds of training example.

If we are to compare the methods that combine spatial-spectral features and pixel-wise classifier only, such as EMP-SVM(SE), and the methods that combine pixel-wise classifier and graphical models only, such as SVM(SE)-MRF, the former always outperformed. However, methods that utilize all three components--spatial-spectral features, classifier, and UGM, were always the best. This shows an advantage of UGM. They can be added to pre-existing models to further improve the performance. The current trend in hyperspectral remote sensing is to develop more accurate spatial-spectral features with deep learning. The performance of these methods could be further improved using UGMs. 

\begin{figure*}[!h]
\centering
	 \subfloat[][RGB image]{\includegraphics{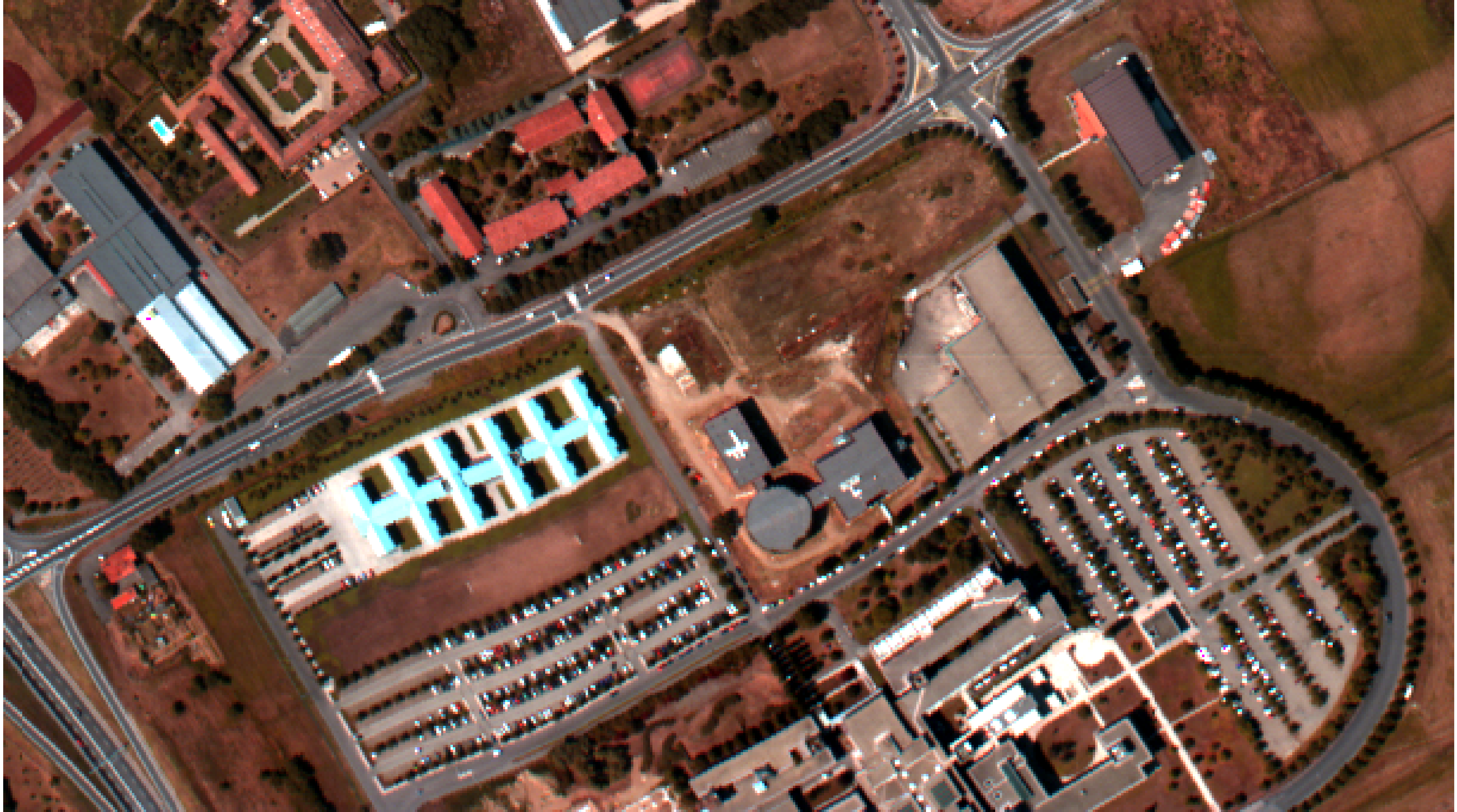}} \hfill
	 \subfloat[][Random forest]{\includegraphics{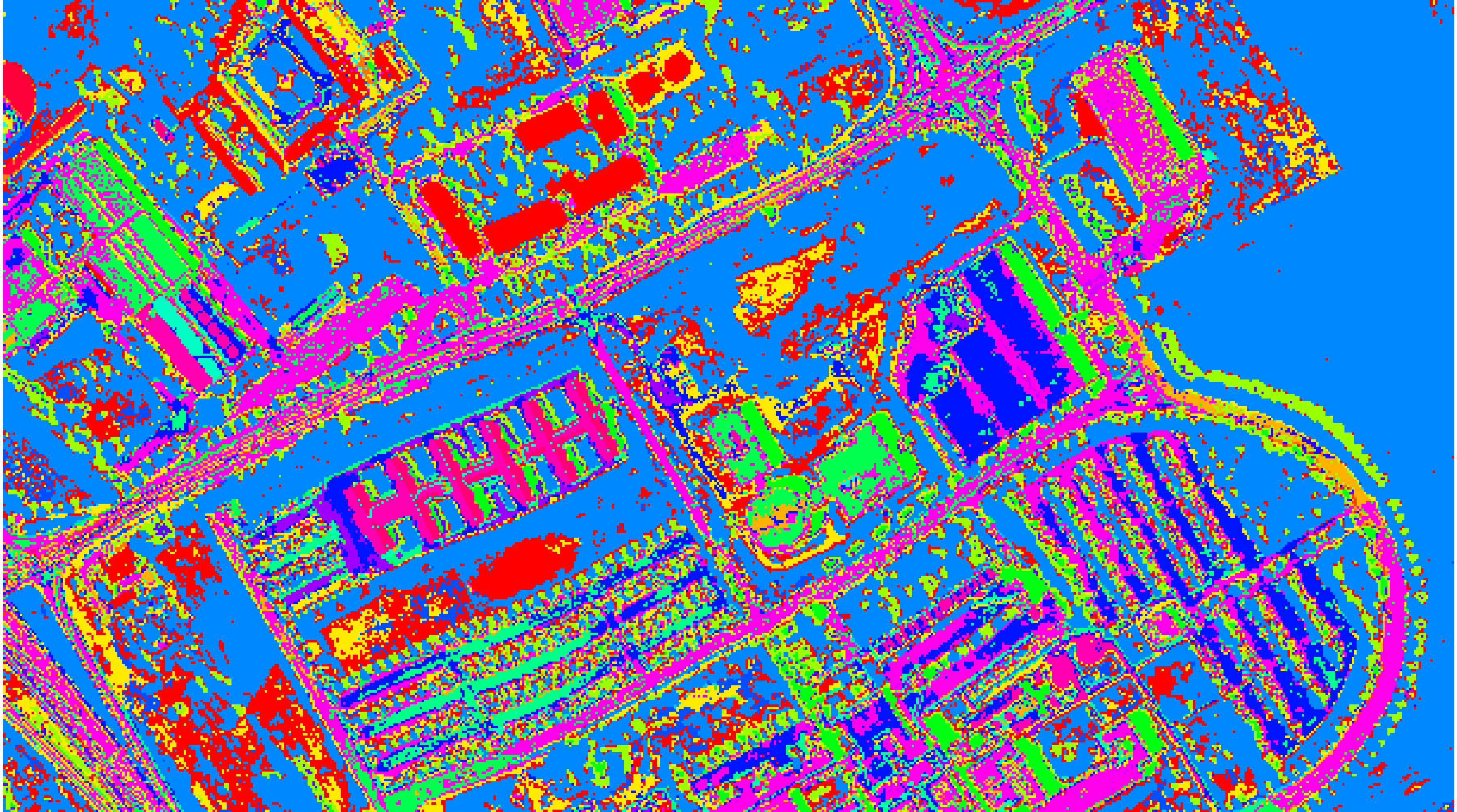}} \hfill
     \subfloat[][MRF]{\includegraphics{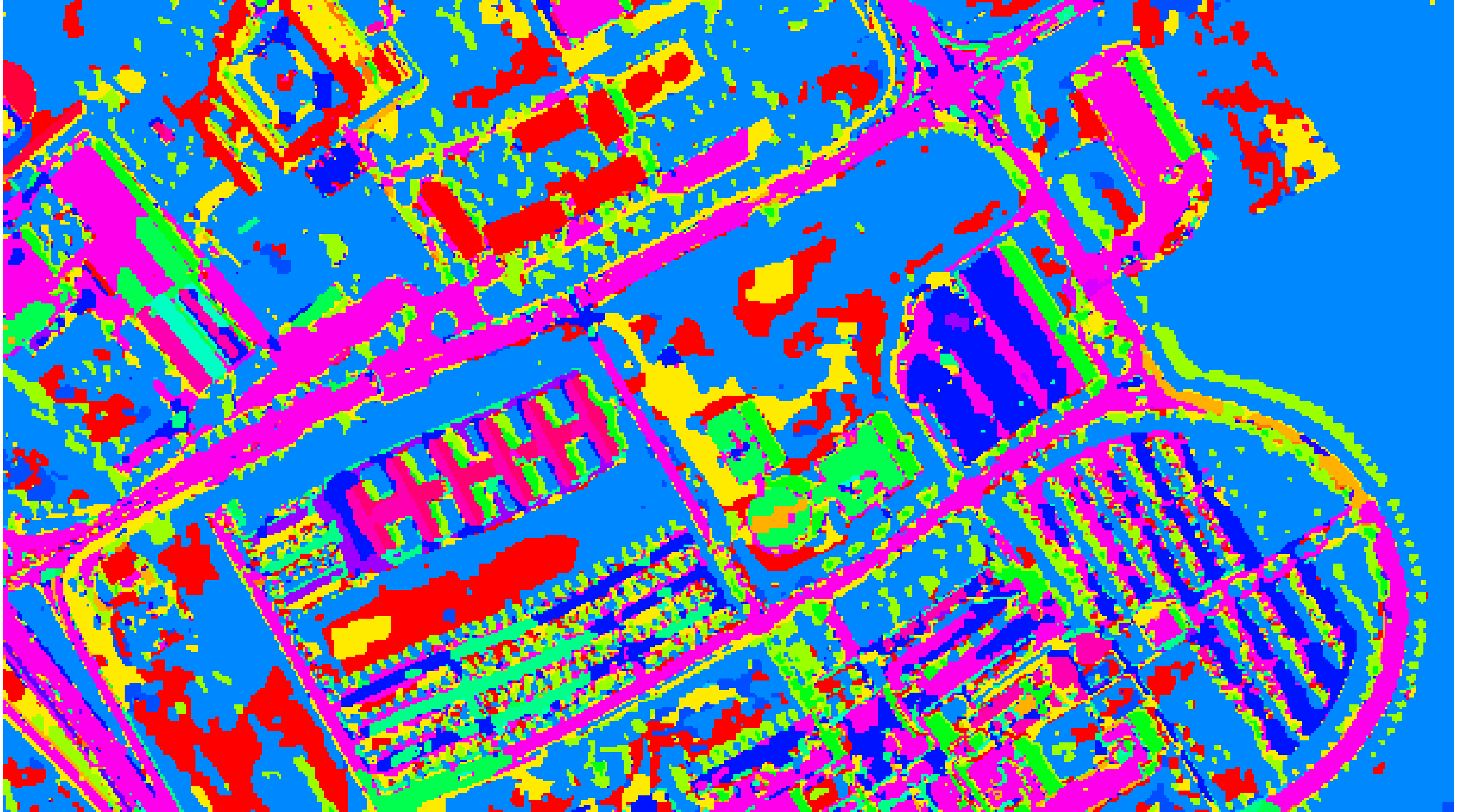}} \\
     \subfloat[][Major land covers]{\includegraphics{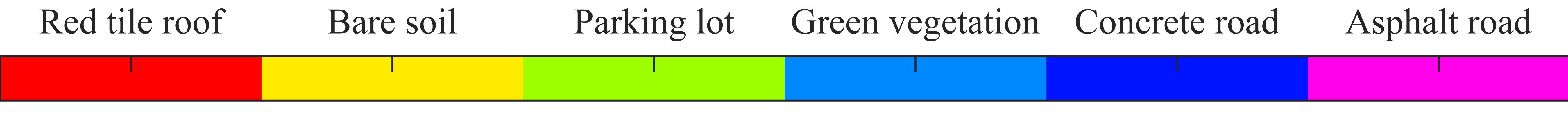}} \hfill
\caption{Land cover mapping using third-party spectral library.}
\label{fig:speclib_map}
\end{figure*}

There are cases where the use of spatial-spectral features is not possible for land cover mapping. If the training set consists of a third-party spectral library or ground spectra colleced from the scene, spatial-spectral features are not applicable for mapping. Same is the case when land cover maps are created by using physics bases models or unmixing techniques. In these cases, a principled approach to introduce spatial contexts is the use of UGMs. For example, the Santa Barbara urban spectral library~\cite{herold2003spectral} was used to map the land covers in the University of Pavia in Figure~\ref{fig:speclib_map}. The spectral library consists of 27 urban covers but the only land cover classes that occupied  more than 5\% of the pixel area in the predicted map have been included in the legend. The maps were generated by random forest classifier and the combination of random forest and MRF with $\beta=1$.

Table~\ref{tab:OATable_elaborate} compares the qualities of land cover maps generated by different methods using a variety of performance metrics used in remote sensing--overall accuracy, kappa coefficient, average precision, average recall, and average F1 score metrics. The training set size is set at 20 and 200 pixels per class and the classifier used is SVM with squared exponential kernel function. As with the previous results, the mean and the standard deviation of 30 trials is reported. The overall accuracy, a metric that is most widely used in remote sensing, measures the fraction of pixels that were correctly classified by the classifier. However, it fails to account if individual classes of materials are accurately classified. For each material class, precision measures the fraction of pixels that were classified to belong to a class that actually belonged to that class while recall measures the fraction of pixels that belonged to a class in the ground truth that were correctly labeled by the classifier to that class. Precision and recall are commonly referred as the user's accuracy and the producer's accuracy in the remote sensing literature. F1 score is the harmonic mean of precision and recall. For compactness, we have only included the class averaged values of these metrics in the table. Class averaged recall is sometimes also called average accuracy in literature. Since the number of pixels of each class in the testing set are equal in our experiments, the class averaged recall is equal to the overall accuracy. $\kappa$ coefficient is similar to overall accuracy and measures statistical agreement between the ground truth labels and the predicted labels.

\subsection{Superpixel-based pairwise MRFs}
\begin{figure}
\centering
\includegraphics[]{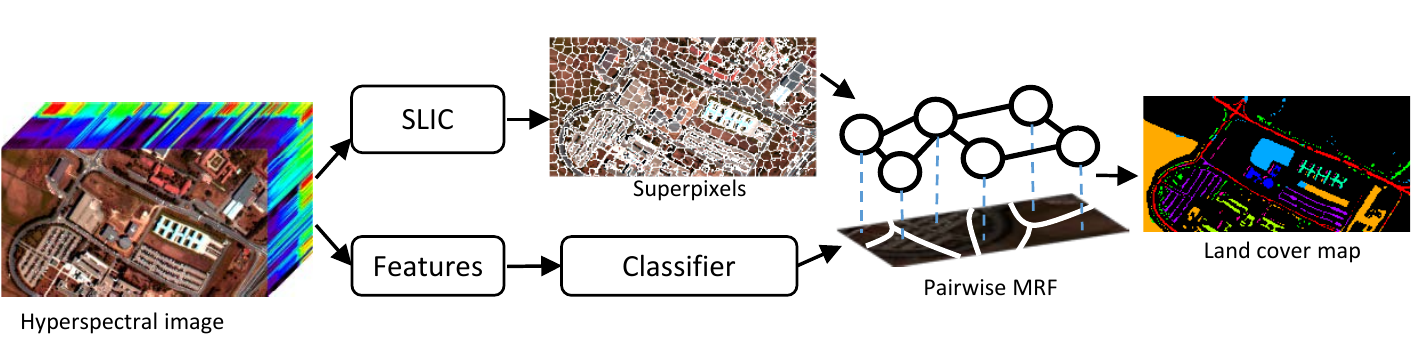}
\caption{Pipeline for superpixel-based pairwise models.}
\label{fig:pipeline_super}
\end{figure}
Even though efficient inference algorithms exists for grid-structured models, real-world aerial and satellite images can be enormous and grid-structured models might be slow for time-critical applications. The computational cost of inference is generally a function of the number of nodes in the graph. Therefore, for very large images it is wiser to group pixels into homogeneous regions, called superpixels~\cite{stutz2017superpixels} and use UGMs to model the distribution of superpixels' labels in order to reduce the total number of nodes. Superpixels are group of similar and connected pixels in the image. Unsupervised segmentation algorithms, such as simple linear iterative clustering (SLIC)~\cite{achanta2012slic}, can be used to decompose the image into superpixels. All the pixels belonging to a superpixel are assumed to have the same class label. This is a reasonable assumption because pixels which are connected and have similar spectra are likely to be of same material. However, if the size of the superpixels are too big, there is a high chance that some of the pixels in the superpixel will be of different classes. 

Larger superpixels leads to fewer total number of superpixels in the image and hence fewer nodes in the graph. This decreases the inference computational cost. On the other hand, larger superpixels decrease the resolution of the predicted map. Any object significantly smaller than the size of a superpixel will be missed. So there is a trade-off between the resolution of the predicted land cover map and the computational complexity, which the users can control. The average size of the superpixels can generally be controlled in the unsupervised segmentation algorithms. Apart from being faster, super-pixel based UGM can model higher level of relationship in the image than pixel-based models because they model relationships between different parts of the image having some semantic meaning rather than just pixels. 

Both MRFs and CRFs can be used with superpixels. However, in this section we experiment with only MRFs because as we saw in the previous section CRFs with log-linear potentials require large amount of training data and are better suited for cases where we have multiple training images. Figure~\ref{fig:pipeline_super} shows the workflow of the superpixel-based pairwise MRF used in the experiments. We experiment with raw spectrum and EMP features. Only SVM with squared exponential kernel is used as classifier in this experimentation as it was the one that performed the best previously. The SLIC algorithm was used to segment the images into superpixels. The SLIC algorithm works by assigning each pixel by a feature vector consisting of weighted concatenation of spatial coordinates plus the spectrum of the pixel and using a localized version of k-means clustering in this feature space to generate superpixels. There are three parameters of the SLIC algorithm. The first is called the regularizer and controls the shape of the superpixels. It was set to 100 in our experiments. The second parameter is the initial region size. It controls the final size and the total number of superpixels. 
The initial region size parameter was calculated in terms of desired number of superpixels in the image. The initial region size is equal to the square root of the total number of pixels in the image divided by the desired number of superpixels. In SLIC algorithm, the number of extracted superpixels is slightly different than the number of superpixels requested to the algorithm. We evaluate the performance of MRFs as a function of the number of superpixels in the experiments. The third parameter is the minimum region size, which was to 9 pixels in the experiments. We used the SLIC implementation in VLFeat library~\cite{vedaldi2010vlfeat} in the experiments. The spectrum was normalized to have zero mean and standard deviation of one at all the wavelengths before using with SLIC algorithm.

In superpixel-based MRF, each superpixel is a node of the undirected graph and there is an edge between the pair of nodes representing the superpixels which touch each other. So, depending upon the superpixel segmentation, the structure of the graph changes. In our experiments, the unary potential at each superpixel was calculated by averaging class probabilities of the pixels inside the superpixel. Averaging is somewhat equivalent to majority voting because if there are many pixels which likely belong to a particular class, the unary potential for the superpixel is higher for that class. If $s_1,s_2,...s_K$ are the $K$ superpixels, the unary energy at the node $s_i$ when it is assigned to class $c$ is given by $E_{s_i}\left(s_i=c\right) = - \ln \left( \frac{1}{N_{s_i}} \sum_{y_k \in s_i} P\left(  y_k=c \mid x_k \right) \right)$, where $N_{s_i}$ is the number of pixels in $s_i$. $x_k$ and $y_k$ are the input features and the labels of the pixels in $s_i$ respectively. The class probability at each pixel in the image is estimated by a pixel-wise SVM classifier. The pairwise energy function used is the Potts model (\ref{eq:potts}): $E_{ij}\left(s_i,s_j\right) = \beta  \, \mathcal{I}\left[ s_i \neq s_j\right]$, where $\mathcal{I}[.]$ is an indicator function and $\beta$ is a  parameter. The pairwise MRFs were implemented using UGM library~\cite{schmidt2012ugm}. For fair comparison, the grid-structured pixel-based pairwise models used as baseline in this section were also implemented using the same library. Inference was performed using graph cuts with alpha-expansion. 

\subsubsection{Results}
\begin{table*}[!htb]
\tiny
\caption{Performance of Superpixel-based MRF on the Indian pines dataset.}
\label{tab:sup_indian}
\centering	
\tabcolsep=0.09cm
\noindent\makebox[\textwidth]{%
\begin{tabular}{@{}llcccccc@{}} 
\toprule
Features & Superpixels & OA & $\kappa$ & Avg.Precision  & Avg. Recall & Avg. F1 & Time (secs)\\
\midrule
\multirow{9}{*}{Spectra} & Pixel-based MRF & 89.05$\pm$2.08 & 88.05$\pm$2.26 & 89.83$\pm$1.90 & 89.05$\pm$2.08 & 88.93$\pm$2.11 & 19.47$\pm$1.16\\
 & 50 & 74.12$\pm$2.82 & 71.77$\pm$3.07 & 72.86$\pm$3.36 & 74.12$\pm$2.82 & 71.92$\pm$3.06 & \textbf{11.15}$\pm$0.12\\
 & 100 & 79.17$\pm$2.48 & 77.27$\pm$2.70 & 82.10$\pm$1.86 & 79.17$\pm$2.48 & 78.76$\pm$2.59 & 11.17$\pm$0.12\\
 & 200 & 85.97$\pm$2.01 & 84.69$\pm$2.19 & 87.28$\pm$1.78 & 85.97$\pm$2.01 & 85.95$\pm$2.05 & 11.21$\pm$0.12\\
 & 400 & 89.13$\pm$1.97 & 88.15$\pm$2.15 & 90.11$\pm$1.69 & 89.13$\pm$1.97 & 89.05$\pm$2.00 & 11.27$\pm$0.12\\
 & 800 & \textbf{89.34}$\pm$1.91 & \textbf{88.37}$\pm$2.08 & \textbf{90.17}$\pm$1.76 & \textbf{89.34}$\pm$1.91 & \textbf{89.22}$\pm$1.96 & 11.38$\pm$0.12\\
 & 1600 & 88.36$\pm$2.27 & 87.30$\pm$2.48 & 89.32$\pm$2.21 & 88.36$\pm$2.27 & 88.20$\pm$2.33 & 11.53$\pm$0.14\\
 & 3200 & 86.41$\pm$1.95 & 85.17$\pm$2.13 & 87.28$\pm$1.89 & 86.41$\pm$1.95 & 86.22$\pm$2.03 & 11.59$\pm$0.16\\

\midrule
\multirow{9}{*}{EMP} & Pixel-based MRF & 95.91$\pm$1.27 & 95.53$\pm$1.39 & 96.11$\pm$1.17 & 95.91$\pm$1.27 & 95.89$\pm$1.27 & 374.95$\pm$2.89\\
 & 50 & 79.74$\pm$1.55 & 77.90$\pm$1.70 & 78.30$\pm$3.22 & 79.74$\pm$1.55 & 77.75$\pm$2.42 & \textbf{366.60}$\pm$2.31\\
 & 100 & 85.28$\pm$1.27 & 83.95$\pm$1.39 & 86.83$\pm$1.19 & 85.28$\pm$1.27 & 84.94$\pm$1.38 & 366.62$\pm$2.31\\
 & 200 & 91.81$\pm$1.14 & 91.07$\pm$1.24 & 92.27$\pm$1.06 & 91.81$\pm$1.14 & 91.78$\pm$1.16 & 366.65$\pm$2.31\\
 & 400 & 95.82$\pm$1.18 & 95.44$\pm$1.29 & 96.08$\pm$1.07 & 95.82$\pm$1.18 & 95.81$\pm$1.18 & 366.72$\pm$2.31\\
 & 800 & \textbf{96.06}$\pm$1.12 & \textbf{95.70}$\pm$1.22 & \textbf{96.23}$\pm$1.06 & \textbf{96.06}$\pm$1.12 & \textbf{96.04}$\pm$1.12 & 366.83$\pm$2.31\\
 & 1600 & 95.71$\pm$1.11 & 95.32$\pm$1.21 & 95.91$\pm$1.04 & 95.71$\pm$1.11 & 95.69$\pm$1.11 & 366.98$\pm$2.30\\
 & 3200 & 93.67$\pm$1.36 & 93.10$\pm$1.49 & 93.94$\pm$1.29 & 93.67$\pm$1.36 & 93.64$\pm$1.38 & 367.05$\pm$2.32\\

\bottomrule
\end{tabular}
}
\end{table*}
\begin{table*}[!htb]
\tiny
\caption{Performance of Superpixel-based MRF on the University of Pavia dataset.}
\label{tab:sup_paviaU}
\centering	
\tabcolsep=0.09cm
\noindent\makebox[\textwidth]{%
\begin{tabular}{@{}llcccccc@{}} 
\toprule
Features & Superpixels & OA & $\kappa$ & Avg.Precision  & Avg. Recall & Avg. F1 & Time (secs)\\
\midrule
\multirow{9}{*}{Spectra} & Pixel-based MRF & 91.23$\pm$4.06 & 90.13$\pm$4.57 & 91.69$\pm$3.92 & 91.23$\pm$4.06 & 91.16$\pm$4.10 & 69.58$\pm$0.86\\
 & 50 & 64.69$\pm$6.18 & 60.27$\pm$6.95 & 72.54$\pm$9.05 & 64.69$\pm$6.18 & 63.18$\pm$7.12 & \textbf{8.29}$\pm$0.71\\
 & 100 & 70.19$\pm$4.22 & 66.46$\pm$4.75 & 69.64$\pm$3.64 & 70.19$\pm$4.22 & 67.43$\pm$4.00 & 8.32$\pm$0.71\\
 & 200 & 79.69$\pm$5.03 & 77.15$\pm$5.66 & 83.90$\pm$3.75 & 79.69$\pm$5.03 & 79.78$\pm$4.82 & 8.37$\pm$0.71\\
 & 400 & 86.46$\pm$3.50 & 84.77$\pm$3.94 & 88.63$\pm$3.14 & 86.46$\pm$3.50 & 86.62$\pm$3.49 & 8.46$\pm$0.71\\
 & 800 & 89.80$\pm$3.31 & 88.53$\pm$3.72 & 90.98$\pm$2.81 & 89.80$\pm$3.31 & 89.83$\pm$3.30 & 8.66$\pm$0.71\\
 & 1600 & 91.56$\pm$3.87 & 90.51$\pm$4.35 & 92.40$\pm$3.56 & 91.56$\pm$3.87 & 91.58$\pm$3.84 & 9.05$\pm$0.72\\
 & 3200 & \textbf{92.83}$\pm$3.67 & \textbf{91.93}$\pm$4.13 & \textbf{93.43}$\pm$3.37 & \textbf{92.83}$\pm$3.67 & \textbf{92.76}$\pm$3.71 & 9.73$\pm$0.72\\
 & 6400 & 92.24$\pm$3.38 & 91.28$\pm$3.80 & 92.77$\pm$3.16 & 92.24$\pm$3.38 & 92.18$\pm$3.43 & 10.78$\pm$0.72\\
& 12800 & 91.26$\pm$3.90 & 90.17$\pm$4.39 & 92.00$\pm$3.61 & 91.26$\pm$3.90 & 91.20$\pm$3.98 & 14.03$\pm$0.73\\

\midrule
\multirow{9}{*}{EMP} & Pixel-based MRF & 95.81$\pm$1.49 & 95.29$\pm$1.68 & 96.04$\pm$1.39 & 95.81$\pm$1.49 & 95.83$\pm$1.48 & 321.66$\pm$0.87\\
 & 50 & 74.30$\pm$4.33 & 71.09$\pm$4.87 & 76.82$\pm$3.92 & 74.30$\pm$4.33 & 72.49$\pm$4.72 & \textbf{261.02}$\pm$0.95\\
 & 100 & 81.24$\pm$4.38 & 78.89$\pm$4.93 & 81.36$\pm$6.09 & 81.24$\pm$4.38 & 80.25$\pm$5.51 & 261.05$\pm$0.95\\
 & 200 & 88.21$\pm$2.75 & 86.74$\pm$3.10 & 89.24$\pm$2.18 & 88.21$\pm$2.75 & 87.87$\pm$2.74 & 261.10$\pm$0.95\\
 & 400 & 90.95$\pm$1.60 & 89.82$\pm$1.80 & 91.63$\pm$1.50 & 90.95$\pm$1.60 & 90.90$\pm$1.63 & 261.18$\pm$0.95\\
 & 800 & 94.67$\pm$1.67 & 94.01$\pm$1.88 & 95.08$\pm$1.42 & 94.67$\pm$1.67 & 94.69$\pm$1.66 & 261.39$\pm$0.95\\
 & 1600 & 95.01$\pm$1.41 & 94.38$\pm$1.59 & 95.38$\pm$1.30 & 95.01$\pm$1.41 & 95.03$\pm$1.41 & 261.78$\pm$0.95\\
 & 3200 & 95.79$\pm$1.44 & 95.27$\pm$1.61 & 96.02$\pm$1.37 & 95.79$\pm$1.44 & 95.80$\pm$1.44 & 262.51$\pm$0.91\\
 & 6400 & \textbf{96.09}$\pm$1.54 & \textbf{95.60}$\pm$1.73 & \textbf{96.31}$\pm$1.44 & \textbf{96.09}$\pm$1.54 & \textbf{96.10}$\pm$1.54 & 263.59$\pm$0.93\\
 & 12800 & 95.93$\pm$1.40 & 95.42$\pm$1.57 & 96.18$\pm$1.28 & 95.93$\pm$1.40 & 95.94$\pm$1.40 & 267.37$\pm$1.25\\

\bottomrule
\end{tabular}
}
\end{table*}
\begin{table*}[!htb]
\tiny
\caption{Performance of Superpixel-based MRF on the Pavia center dataset.}
\label{tab:sup_paviaC}
\centering	
\tabcolsep=0.09cm
\noindent\makebox[\textwidth]{%
\begin{tabular}{@{}llcccccc@{}} 
\toprule
Features & Superpixels & OA & $\kappa$ & Avg.Precision  & Avg. Recall & Avg. F1 & Time (secs)\\
\midrule
\multirow{9}{*}{Spectra} & Pixel-based MRF & 95.30$\pm$1.60 & 94.71$\pm$1.80 & 95.59$\pm$1.49 & 95.30$\pm$1.60 & 95.30$\pm$1.60 & 86.87$\pm$4.19\\
 & 50 & 61.90$\pm$1.97 & 57.14$\pm$2.22 & 62.72$\pm$3.29 & 61.90$\pm$1.97 & 57.83$\pm$2.39 & \textbf{20.36}$\pm$3.80\\
 & 100 & 75.59$\pm$1.85 & 72.54$\pm$2.08 & 74.24$\pm$3.66 & 75.59$\pm$1.85 & 73.16$\pm$2.50 & 20.43$\pm$3.80\\
 & 200 & 81.79$\pm$2.41 & 79.51$\pm$2.71 & 83.31$\pm$2.74 & 81.79$\pm$2.41 & 80.73$\pm$3.06 & 20.56$\pm$3.79\\
 & 400 & 84.59$\pm$1.51 & 82.67$\pm$1.70 & 86.16$\pm$1.55 & 84.59$\pm$1.51 & 84.17$\pm$1.69 & 20.81$\pm$3.79\\
 & 800 & 86.67$\pm$1.52 & 85.00$\pm$1.71 & 87.79$\pm$1.24 & 86.67$\pm$1.52 & 86.39$\pm$1.62 & 21.34$\pm$3.79\\
 & 1600 & 90.61$\pm$1.64 & 89.43$\pm$1.85 & 91.14$\pm$1.43 & 90.61$\pm$1.64 & 90.52$\pm$1.67 & 22.37$\pm$3.74\\
 & 3200 & 93.18$\pm$1.70 & 92.33$\pm$1.91 & 93.59$\pm$1.45 & 93.18$\pm$1.70 & 93.15$\pm$1.72 & 24.27$\pm$3.75\\
 & 6400 & 95.11$\pm$1.61 & 94.50$\pm$1.81 & 95.38$\pm$1.48 & 95.11$\pm$1.61 & 95.11$\pm$1.60 & 28.37$\pm$3.73\\
 & 12800 & \textbf{95.53}$\pm$1.38 & \textbf{94.97}$\pm$1.55 & \textbf{95.78}$\pm$1.29 & \textbf{95.53}$\pm$1.38 & \textbf{95.53}$\pm$1.37 & 35.62$\pm$3.74\\
 & 25600 & 95.40$\pm$1.59 & 94.83$\pm$1.79 & 95.68$\pm$1.45 & 95.40$\pm$1.59 & 95.40$\pm$1.58 & 48.02$\pm$3.87\\
 
\midrule
\multirow{9}{*}{EMP} & Pixel-based MRF & 96.00$\pm$1.44 & 95.50$\pm$1.62 & 96.14$\pm$1.40 & 96.00$\pm$1.44 & 96.00$\pm$1.43 & 823.60$\pm$3.55\\
 & 50 & 61.64$\pm$2.70 & 56.84$\pm$3.04 & 63.99$\pm$3.23 & 61.64$\pm$2.70 & 57.58$\pm$2.93 & \textbf{759.85}$\pm$3.68\\
 & 100 & 77.07$\pm$2.85 & 74.21$\pm$3.20 & 77.39$\pm$2.93 & 77.07$\pm$2.85 & 75.84$\pm$3.21 & 759.93$\pm$3.68\\
 & 200 & 82.56$\pm$2.65 & 80.38$\pm$2.98 & 84.57$\pm$2.32 & 82.56$\pm$2.65 & 81.87$\pm$3.16 & 760.05$\pm$3.68\\
 & 400 & 86.36$\pm$2.04 & 84.65$\pm$2.29 & 87.63$\pm$1.73 & 86.36$\pm$2.04 & 86.14$\pm$2.19 & 760.30$\pm$3.68\\
 & 800 & 89.14$\pm$1.86 & 87.78$\pm$2.09 & 90.00$\pm$1.51 & 89.14$\pm$1.86 & 88.91$\pm$1.96 & 760.83$\pm$3.68\\
 & 1600 & 92.30$\pm$1.59 & 91.34$\pm$1.79 & 92.72$\pm$1.44 & 92.30$\pm$1.59 & 92.23$\pm$1.63 & 761.84$\pm$3.66\\
 & 3200 & 94.62$\pm$1.67 & 93.95$\pm$1.88 & 94.87$\pm$1.56 & 94.62$\pm$1.67 & 94.60$\pm$1.70 & 763.78$\pm$3.72\\
 & 6400 & 95.93$\pm$1.58 & 95.42$\pm$1.78 & 96.10$\pm$1.51 & 95.93$\pm$1.58 & 95.93$\pm$1.59 & 767.98$\pm$3.70\\
 & 12800 & \textbf{96.05}$\pm$1.48 & \textbf{95.56}$\pm$1.67 & \textbf{96.21}$\pm$1.42 & \textbf{96.05}$\pm$1.48 & \textbf{96.05}$\pm$1.48 & 773.64$\pm$3.16\\
 & 25600 & 95.98$\pm$1.73 & 95.47$\pm$1.95 & 96.12$\pm$1.70 & 95.98$\pm$1.73 & 95.98$\pm$1.73 & 787.12$\pm$3.51\\
\bottomrule
\end{tabular}
}
\end{table*}
\begin{table*}[!htb]
\tiny
\caption{Performance of Superpixel-based MRF on the Salinas dataset.}
\label{tab:sup_salinas}
\centering	
\tabcolsep=0.09cm
\noindent\makebox[\textwidth]{%
\begin{tabular}{@{}llcccccc@{}} 
\toprule
Features & Superpixels & OA & $\kappa$ & Avg.Precision  & Avg. Recall & Avg. F1 & Time (secs)\\
\midrule
\multirow{9}{*}{Spectra} & Pixel-based MRF & 97.42$\pm$1.37 & 97.24$\pm$1.46 & 97.53$\pm$1.63 & 97.42$\pm$1.37 & 97.34$\pm$1.60 & 101.13$\pm$2.02\\
 & 50 & 73.26$\pm$2.19 & 71.48$\pm$2.33 & 67.21$\pm$4.33 & 73.26$\pm$2.19 & 67.95$\pm$3.39 & \textbf{31.85}$\pm$1.99\\
 & 100 & 79.59$\pm$1.59 & 78.23$\pm$1.69 & 81.03$\pm$2.23 & 79.59$\pm$1.59 & 76.39$\pm$2.03 & 31.90$\pm$1.99\\
 & 200 & 89.31$\pm$1.83 & 88.60$\pm$1.95 & 91.11$\pm$1.78 & 89.31$\pm$1.83 & 89.24$\pm$2.12 & 32.00$\pm$1.99\\
 & 400 & 96.84$\pm$1.29 & 96.63$\pm$1.38 & 96.95$\pm$1.58 & 96.84$\pm$1.29 & 96.72$\pm$1.61 & 32.17$\pm$1.99\\
 & 800 & 97.19$\pm$1.42 & 97.00$\pm$1.51 & 97.22$\pm$1.75 & 97.19$\pm$1.42 & 97.10$\pm$1.69 & 32.61$\pm$1.96\\
 & 1600 & 96.68$\pm$1.41 & 96.46$\pm$1.51 & 96.86$\pm$1.64 & 96.68$\pm$1.41 & 96.57$\pm$1.67 & 33.78$\pm$1.94\\
 & 3200 & 97.67$\pm$1.54 & 97.52$\pm$1.64 & 97.75$\pm$1.79 & 97.67$\pm$1.54 & 97.59$\pm$1.79 & 35.96$\pm$1.89\\
 & 6400 & \textbf{98.04}$\pm$1.59 & \textbf{97.91}$\pm$1.70 & \textbf{98.07}$\pm$1.88 & \textbf{98.04}$\pm$1.59 & \textbf{97.95}$\pm$1.85 & 43.61$\pm$1.94\\
 & 12800 & 96.58$\pm$1.75 & 96.35$\pm$1.87 & 96.74$\pm$1.97 & 96.58$\pm$1.75 & 96.46$\pm$2.03 & 44.49$\pm$1.96\\

\midrule
\multirow{9}{*}{EMP} & Pixel-based MRF & 98.78$\pm$0.78 & 98.69$\pm$0.83 & 98.83$\pm$0.74 & 98.78$\pm$0.78 & 98.77$\pm$0.78 & 481.14$\pm$2.18\\
 & 50 & 74.76$\pm$1.13 & 73.08$\pm$1.21 & 66.71$\pm$2.48 & 74.76$\pm$1.13 & 68.91$\pm$1.76 & \textbf{412.69}$\pm$2.20\\
 & 100 & 81.07$\pm$1.06 & 79.81$\pm$1.13 & 82.24$\pm$1.49 & 81.07$\pm$1.06 & 78.35$\pm$1.26 & 412.74$\pm$2.20\\
 & 200 & 90.89$\pm$0.87 & 90.28$\pm$0.93 & 91.99$\pm$0.80 & 90.89$\pm$0.87 & 90.88$\pm$0.88 & 412.84$\pm$2.20\\
 & 400 & 98.11$\pm$0.49 & 97.99$\pm$0.53 & 98.20$\pm$0.46 & 98.11$\pm$0.49 & 98.11$\pm$0.50 & 413.01$\pm$2.20\\
 & 800 & 98.62$\pm$0.63 & 98.52$\pm$0.67 & 98.70$\pm$0.56 & 98.62$\pm$0.63 & 98.62$\pm$0.63 & 413.39$\pm$2.20\\
 & 1600 & 98.54$\pm$0.56 & 98.44$\pm$0.59 & 98.62$\pm$0.53 & 98.54$\pm$0.56 & 98.54$\pm$0.56 & 414.60$\pm$2.23\\
 & 3200 & 98.91$\pm$0.52 & 98.84$\pm$0.55 & 98.96$\pm$0.49 & 98.91$\pm$0.52 & 98.91$\pm$0.51 & 417.13$\pm$2.25\\
 & 6400 & \textbf{98.93}$\pm$0.54 & \textbf{98.86}$\pm$0.57 & \textbf{98.98}$\pm$0.51 & \textbf{98.93}$\pm$0.54 & \textbf{98.93}$\pm$0.54 & 422.66$\pm$2.78\\
 & 12800 & 98.03$\pm$0.58 & 97.89$\pm$0.61 & 98.10$\pm$0.56 & 98.03$\pm$0.58 & 98.03$\pm$0.58 & 424.44$\pm$2.33\\
\bottomrule
\end{tabular}
}
\end{table*}

Table~\ref{tab:sup_indian}, Table~\ref{tab:sup_paviaU}, Table~\ref{tab:sup_paviaC}, and Table~\ref{tab:sup_salinas} compare the performance of superpixel-based MRF to the grid-structured pixel-based MRF (referred as pixel-based MRF). The number of superpixels used is varied.  The SVM classifier were trained using 50 ground truth pixels per class. The hyperparameters of the SVM and the MRFs were tuned using grid search similarly to previous experiments. The mean and the standard deviation of the performance metrics computed over 30 independent random trials are reported. 

\begin{figure*}[!h]
\centering
	 \subfloat[][RGB image]{\includegraphics{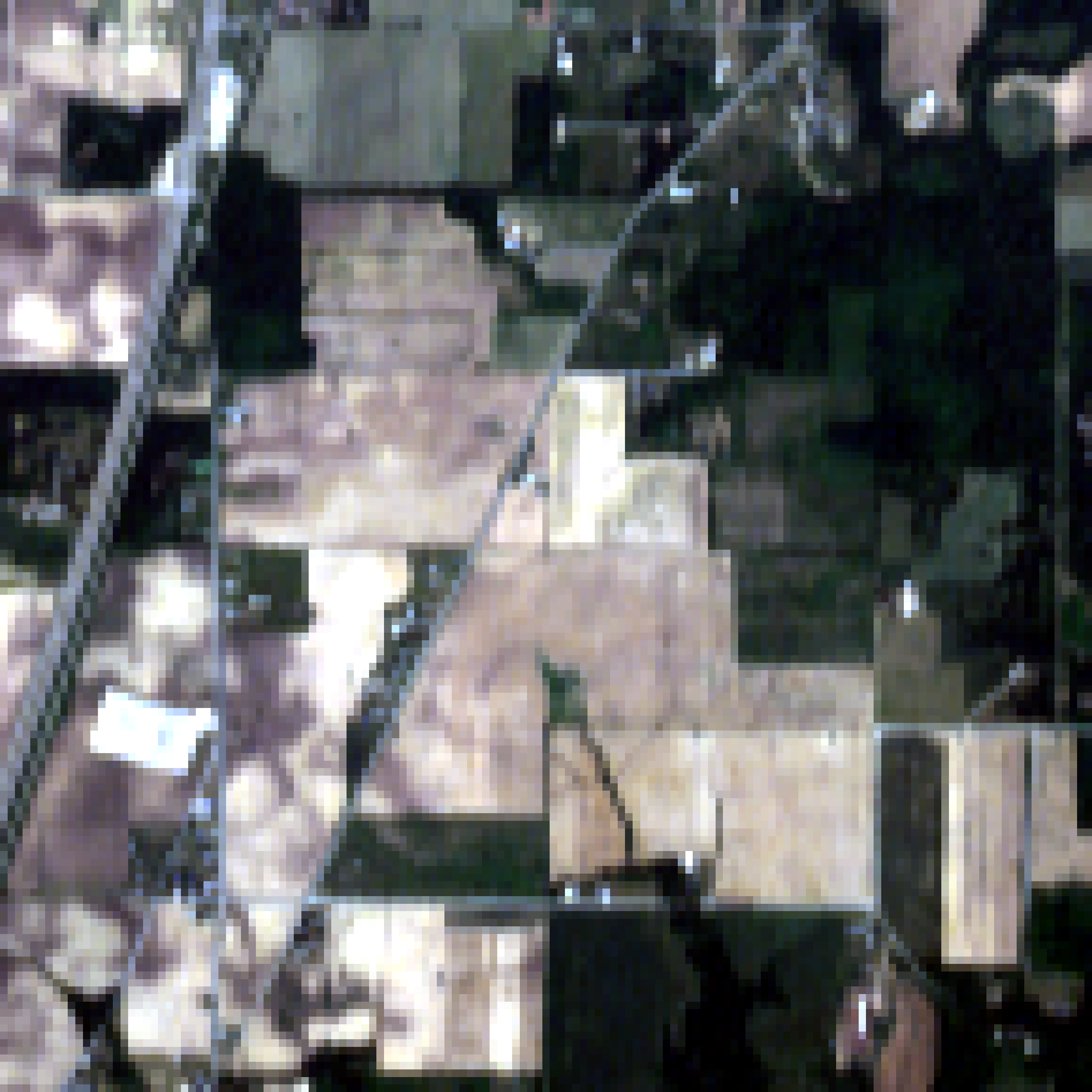}} \hfill
	 \subfloat[][Ground truth]{\includegraphics{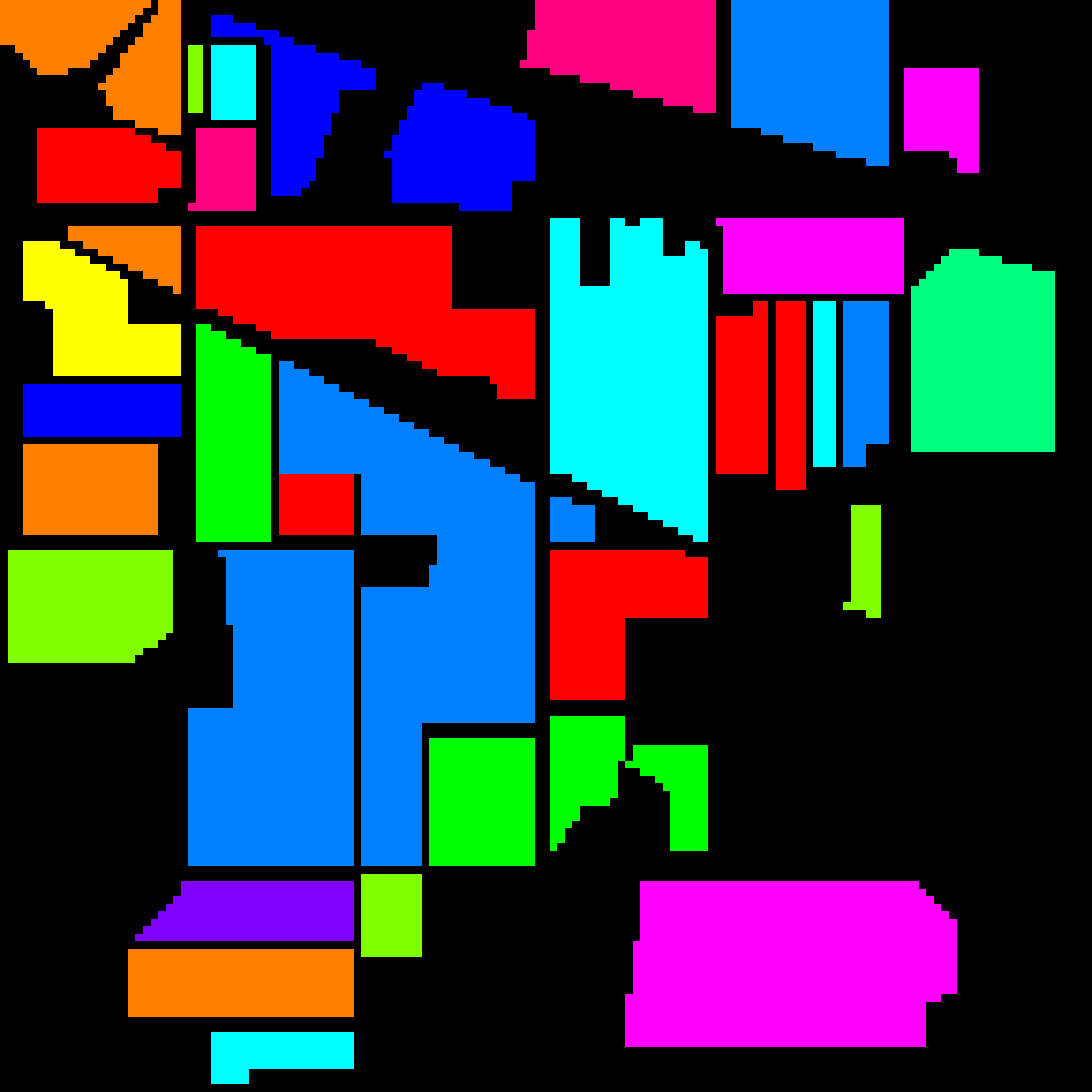}} \hfill
     \subfloat[][SVM]{\includegraphics{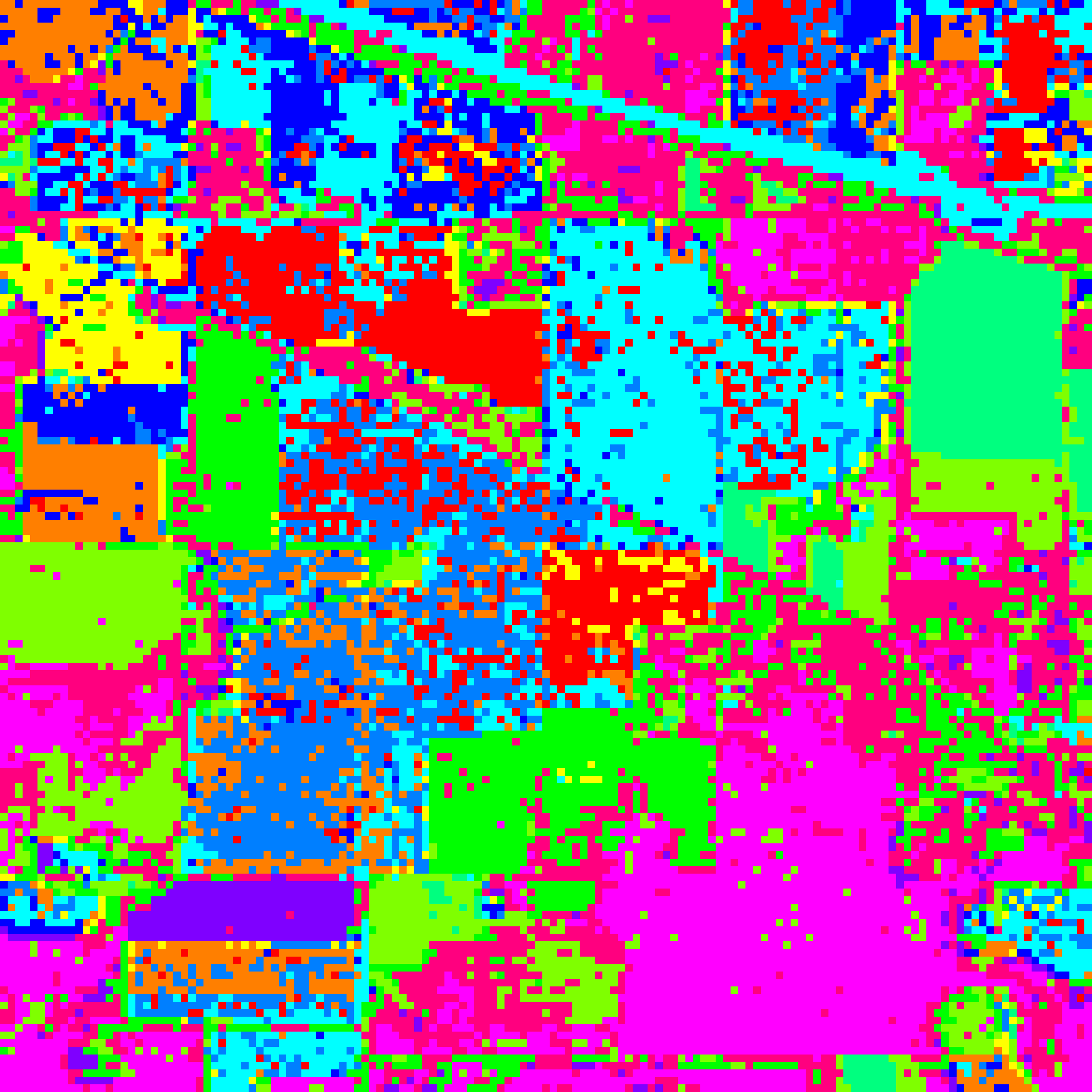}} \hfill
     \subfloat[][Pixel-based MRF]{\includegraphics{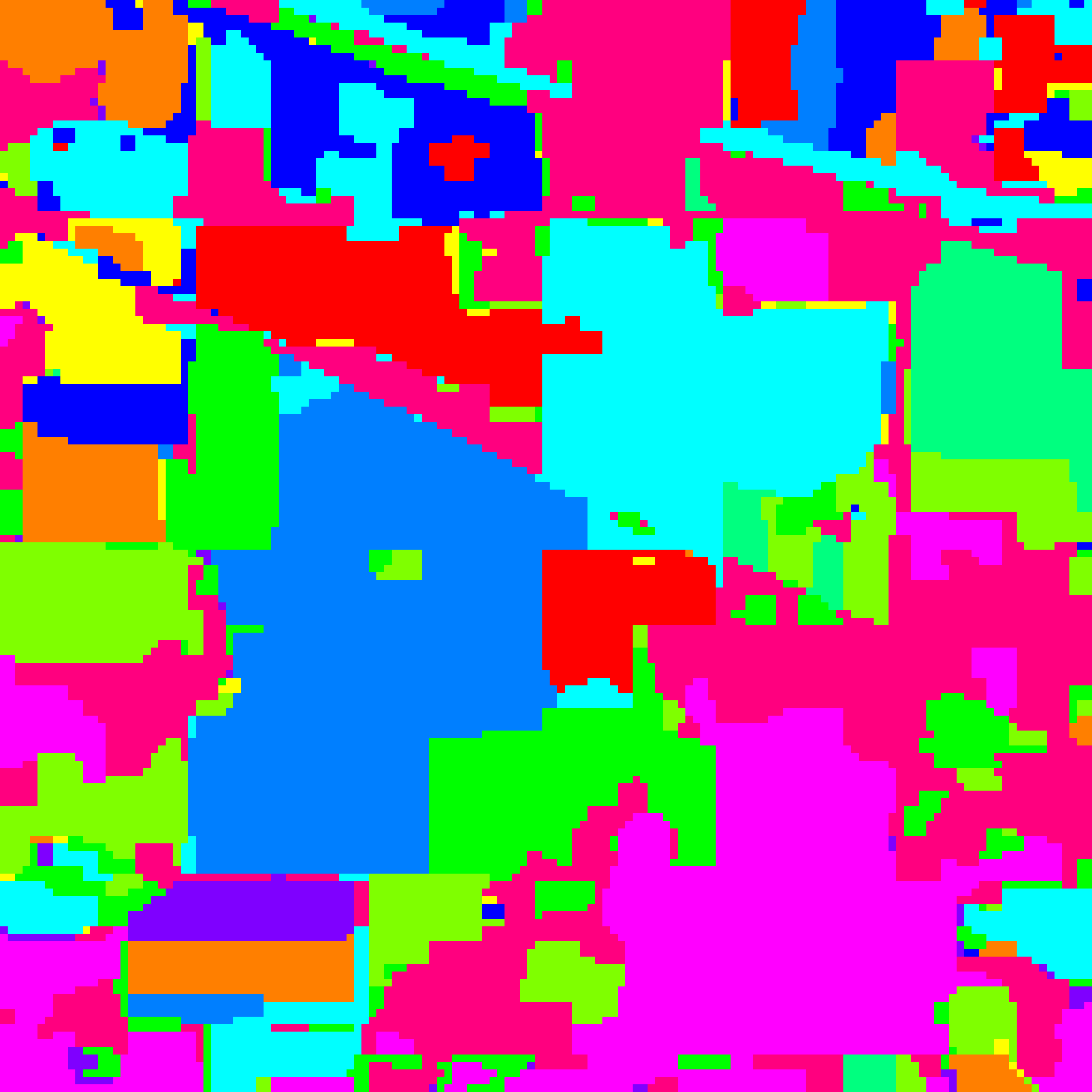}} \hfill
     \subfloat[][50 superpixels]{\includegraphics{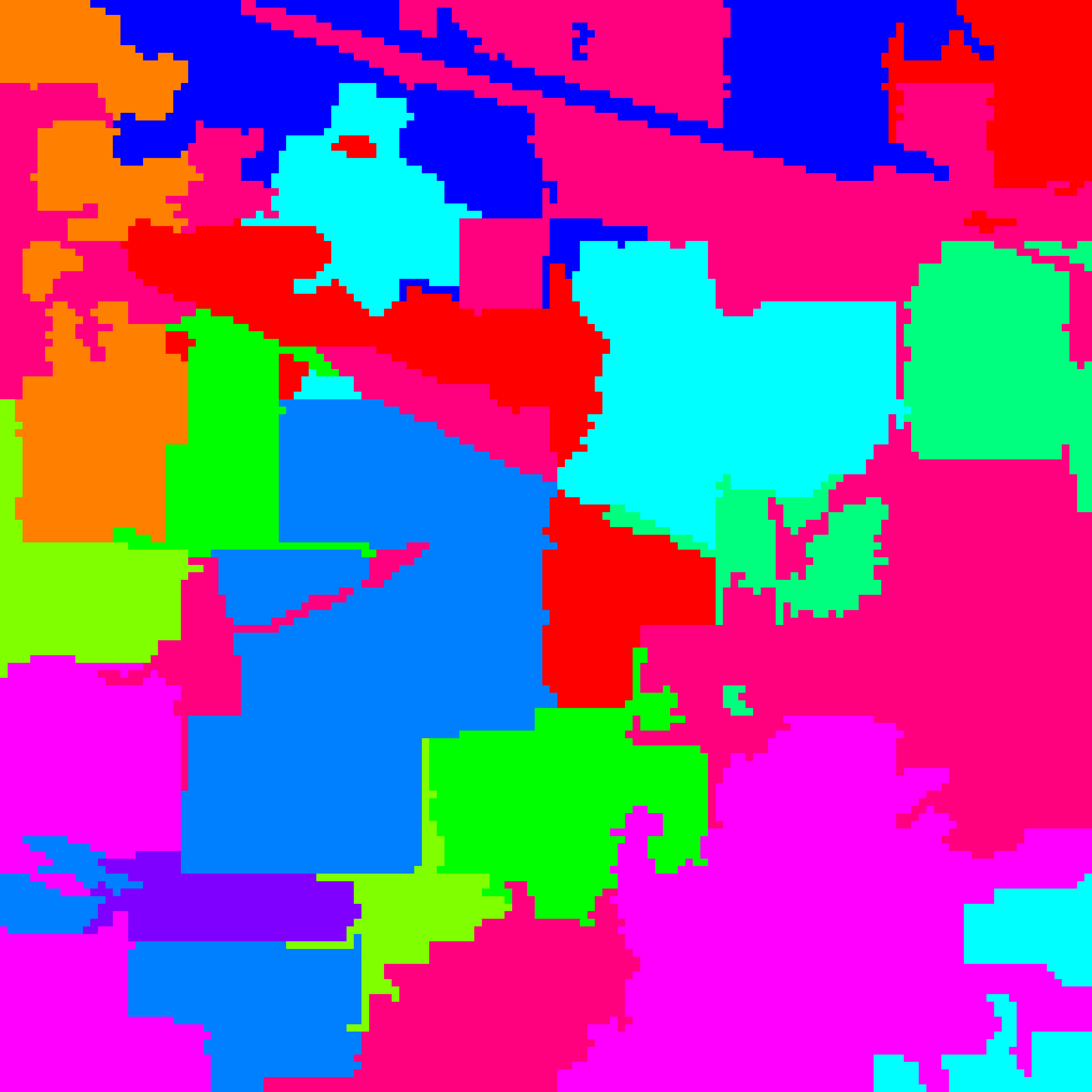}} \hfill
     \subfloat[][100 superpixels]{\includegraphics{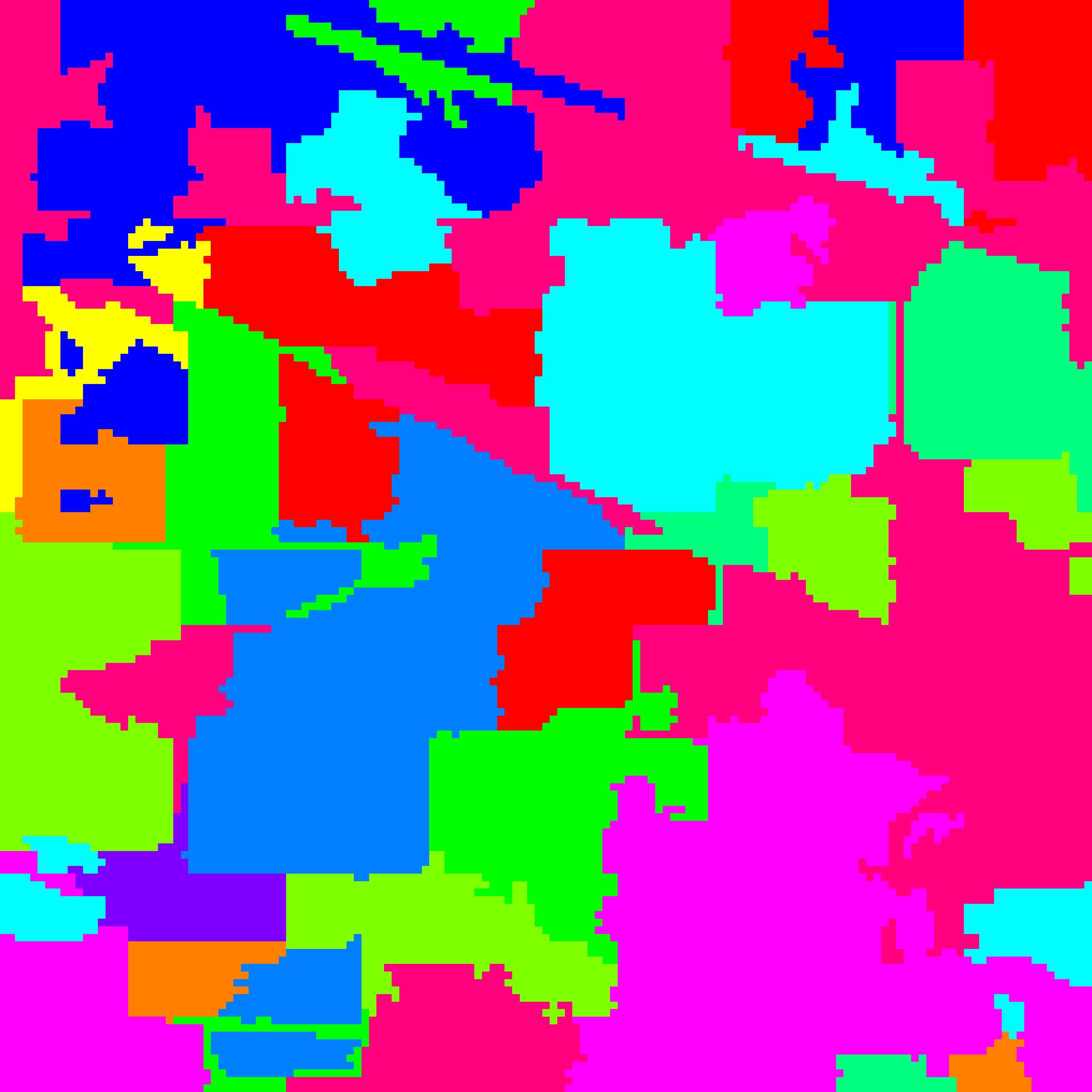}} \hfill
     \subfloat[][200 superpixels]{\includegraphics{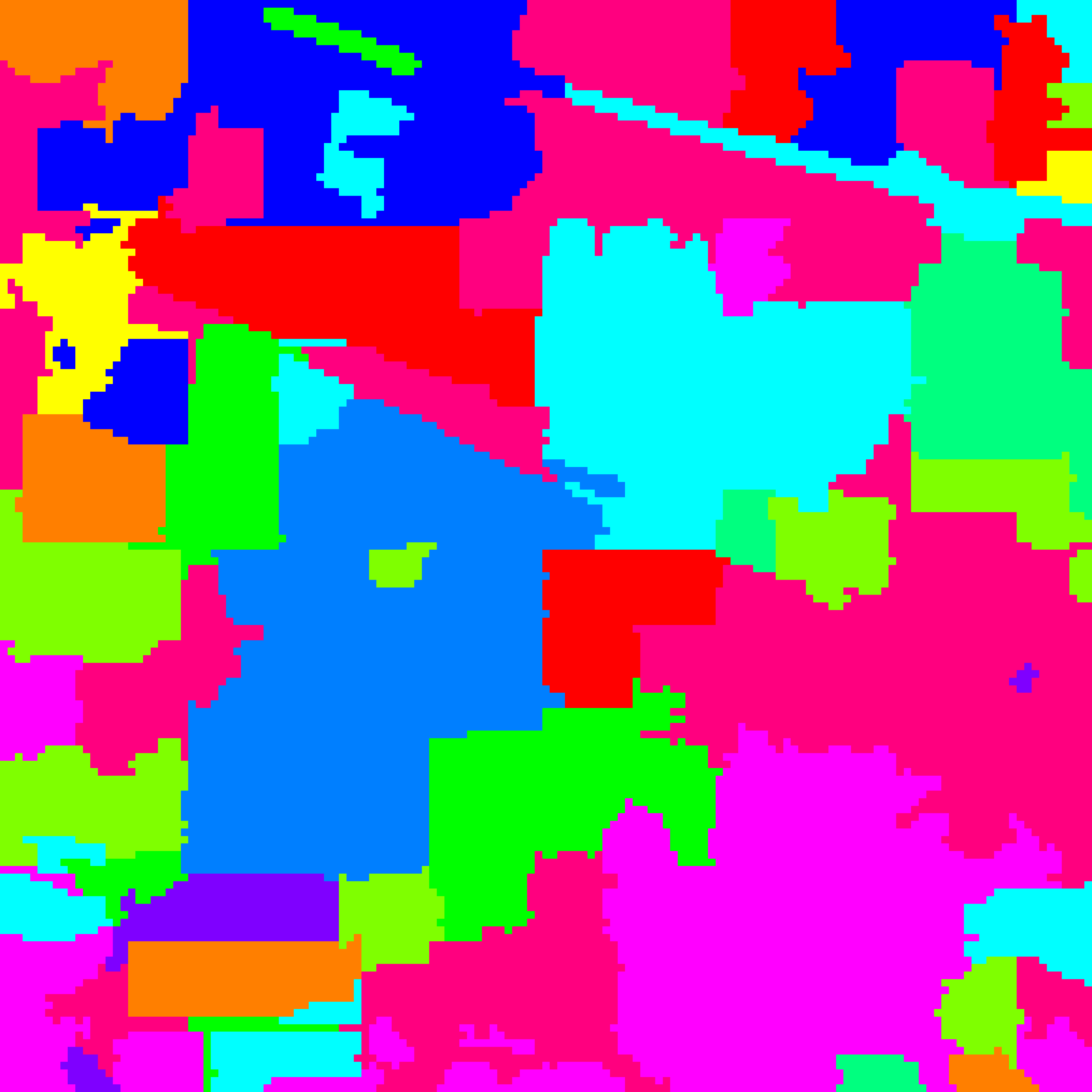}} \hfill
     \subfloat[][400 superpixels]{\includegraphics{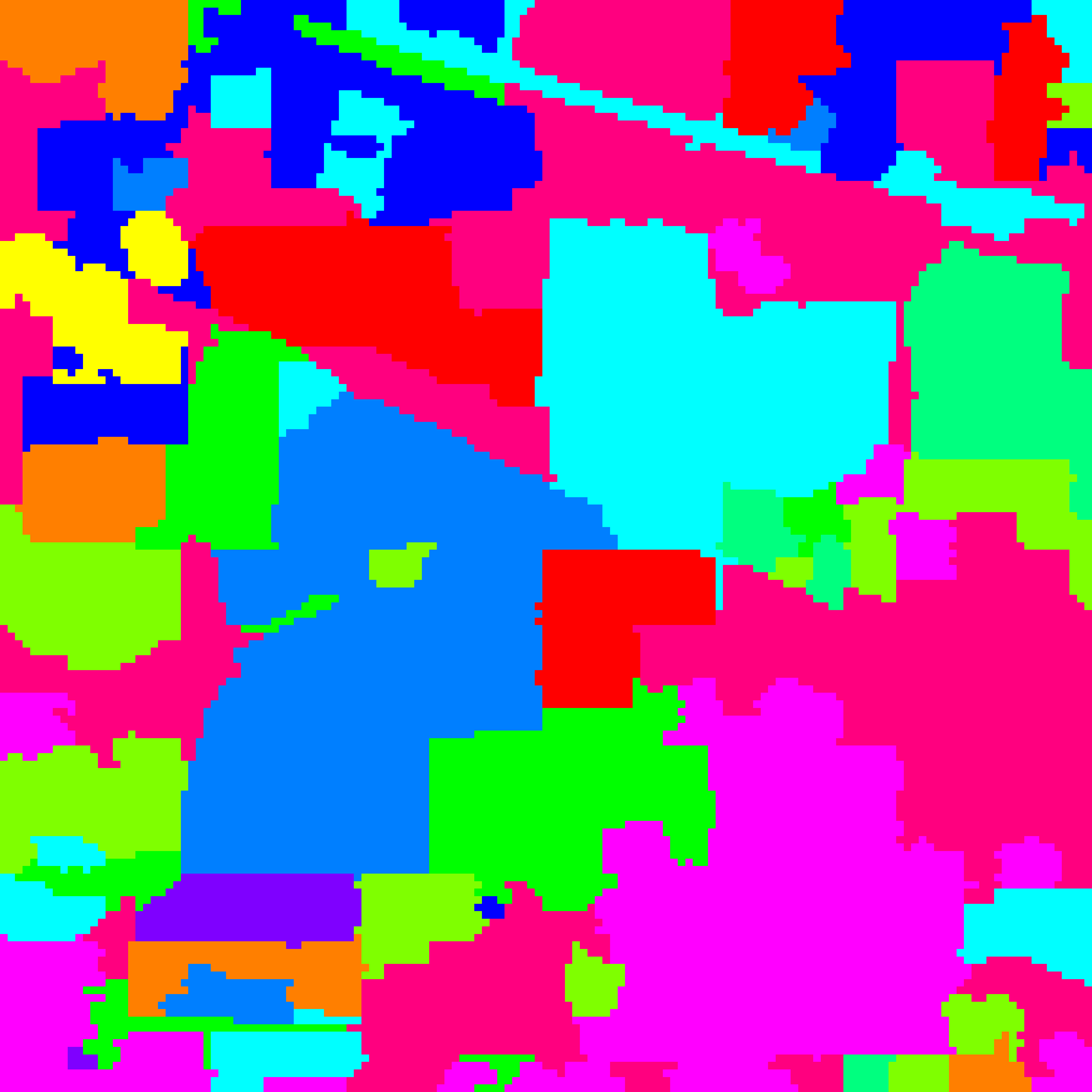}} \hfill
     \subfloat[][800 superpixels]{\includegraphics{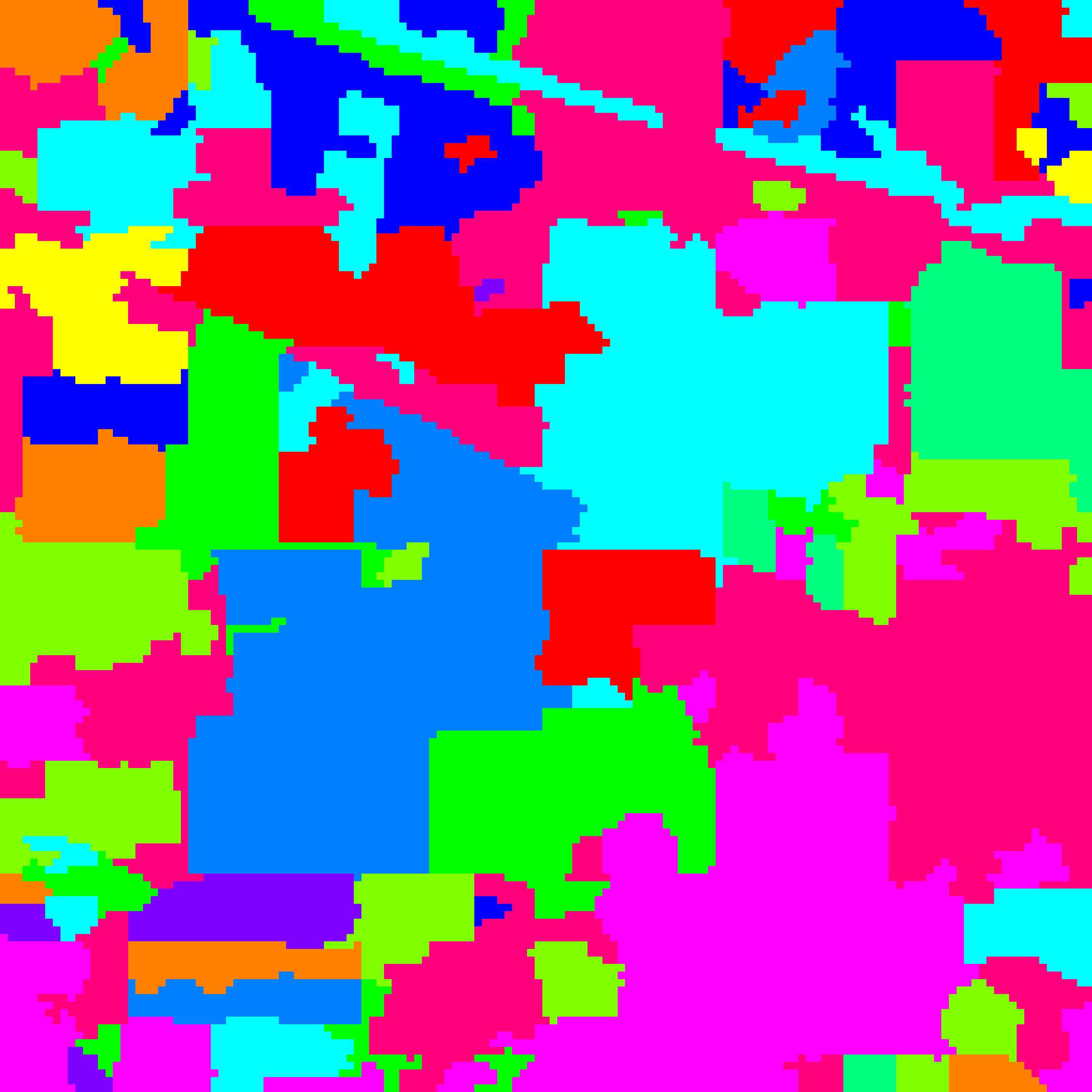}} \hfill
     \subfloat[][1600 superpixels]{\includegraphics{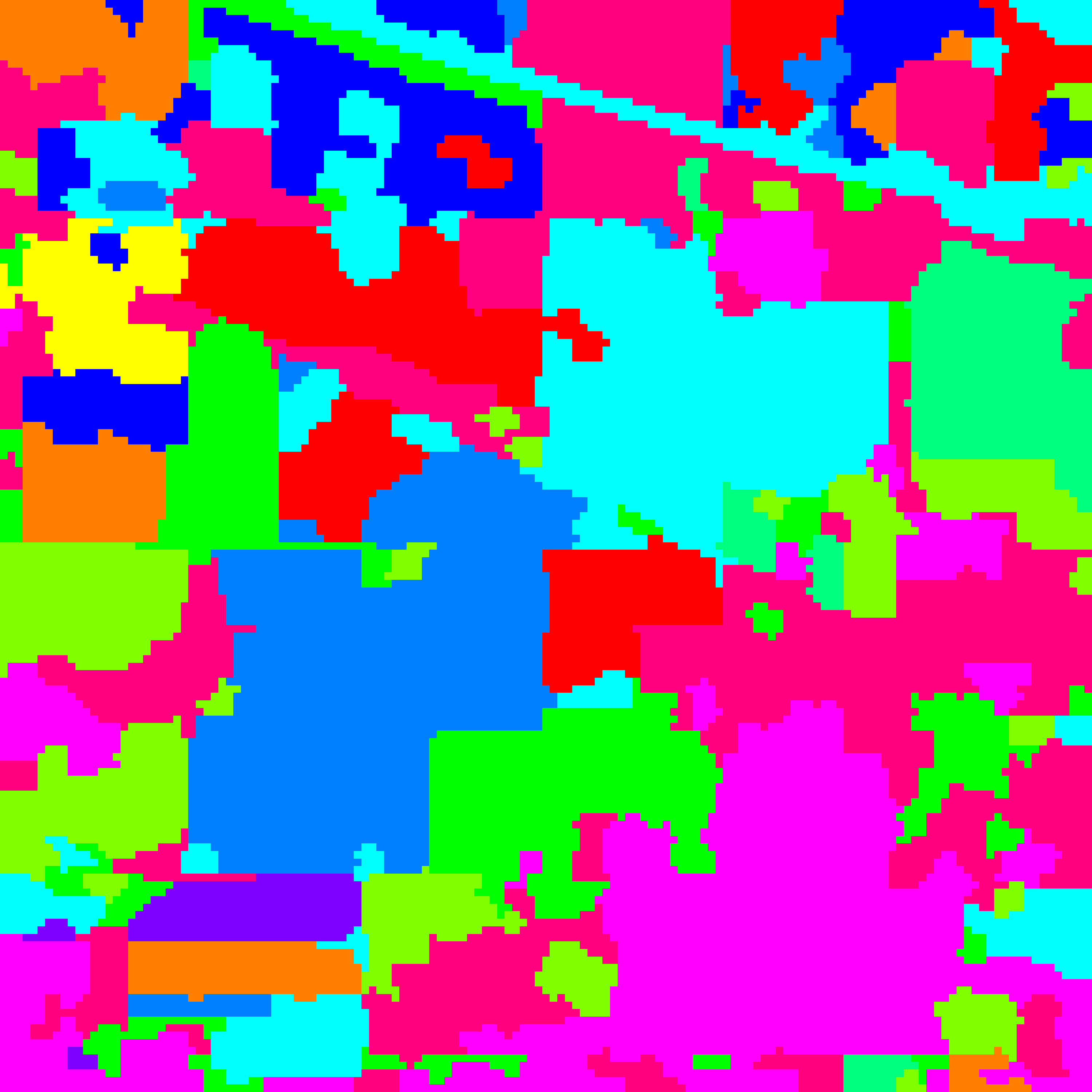}} \hfill
     \subfloat[][3200 superpixels]{\includegraphics{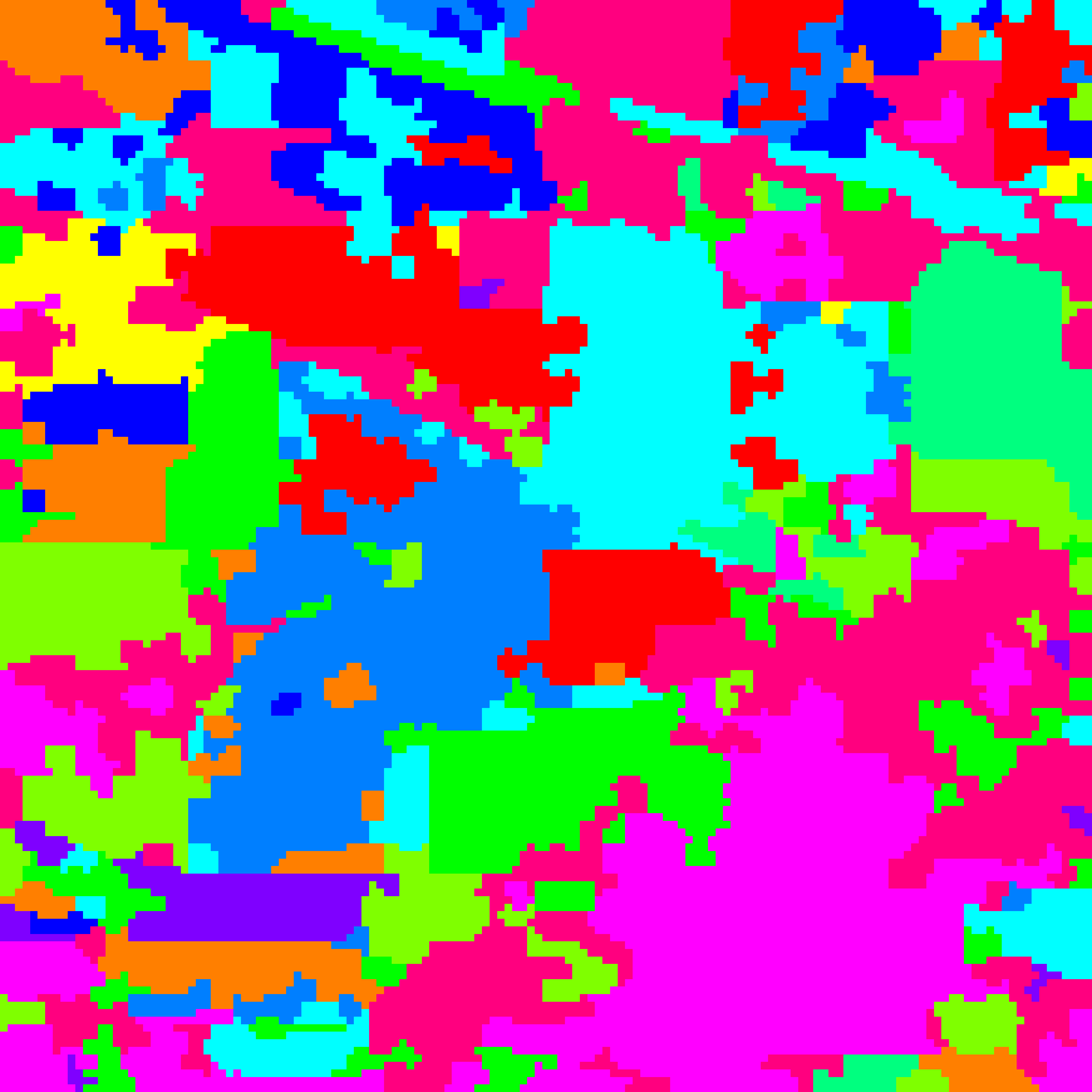}} \hfill
     
\caption{Superpixel-based MRF applied on the Indian pines image.}
\label{fig:sup_pix_map}
\end{figure*}
\begin{figure*}[!h]
\centering
	 \subfloat[][RGB image]{\includegraphics{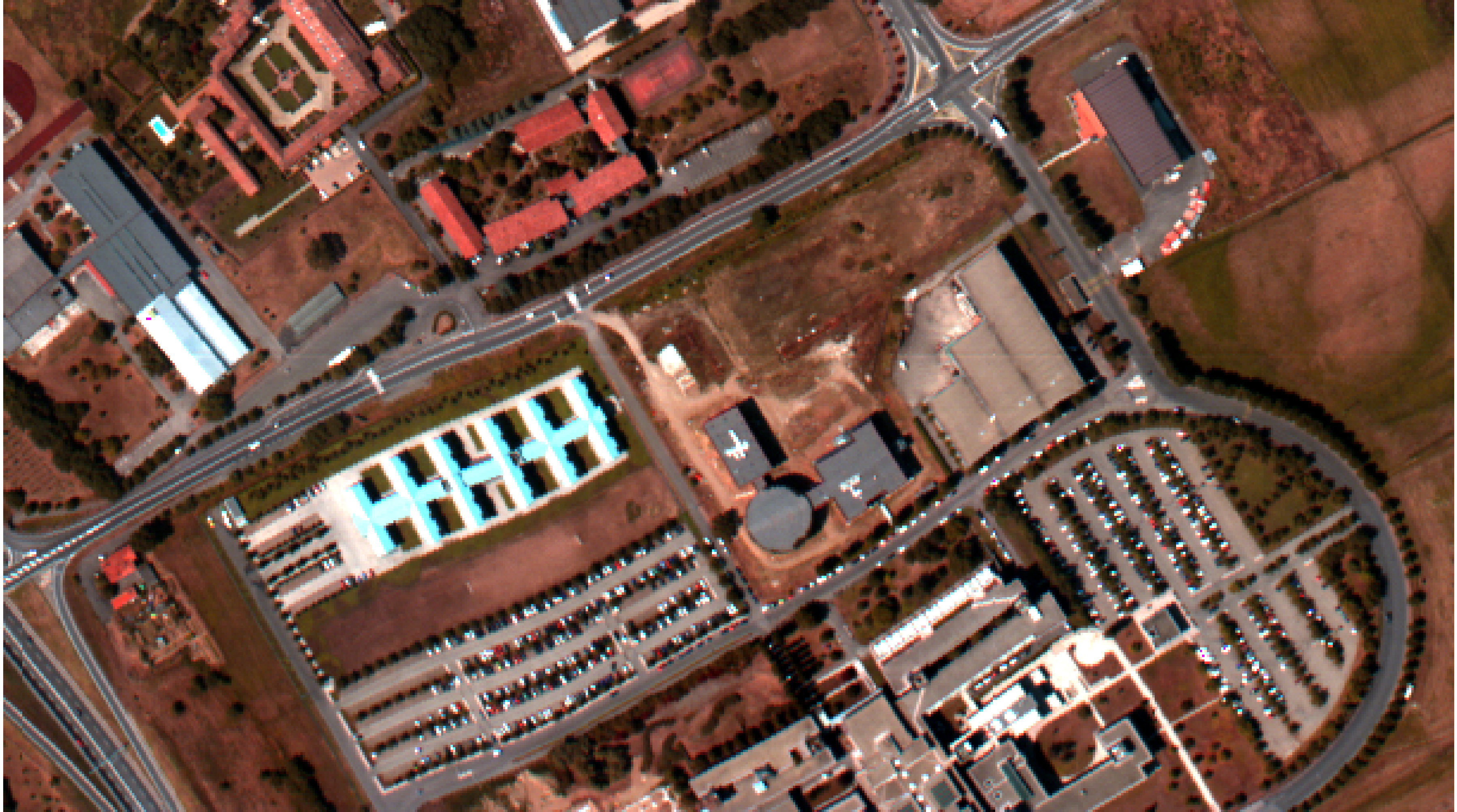}} \hfill
	 \subfloat[][Ground truth]{\includegraphics{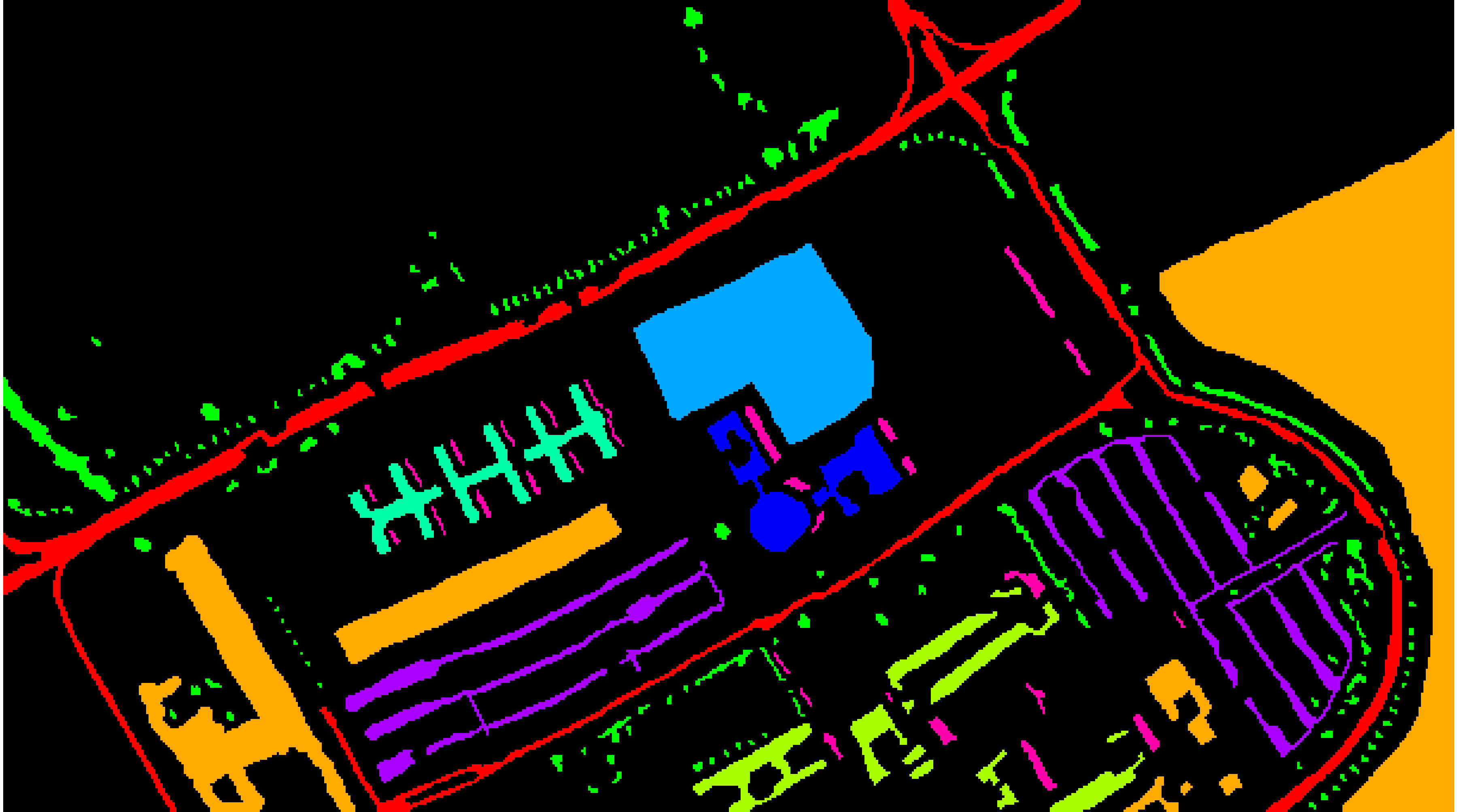}} \hfill
     \subfloat[][SVM]{\includegraphics{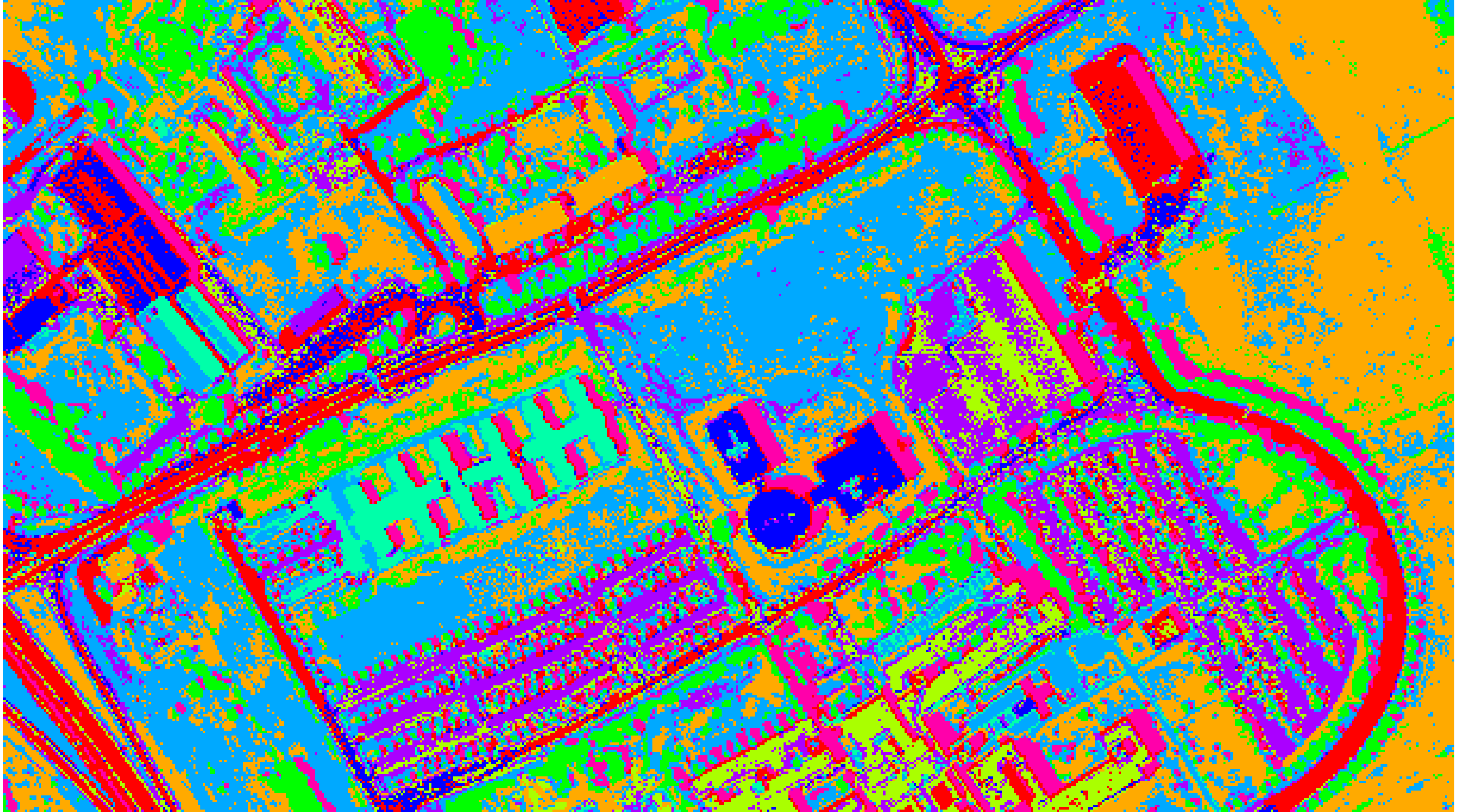}} \hfill
     \subfloat[][Pixel-based MRF]{\includegraphics{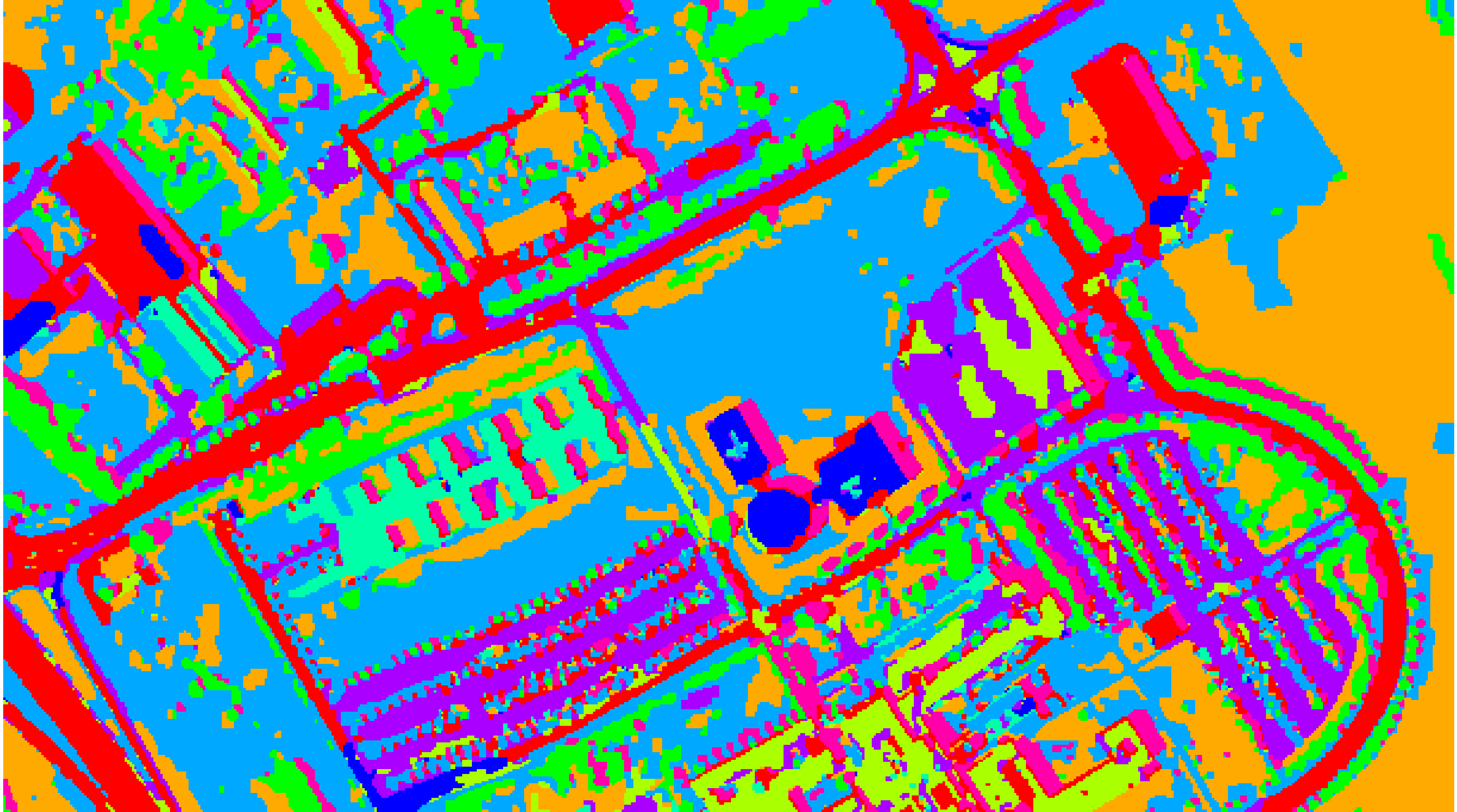}} \hfill
     \subfloat[][50 superpixels]{\includegraphics{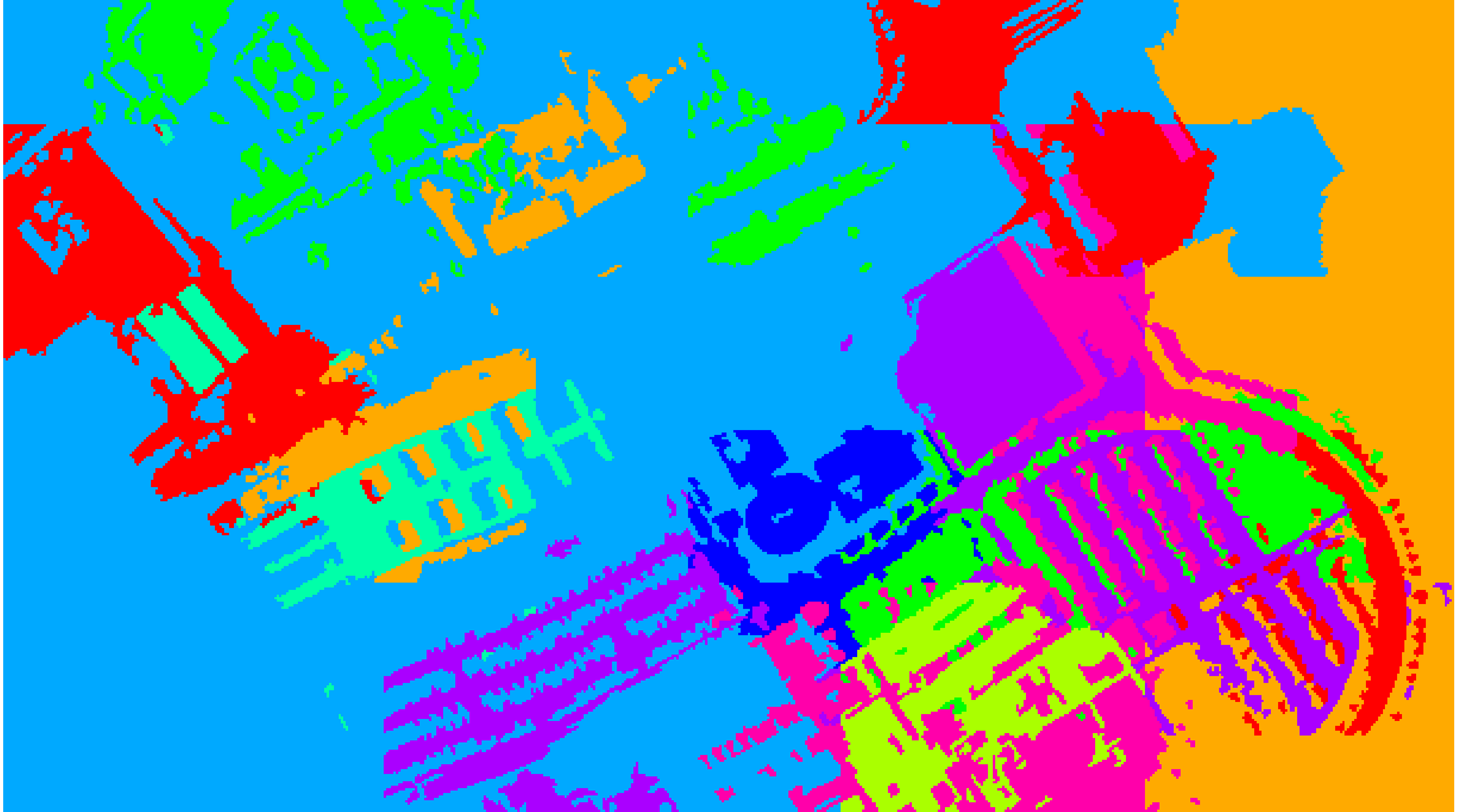}} \hfill
     \subfloat[][100 superpixels]{\includegraphics{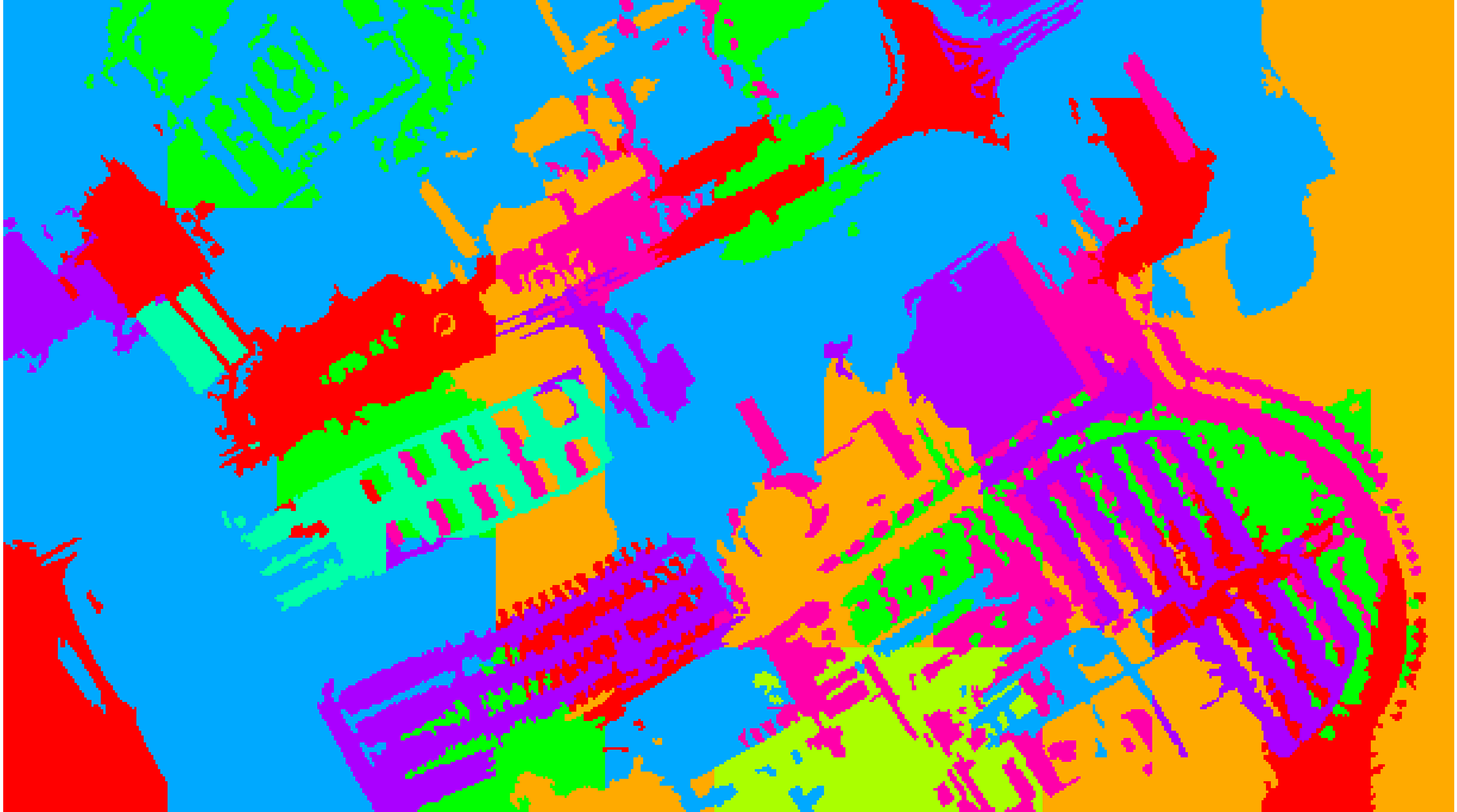}} \hfill
     \subfloat[][200 superpixels]{\includegraphics{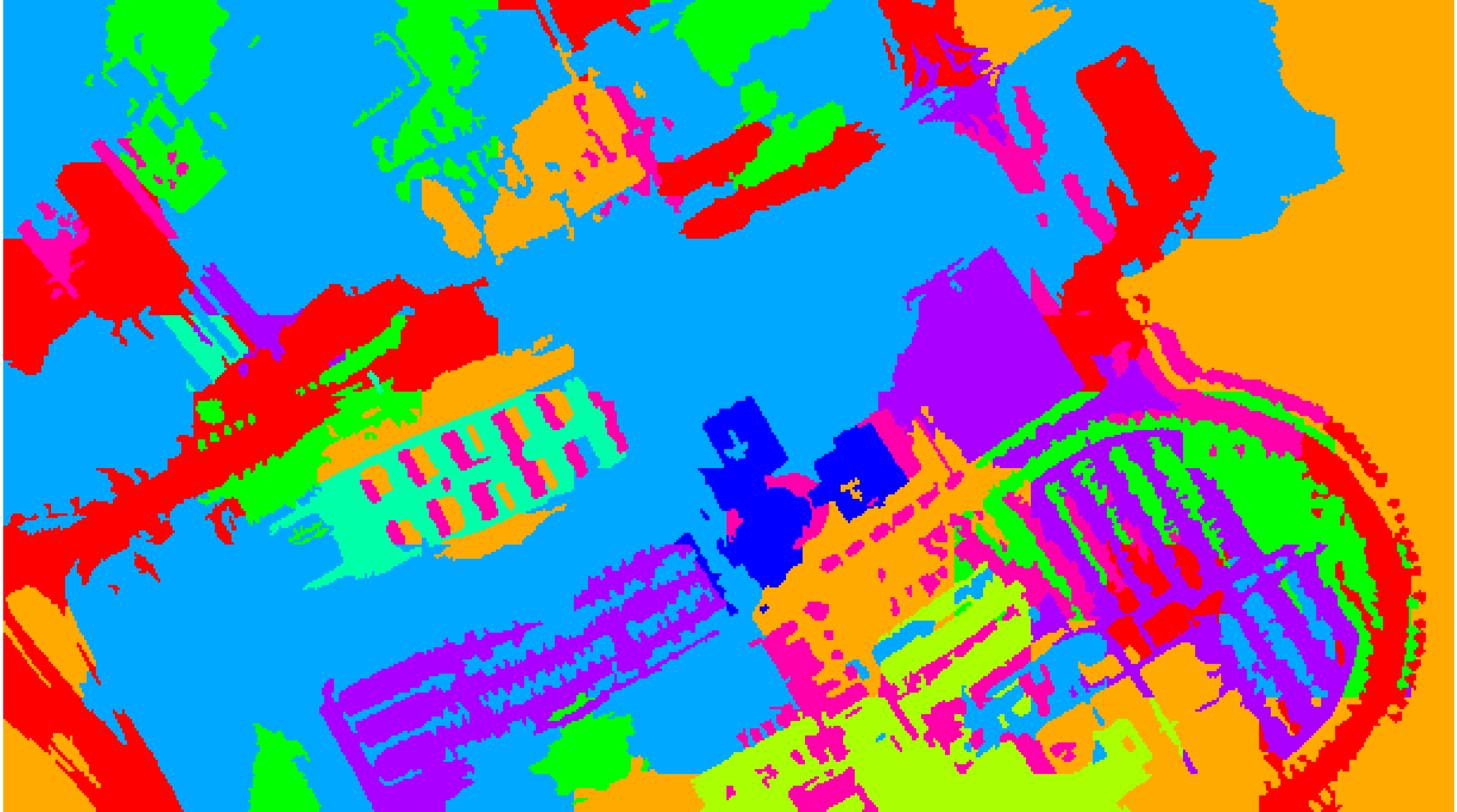}} \hfill
     \subfloat[][400 superpixels]{\includegraphics{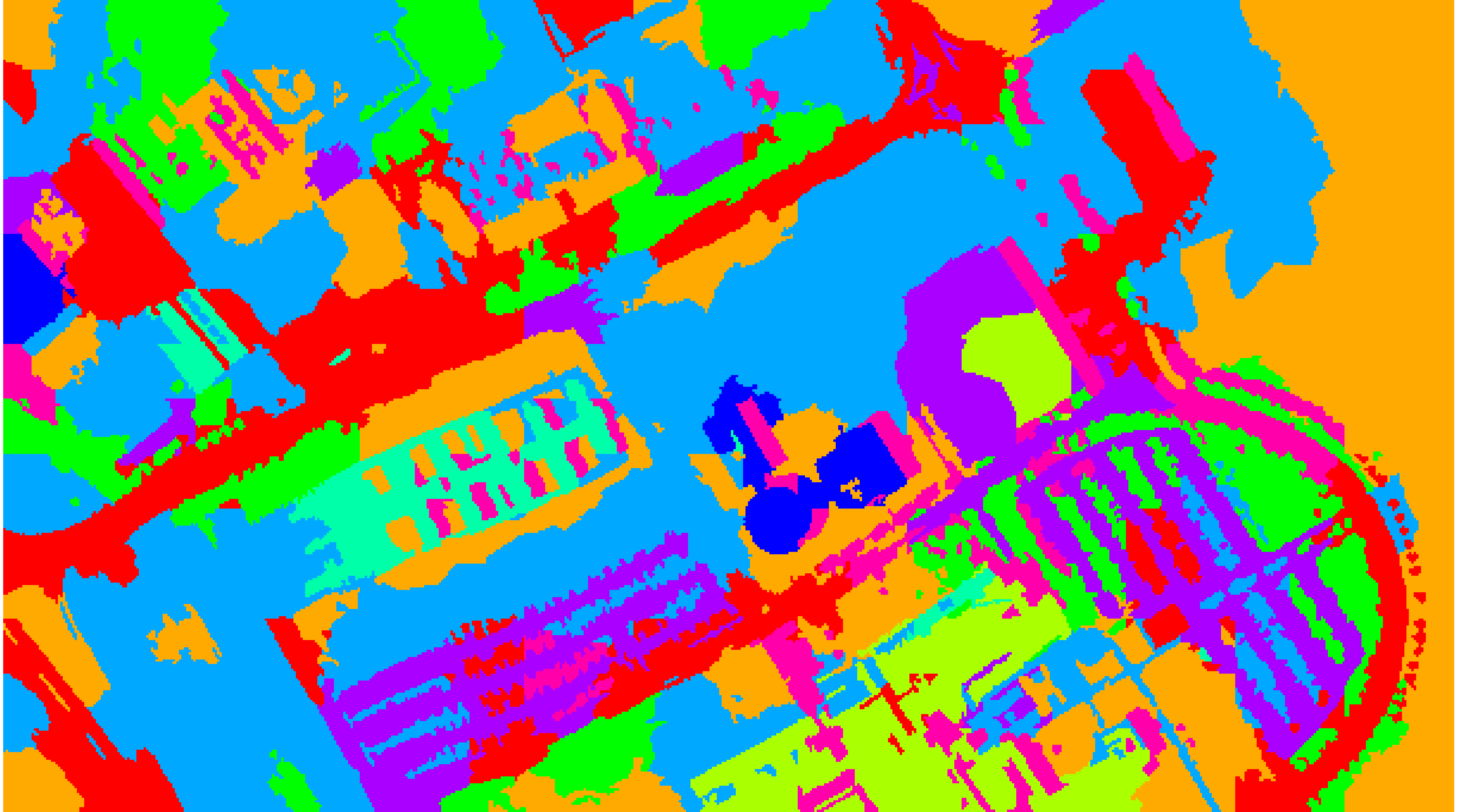}} \hfill
     \subfloat[][800 superpixels]{\includegraphics{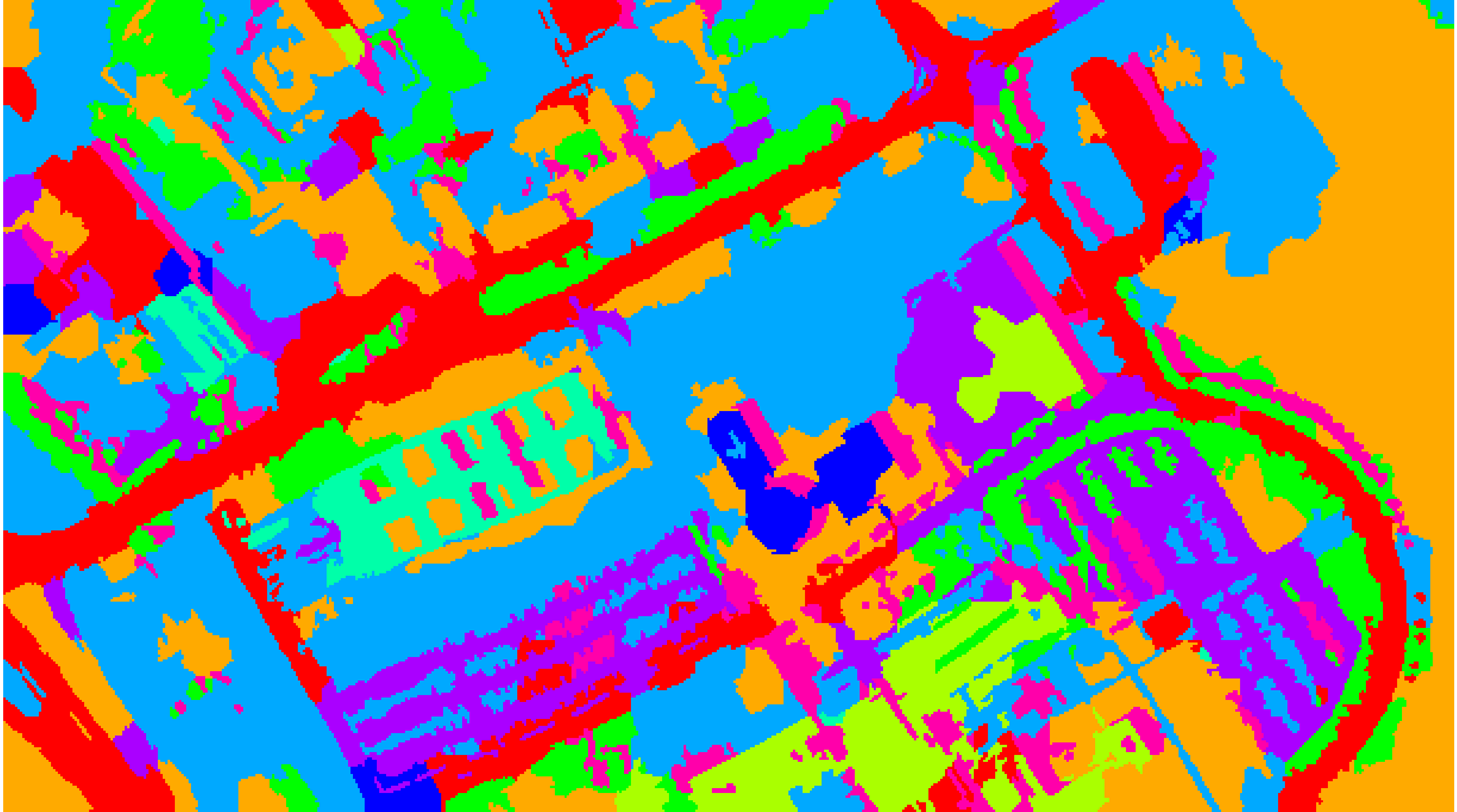}} \hfill
     \subfloat[][1600 superpixels]{\includegraphics{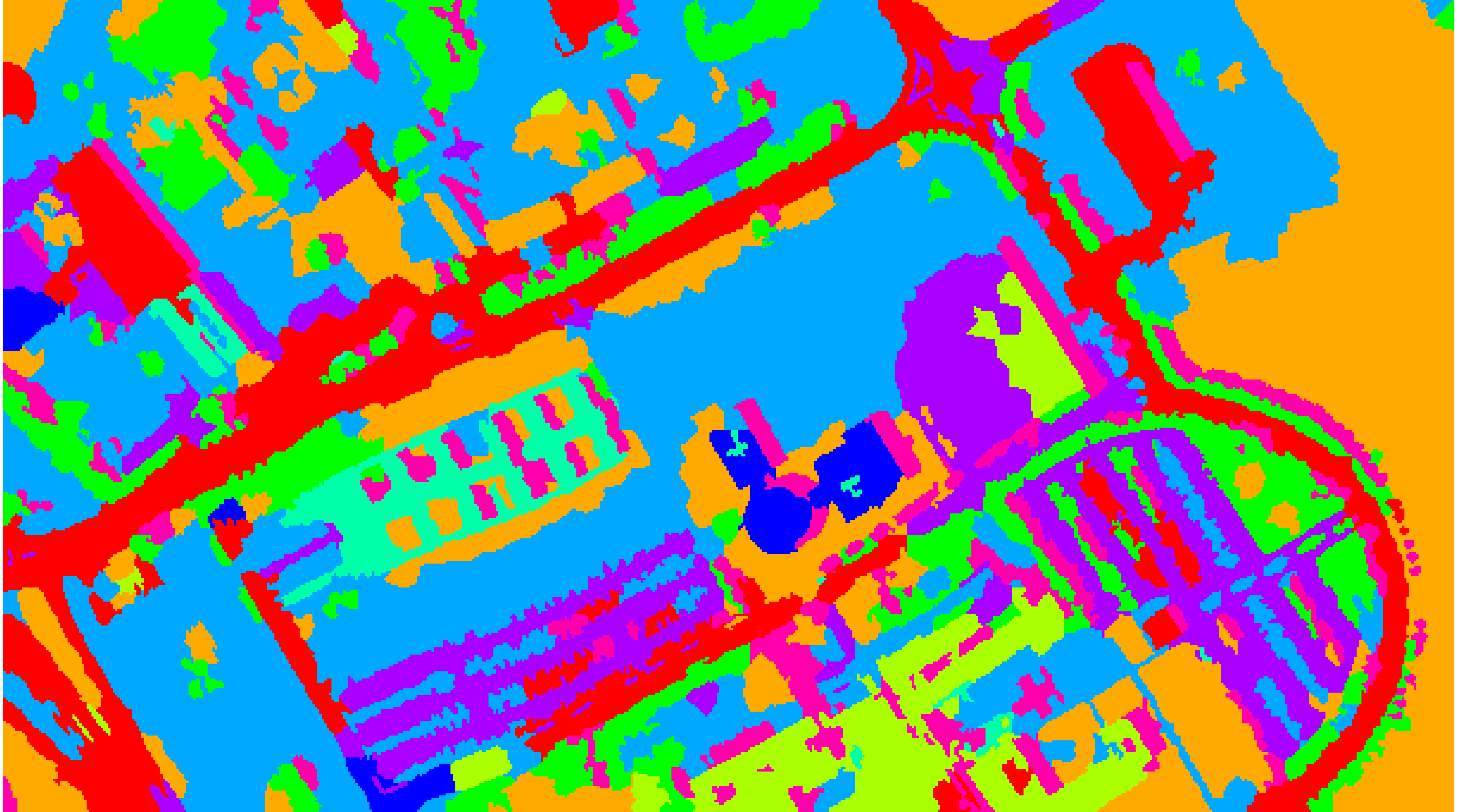}} \hfill
     \subfloat[][3200 superpixels]{\includegraphics{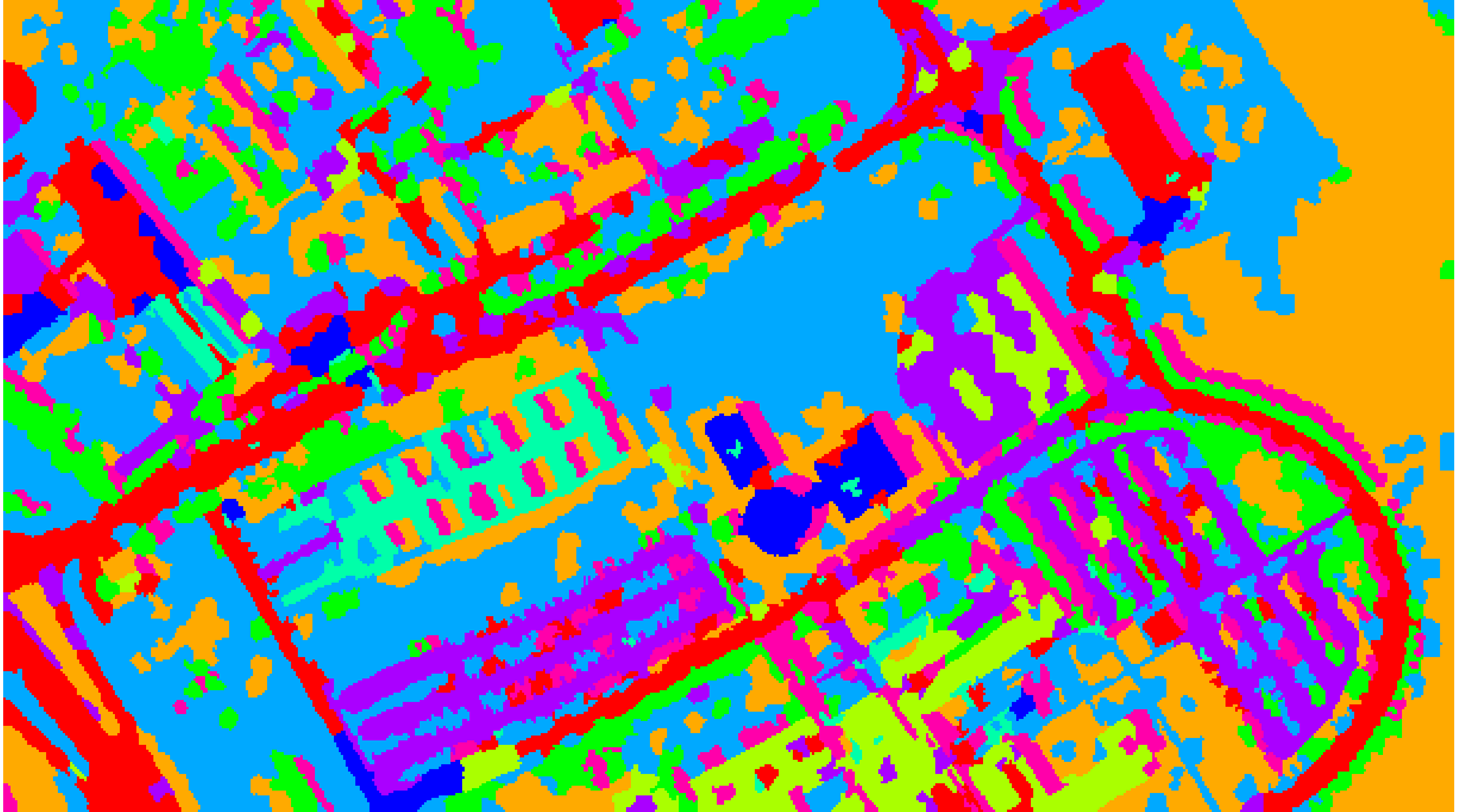}} \hfill
     \subfloat[][6400 superpixels]{\includegraphics{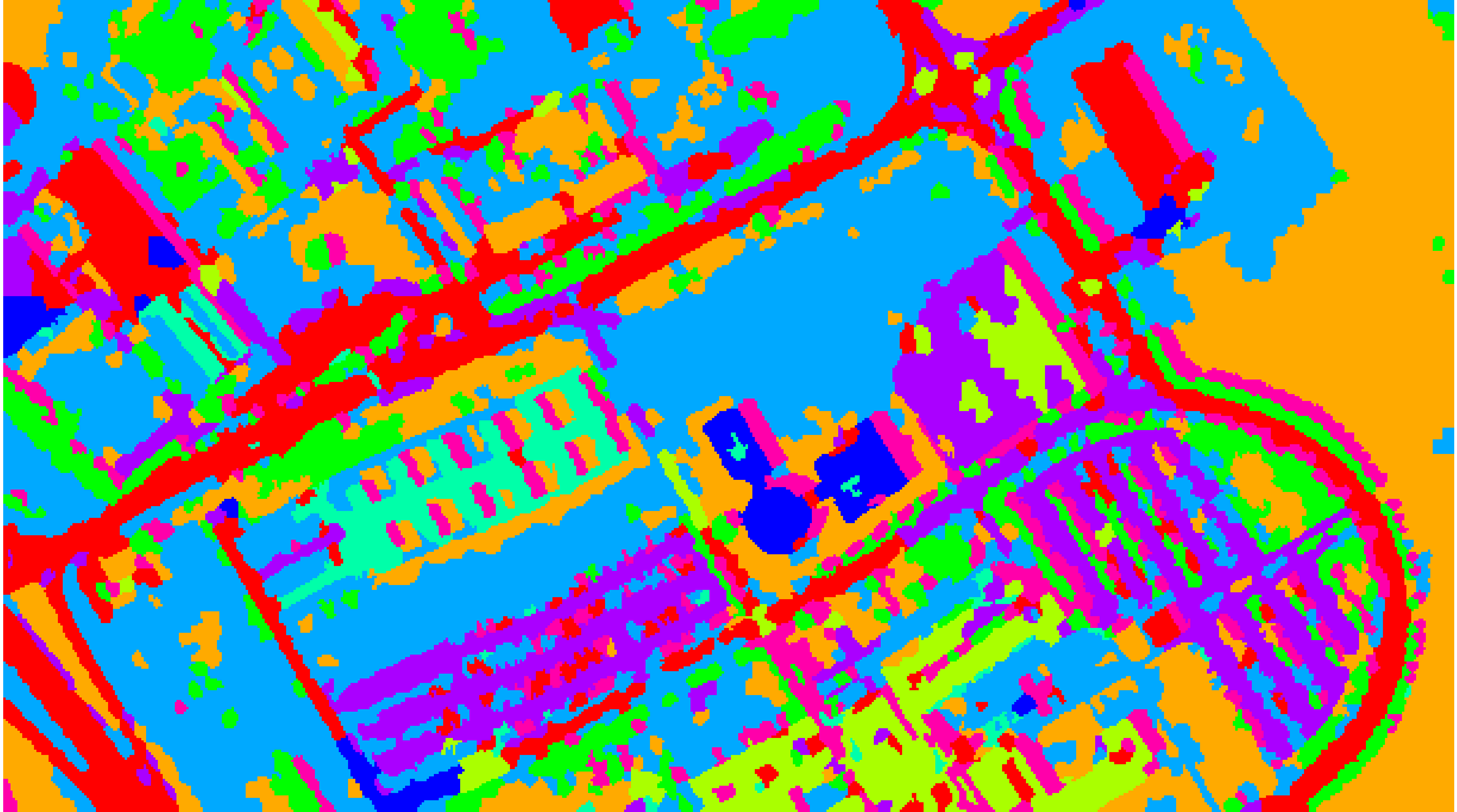}} \hfill
\caption{Superpixel-based MRF applied on the University of Pavia image.}
\label{fig:sup_pix_map2}
\end{figure*}
Figure~\ref{fig:sup_pix_map} and Figure~\ref{fig:sup_pix_map} show the quality of land cover maps produce by superpixel-based MRFs on the Indian pines and the University of Pavia images during one of the trials. For visual comparison, a rendered RGB image and the ground truth map of the datasets have been included.

\subsubsection{Discussion}
The results show consistent patterns for all datasets. The performance of the superpixel-based MRF is poor when there are few superpixels. As the number of superpixels is increased, the performance grows reaching a peak performance, which is equal or better than that of pixel-wise MRF. However, after reaching the peak the performance decreases on increasing the number of superpixels. Furthermore, the superpixel-based MRFs are significantly faster than pixel-based MRFs. It should be noted that the time taken to perform the superpixel segmentation is included in the reported method's time in the table.

Pixel-based MRFs learn spatial relationship between the pixels while the superpixel-based MRFs learn spatial relationship between superpixels, which are group of homogeneous pixels and represent higher order structures. Superpixels may represent objects or different parts of objects. Therefore, superpixel-based MRF should perform better than pixel-wise MRF when the size of the superpixels are optimally representing different components in the image. The results confirm this hypothesis. We see that at optimal scale (size of the superpixel), the superpixel-based MRF performs as good as or outperforms the pixel-wise MRF. The optimal scale is different for different images depending upon the size of objects in the image. It should be also noted that the performance of superpixel-based MRF is dependent on the performance of the superpixel segmentation algorithm. Additionally, since the number of nodes if much smaller in superpixel-based MRF is it much faster than pixel-based MRF, which is a great advantage when we are trying to label very large images. The gain in speed of inference in order to get the same or better performance than pixel-based model is much higher in larger images (Salinas and Pavia city) compared to smaller images (Indian pines) indicating superpixel-based UGM become essential while dealing enormous remote sensing images. 

In Figure~\ref{fig:sup_pix_map} and Figure~\ref{fig:sup_pix_map2}, we see that when fewer superpixels are used the estimated land cover map is of low resolution and small objects in the image are ignored. As the number of superpixels is increased smaller objects become visible in the land cover maps. So superpixel-based MRFs give users an option to control the resolution of the land cover maps and a chance to obtain faster computational time at the cost of decreased resolution.

\section{Summary}
\label{tutorial_summary}
In this tutorial, we provided a broad introduction to modeling and inference in undirected graphical models (UGMs), reviewed UGM based methods developed for remote sensing applications, explained pixel-based and superpixel-based pairwise UGMs in detail, and experimentally evaluated those models on popular hyperspectral datasets. To make it easier for the readers to have hands-on experience with the discussed models, the source code used to implement the methods for the experiments have been published. The elaborate set of experimental results included in this tutorial can serve as baselines for future UGM-based or any other techniques for spatial-spectral classification.  

Pairwise Markov random fields with grid-search for parameter learning seems to be best suited approach for current hyperspectral datasets. Current datasets consists of a moderately sized image with some labeled pixels for training and some for testing. Therefore, only simpler UGMs can be properly trained on them because models with more complex graph and more expressive potential functions have large number of parameters which have to be learning from the data. However, it would be more useful to train and test models on different images, so that once the model is trained it could be applied to any new images obtained from the sensor. Such models would require large amount of labeled images for training. Researchers have already started to develop this kind of datasets for color and multi-spectral images, for example, SpaceNet, Inria aerial image labeling dataset~\cite{maggiori2017can}, and 2017 IEEE GRSS data fusion contest dataset~\cite{tuia20172017}. Unfortunately, such datasets are unavailable for hyperspectral imagery and this have limited the growth of development of more sophisticated and robust mapping methods. With enough labeled training example, in theory we should be able to build mapping models that are robust to variation in geographic locations, image acquisition season and time, image resolution, weather conditions, and sensor technologies. Therefore, development of new benchmark datasets should be one of the top priories of researchers.
 
Data fusion is a fields where UGMs could have a major impact. They have been successfully used to fuse multi-modal images for land cover classification~\cite{liu2017dense,wegner2011building}. However, more interesting applications would arise from the fusion of ground level data, such as digital maps (e.g., OpenStreetMaps) and geotagged data (e.g., photos, online reviews), with the remotely sensed images. The class labels used to describe the points of interest in different sources of data would not have to be the same because UGMs could automatically learn the hierarchical relationships between different them as in \cite{albert2017higher}.

The current popular trend in spatial-spectral classification is to develop deep neural network for feature extraction~\cite{zhu2017deep}. UGMs can be used to complement these methods. As we saw in the experiments, UGMs can boost the performance of spatial-spectral feature based classifiers. Moreover, for cases where the training set consists of spectra from third-party spectral library or ground spectra collected from the scene, spectral features either hand-designed or learned using deep network can not be used. In these cases, UGMs become a useful tool to apply spatial contextual information.  

\bibliographystyle{plain}
\bibliography{tutorial}

\end{document}